\pdfoutput=1
\documentclass{article}

\usepackage{iclr2027_conference}
\usepackage{times}
\usepackage{amsmath,amssymb,amsthm,mathtools}
\usepackage{mathrsfs}
\allowdisplaybreaks
\usepackage{booktabs,tabularx,array,multirow,longtable}
\usepackage{graphicx}
\usepackage{float}
\usepackage{xcolor}
\usepackage{microtype}

\usepackage{hyperref}
\usepackage{cleveref}
\usepackage{listings}
\usepackage{lean_equation_links}
\newcommand{\TrapMasses}{(1.000437388123,\,1.390797880326,\,0.604352649479)}
\newcommand{\TrapAnglesOverPi}{(1.777333876184,\,1.056213642576,\,0.363169886907)}
\newcommand{\TrapBoxRadius}{1/50000}
\newcommand{\TrapGramScale}{10^{-6}}
\newcommand{\TrapGramMatrix}{\begin{pmatrix}
 1744016&118937&353914&0&536774&-434587 \\
 0&1739956&132152&-387017&-36692&304924 \\
 0&0&1702608&766945&-755979&66668 \\
 0&0&0&1414697&-907422&-512806 \\
 0&0&0&0&1844853&-717181 \\
 0&0&0&0&0&339667
 \end{pmatrix}}
\newcommand{\TrapHessianShift}{\frac{1}{10}}
\newcommand{\TrapHessianBound}{\frac{1}{250}}

\newcommand{\TrapCurvatureFloorInline}{1/16}
\newcommand{\TrapShiftMinusSixBound}{\frac{19}{250}}
\newcommand{\TrapSosRemainder}{\frac{27}{2000}}
\newcommand{\TrapGradientScale}{10^{-14}}
\newcommand{\TrapGradientBounds}{(14571,14180,14985,8537,11601,5258)}

\newcommand{\leanbase}{https://github.com/JayPiZimmermann/Removing-spurious-minima-for-planar-features-by-skip-connections/blob/main}
\definecolor{leangray}{gray}{0.35}
\newcommand{\leanref}[3]{\href{\leanbase/#1\#L#2}{\textcolor{leangray}{\texttt{#3}}}}

\newcommand{\leanlink}[3]{\leanref{#1}{#2}{#3}}
\newcounter{leanlinkno}
\newcommand{\lean}[2]{\stepcounter{leanlinkno}\leanref{#1}{#2}{Lean\_\theleanlinkno}}
\newcommand{\leandecl}[3]{\lean{#1}{#2}}
\newcommand{\leanlinkdecl}[4]{\leanlink{#1}{#2}{#4}}
\DeclareLeanEquation{eq:models-raw}{Preliminaries.lean}{845}
\DeclareLeanEquation{eq:models}{Preliminaries.lean}{850}
\DeclareLeanEquation{eq:parameter-classes}{Preliminaries.lean}{793}
\DeclareLeanEquation{eq:losses}{Skip.lean}{399}
\DeclareLeanEquation{eq:q-losses}{Preliminaries.lean}{1613}
\DeclareLeanEquation{eq:raw-chain-rule}{Preliminaries.lean}{2173}
\DeclareLeanEquation{eq:skip-centered-split}{Skip.lean}{388}
\DeclareLeanEquation{eq:skip-connection}{Skip.lean}{443}
\DeclareLeanEquation{eq:loss-connection}{Skip.lean}{498}
\DeclareLeanEquation{eq:kernels}{Preliminaries.lean}{1622}
\DeclareLeanEquation{eq:ambient-residual-main}{Preliminaries.lean}{963}
\DeclareLeanEquation{eq:residual-loss}{Preliminaries.lean}{1732}
\DeclareLeanEquation{eq:residual-mass-gradient}{Preliminaries.lean}{1743}
\DeclareLeanEquation{eq:pair-second-derivative}{Geometry.lean}{547}
\DeclareLeanEquation{eq:pair-closed-form}{Geometry.lean}{558}
\DeclareLeanEquation{eq:term-trace}{Geometry.lean}{1064}
\DeclareLeanEquation{eq:global-trace}{Geometry.lean}{1112}
\DeclareLeanEquation{eq:loss-first-derivative}{Geometry.lean}{3182}
\DeclareLeanEquation{eq:student-first-derivative}{Geometry.lean}{3054}
\DeclareLeanEquation{eq:mixed-second-moment}{Geometry.lean}{1333}
\DeclareLeanEquation{eq:second-moment-pair}{Geometry.lean}{2243}
\DeclareLeanEquation{eq:stretching-convex}{Geometry.lean}{2422}
\DeclareLeanEquation{eq:midpoint-convexity}{Geometry.lean}{2450}
\DeclareLeanEquation{eq:reflection-invariance}{Geometry.lean}{2462}
\DeclareLeanEquation{eq:normal-stretching}{Geometry.lean}{2481}
\DeclareLeanEquation{eq:stretching-secant}{Geometry.lean}{2603}
\DeclareLeanEquation{eq:revival-signed}{Geometry.lean}{4636}
\DeclareLeanEquation{eq:zero-arc-gap-occupancy}{Planar.lean}{3650}
\DeclareLeanEquation{eq:zero-arc-no-negative}{Geometry.lean}{3685}
\DeclareLeanEquation{eq:zero-mass-pairing}{Geometry.lean}{4161}
\DeclareLeanEquation{eq:net-weight-main}{Planar.lean}{2082}
\DeclareLeanEquation{eq:residual-potential-paper}{Preliminaries.lean}{969}
\DeclareLeanEquation{eq:definition_beam_operator}{Planar.lean}{92}
\DeclareLeanEquation{eq:definition_V}{Beam.lean}{82}
\DeclareLeanEquation{eq:gap-green-representation}{Planar.lean}{4715}
\DeclareLeanEquation{eq:LFC}{Planar.lean}{2060}
\DeclareLeanEquation{eq:kernel-jet}{Beam.lean}{642}
\DeclareLeanEquation{eq:LFC-jump}{Planar.lean}{2116}
\DeclareLeanEquation{eq:green-linear-system}{Planar.lean}{1242}
\DeclareLeanEquation{eq:green-determinant}{Planar.lean}{1005}
\DeclareLeanEquation{eq:green-M0}{Planar.lean}{997}
\DeclareLeanEquation{eq:green-A}{Planar.lean}{1001}
\DeclareLeanEquation{eq:green-function}{Planar.lean}{1415}
\DeclareLeanEquation{eq:wronski_beauty}{Planar.lean}{49}
\DeclareLeanEquation{eq:green-pairing}{Planar.lean}{1466}
\DeclareLeanEquation{eq:gap-second-derivative-green}{Planar.lean}{3382}
\DeclareLeanEquation{eq:gap-derivative}{Planar.lean}{3421}
\DeclareLeanEquation{eq:opposed-rotation-curvature}{Planar.lean}{2735}
\DeclareLeanEquation{eq:diagonal-angle-hessian}{Planar.lean}{2631}
\DeclareLeanEquation{eq:opposed-rotation-computation}{Planar.lean}{2705}
\DeclareLeanEquation{eq:beam-descent-direction}{Planar.lean}{5551}
\DeclareLeanEquation{eq:beam-feature-variation}{Beam.lean}{224}
\DeclareLeanEquation{eq:definition_E}{Beam.lean}{238}
\DeclareLeanEquation{eq:beam-second-variation-raw}{Beam.lean}{2523}
\DeclareLeanEquation{eq:beam-gram}{Beam.lean}{2202}
\DeclareLeanEquation{eq:E_first_sum}{Beam.lean}{2217}
\DeclareLeanEquation{eq:beam-hessian-offdiagonal}{Beam.lean}{2400}
\DeclareLeanEquation{eq:beam-hessian-diagonal}{Beam.lean}{2411}
\DeclareLeanEquation{eq:beam-second-variation}{Beam.lean}{2552}
\DeclareLeanEquation{eq:beam-gap-sinusoid}{Beam.lean}{1511}
\DeclareLeanEquation{eq:beam-energy-gapwise}{Beam.lean}{2842}
\DeclareLeanEquation{eq:beam-feature-jumps}{Beam.lean}{1426}
\DeclareLeanEquation{eq:beam-constants}{Beam.lean}{3053}
\DeclareLeanEquation{eq:beam-direction-angle}{Beam.lean}{974}
\DeclareLeanEquation{eq:beam-direction}{Beam.lean}{3079}
\DeclareLeanEquation{eq:beam-energy}{Beam.lean}{3041}
\DeclareLeanEquation{eq:beam-branch}{Beam.lean}{957}
\DeclareLeanEquation{eq:beam-branch-start}{Beam.lean}{1013}
\DeclareLeanEquation{eq:beam-branch-end}{Beam.lean}{1027}
\DeclareLeanEquation{eq:beam-branch-jumps}{Beam.lean}{1619}
\DeclareLeanEquation{eq:beam-glued-profile}{Beam.lean}{2926}
\DeclareLeanEquation{eq:beam-glued-feature}{Beam.lean}{1804}
\DeclareLeanEquation{eq:beam-periodic-antiperiodic}{Beam.lean}{1253}
\DeclareLeanEquation{eq:beam-branch-energy}{Beam.lean}{1123}
\DeclareLeanEquation{eq:beam-energy-proof}{Beam.lean}{2962}
\DeclareLeanEquation{eq:beam-descent-identity}{Beam.lean}{3188}
\DeclareLeanEquation{eq:beam-target}{Beam.lean}{3154}
\DeclareLeanEquation{eq:beam-gap-data}{Planar.lean}{3856}
\DeclareLeanEquation{eq:beam-curvature-ratio}{Beam.lean}{2638}
\DeclareLeanEquation{eq:beam-split}{Beam.lean}{2661}
\DeclareLeanEquation{eq:beam-gap-sum}{Beam.lean}{2690}
\DeclareLeanEquation{eq:beam-gap-inequality}{Beam.lean}{2625}
\DeclareLeanEquation{eq:beam-bracket-factorization}{Beam.lean}{508}
\DeclareLeanEquation{eq:beam-ratio-derivative}{Beam.lean}{376}
\DeclareLeanEquation{eq:trap-taylor}{Plain.lean}{380}
\DeclareLeanEquation{eq:merge-cross-term}{Plain.lean}{9306}
\DeclareLeanEquation{eq:trap-student-masses}{Plain.lean}{154}
\DeclareLeanEquation{eq:trap-student-angles}{Plain.lean}{158}
\DeclareLeanEquation{eq:trap-positive-loss}{Plain.lean}{739}
\DeclareLeanEquation{eq:sine-load-identity}{Plain.lean}{2871}
\DeclareLeanEquation{eq:trap-seed-curvature}{Plain.lean}{9070}
\DeclareLeanEquation{eq:ceiling-count}{Plain.lean}{2118}
\DeclareLeanEquation{eq:corner-free-boundary}{Plain.lean}{1285}
\DeclareLeanEquation{eq:jump-summation}{Plain.lean}{1387}
\DeclareLeanEquation{eq:duhamel-sum}{Plain.lean}{1560}
\DeclareLeanEquation{eq:gap-orthogonality}{Plain.lean}{2901}
\DeclareLeanEquation{eq:empirical-loss}{Empirical.lean}{615}
\DeclareLeanEquation{eq:empirical-loss-mismatch}{Empirical.lean}{207}
\DeclareLeanEquation{eq:modulus}{Empirical.lean}{584}
\DeclareLeanEquation{eq:max-product-distance}{Empirical.lean}{159}
\DeclareLeanEquation{eq:data-scaling}{Empirical.lean}{860}
\DeclareLeanEquation{eq:data-scaling-loss}{Empirical.lean}{893}
\DeclareLeanEquation{eq:data-coercivity}{Empirical.lean}{1090}
\DeclareLeanEquation{eq:data-parameter-bounds}{Empirical.lean}{1216}
\DeclareLeanEquation{eq:uniform-comparison}{Empirical.lean}{665}
\DeclareLeanEquation{stp:zm-revival}{Geometry.lean}{4738}
\DeclareLeanEquation{stp:zm-propagation}{Geometry.lean}{4517}
\DeclareLeanEquation{stp:zm-pairing}{Geometry.lean}{4763}
\DeclareLeanEquation{stp:green-beam-equation}{Planar.lean}{111}
\DeclareLeanEquation{stp:green-function}{Planar.lean}{1415}
\DeclareLeanEquation{stp:green-representation}{Planar.lean}{1510}
\DeclareLeanEquation{stp:green-signs}{Planar.lean}{3480}
\DeclareLeanEquation{stp:beam-second-variation}{Beam.lean}{2581}
\DeclareLeanEquation{stp:beam-mode}{Beam.lean}{3097}
\DeclareLeanEquation{stp:beam-gap-inequality}{Beam.lean}{3260}
\DeclareLeanEquation{stp:plain-box}{Plain.lean}{9070}
\DeclareLeanEquation{stp:plain-curvature}{Plain.lean}{9070}
\DeclareLeanEquation{stp:plain-splitting}{Plain.lean}{10152}
\DeclareLeanEquation{stp:finite-data-compactness}{Empirical.lean}{1216}
\DeclareLeanEquation{stp:finite-data-uniform}{Empirical.lean}{708}
\DeclareLeanEquation{stp:finite-data-stability}{Empirical.lean}{386}
\DeclareLeanEquation{stp:green-positivity}{Planar.lean}{1137}
\DeclareLeanEquation{eq:relu-pair-angular}{Preliminaries.lean}{1554}
\DeclareLeanEquation{eq:loss-quadratic}{Preliminaries.lean}{1692}
\DeclareLeanEquation{eq:plain-correction-trace}{Geometry.lean}{1806}
\DeclareLeanEquation{eq:proportional-pair-quadratic}{Geometry.lean}{592}
\DeclareLeanEquation{eq:green-ansatz}{Planar.lean}{1384}
\DeclareLeanEquation{eq:green-ratio-derivative}{Planar.lean}{1044}
\DeclareLeanEquation{eq:green-numerator-ratio}{Planar.lean}{1105}
\DeclareLeanEquation{eq:width-one-centered-loss}{Geometry.lean}{3959}
\DeclareLeanEquation{eq:beam-hessian-offdiagonal-masses}{Beam.lean}{2400}
\DeclareLeanEquation{eq:beam-hessian-offdiagonal-mass-angle}{Beam.lean}{2400}
\DeclareLeanEquation{eq:beam-hessian-diagonal-mass}{Beam.lean}{2411}
\DeclareLeanEquation{eq:beam-hessian-diagonal-mass-angle}{Beam.lean}{2411}
\DeclareLeanEquation{eq:beam-gap-scalar}{Beam.lean}{496}
\DeclareLeanEquation{eq:beam-M}{Beam.lean}{437}
\DeclareLeanEquation{eq:beam-Theta}{Beam.lean}{416}
\DeclareLeanEquation{eq:one-student-quadratic}{Plain.lean}{10665}
\DeclareLeanEquation{eq:one-student-arc}{Plain.lean}{10788}
\DeclareLeanEquation{eq:teacher-potential-expansion}{Plain.lean}{10702}
\DeclareLeanEquation{eq:box-hypotheses}{Plain.lean}{582}
\DeclareLeanEquation{eq:box-boundary-gap}{Plain.lean}{429}
\DeclareLeanEquation{eq:positive-partition}{Theorems.lean}{465}
\DeclareLeanEquation{eq:plain-loss-normalization}{Plain.lean}{8659}
\DeclareLeanEquation{eq:child-barycenter}{Plain.lean}{9182}
\DeclareLeanEquation{eq:split-local-comparison}{Plain.lean}{9518}
\DeclareLeanEquation{eq:seed-gram-matrix}{Plain.lean}{8167}
\DeclareLeanEquation{eq:seed-hessian-enclosure}{Plain.lean}{8981}
\DeclareLeanEquation{eq:seed-hessian-sos}{Plain.lean}{8441}
\DeclareLeanEquation{eq:seed-curvature-margin}{Plain.lean}{327}
\DeclareLeanEquation{eq:data-moment-absorption}{Empirical.lean}{1158}
\DeclareLeanEquation{eq:residual-angle-derivative}{Preliminaries.lean}{1994}
\DeclareLeanEquation{eq:directional-hessian-notation}{Geometry.lean}{1171}
\DeclareLeanEquation{eq:jumper_function}{Planar.lean}{984}
\DeclareLeanEquation{eq:jumper_derivative}{Planar.lean}{1010}
\DeclareLeanEquation{eq:reciprocity_green}{Planar.lean}{1343}
\DeclareLeanEquation{eq:green-sign-claims}{Planar.lean}{3372}
\DeclareLeanEquation{eq:Mzero_positive}{Planar.lean}{1074}
\DeclareLeanEquation{eq:data-teacher-bound}{Empirical.lean}{1014}
\DeclareLeanEquation{stp:reductions}{Geometry.lean}{1182}
\DeclareLeanEquation{stp:zn_hessian_kernel}{Geometry.lean}{1281}
\DeclareLeanEquation{stp:conclusion}{Geometry.lean}{1654}
\DeclareLeanEquation{stp:comments_plain_relu}{Geometry.lean}{1944}
\DeclareLeanEquation{eq:loss-quadratic-regrouped}{Preliminaries.lean}{1732}
\DeclareLeanEquation{eq:ring-operator}{BentRing.lean}{26}
\DeclareLeanEquation{eq:angular-kernel-d2+1}{Preliminaries.lean}{1208}
\DeclareLeanEquation{eq:green-decomposition}{Planar.lean}{1510}
\DeclareLeanEquation{eq:green-system}{Planar.lean}{1242}
\DeclareLeanEquation{eq:green-construction}{Planar.lean}{1384}
\DeclareLeanEquation{eq:net-weight-plain}{Plain.lean}{2342}
\DeclareLeanEquation{eq:sine-load-jump}{Plain.lean}{1525}
\DeclareLeanEquation{eq:angular-kernel}{Planar.lean}{425}
\lstdefinestyle{lean}{
  basicstyle=\fontencoding{T1}\fontfamily{fvm}\fontsize{7.6pt}{9pt}\selectfont,
  columns=fullflexible,
  keepspaces=true,
  breaklines=true,
  breakindent=2em,
  breakatwhitespace=true,
  showstringspaces=false,
  sensitive=true,
  upquote=true,
  escapeinside={(*@}{@*)},
  aboveskip=4pt,
  belowskip=4pt,
  xleftmargin=0.5em,
  morekeywords={theorem,def,abbrev,inductive,where,deriving,fun,namespace,end,open,local,instance,Prop,Type,in},
  keywordstyle=\color[rgb]{0.35,0.15,0.6}\bfseries,
  morecomment=[s]{/-}{-/},
  morecomment=[l]{--},
  commentstyle=\color[gray]{0.42}\itshape,
  literate=
    {ℝ}{{$\mathbb{R}$}}1
    {ℕ}{{$\mathbb{N}$}}1
    {ℤ}{{$\mathbb{Z}$}}1
    {∀}{{$\forall$}}1
    {∃}{{$\exists$}}1
    {∧}{{$\wedge$}}1
    {∨}{{$\vee$}}1
    {¬}{{$\neg$}}1
    {→}{{$\rightarrow$}}1
    {↔}{{$\leftrightarrow$}}1
    {↦}{{$\mapsto$}}1
    {×}{{$\times$}}1
    {‖}{{$\|$}}1
    {⟪}{{$\langle\!\langle$}}1
    {⟫}{{$\rangle\!\rangle$}}1
    {⟨}{{$\langle$}}1
    {⟩}{{$\rangle$}}1
    {∑}{{$\textstyle\sum$}}1
    {∫}{{$\textstyle\int$}}1
    {⁻¹}{{$^{-1}$}}2
    {•}{{$\bullet$}}1
    {∘}{{$\circ$}}1
    {≤}{{$\le$}}1
    {≥}{{$\ge$}}1
    {≠}{{$\ne$}}1
    {∈}{{$\in$}}1
    {∉}{{$\notin$}}1
    {⊆}{{$\subseteq$}}1
    {π}{{$\pi$}}1
    {θ}{{$\theta$}}1
    {…}{{$\dots$}}1
    {β}{{$\beta$}}1
    {γ}{{$\gamma$}}1
    {δ}{{$\delta$}}1
    {ε}{{$\varepsilon$}}1
    {τ}{{$\tau$}}1
    {α}{{$\alpha$}}1
    {μ}{{$\mu$}}1
    {σ}{{$\sigma$}}1
    {ω}{{$\omega$}}1
    {₀}{{$_0$}}1
    {₁}{{$_1$}}1
    {₂}{{$_2$}}1
    {��}{{$\mathcal{N}$}}1
    {ᶠ}{{$^{\mathrm{f}}$}}1
    {∞}{{$\infty$}}1
}
\hypersetup{
  hidelinks,
  pdftitle={Removing spurious minima for planar features by skip connections},
  pdfauthor={Jakob Paul Zimmermann, Moritz Grillo, Andrei Balakin, Georg Loho}
}
\usepackage{url}

\title{Removing spurious minima for planar features by skip connections}

\author{Jakob Paul Zimmermann$^{1,2}$ \quad
        Moritz Grillo$^{3}$ \quad
        Andrei Balakin$^{2}$ \quad
        Georg Loho$^{1}$ \\
        $^{1}$Freie Universit\"at Berlin \quad
        $^{2}$Technische Universit\"at Berlin \\
        $^{3}$Max Planck Institute for Mathematics in the Sciences, Leipzig}

\newtheorem{theorem}{Theorem}
\newtheorem{proposition}[theorem]{Proposition}
\newtheorem{lemma}[theorem]{Lemma}

\newtheorem{conjecture}[theorem]{Conjecture}
\theoremstyle{definition}
\newtheorem{definition}[theorem]{Definition}
\newtheorem{remark}[theorem]{Remark}

\usepackage{refcount}
\newcommand{\pfnum}{??}
\newcounter{pfclaim}\newcounter{pfstep}
\newcounter{pfblock}
\renewcommand{\thepfclaim}{\pfnum.\arabic{pfclaim}}
\renewcommand{\thepfstep}{\pfnum.\arabic{pfstep}}

\newenvironment{Proof}[2]{%
  \stepcounter{pfblock}%
  \setcounter{pfclaim}{0}\setcounter{pfstep}{0}%
  \xdef\pfnum{\getrefnumber{#2}}%
  \begin{proof}[\normalfont\bfseries\phantomsection\label{pf:#2}Proof of #1~\ref{#2}]%
}{\end{proof}}
\newenvironment{ProofSketch}[2]{%
  \stepcounter{pfblock}%
  \setcounter{pfclaim}{0}\setcounter{pfstep}{0}%
  \xdef\pfnum{\getrefnumber{#2}}%
  \begin{proof}[\normalfont\bfseries\phantomsection\label{pf:#2}Proof ideas of #1~\ref{#2}]%
}{\end{proof}}

\newcommand{\pfstep}[2]{\par\medskip\refstepcounter{pfstep}%
  \noindent\textbf{\LeanLinkIfDeclared{#1}{Step \thepfstep}\ (#2).}\label{#1}\enspace\ignorespaces}
\newcommand{\finishes}[2]{This finishes the proof of #1~\ref{#2}.}
 {\par}
 {\end{proof}}

\newcommand{\R}{\mathbb{R}}
\newcommand{\sphere}{\mathbb{S}}
\newcommand{\E}{\mathbb{E}}

\newcommand{\LC}{\mathscr{L}_{C}}
\newcommand{\LR}{\mathscr{L}_{R}}
\newcommand{\Lskip}{\mathscr{L}_{\mathrm{skip}}}

\newcommand{\Nskip}[2]{\mathcal{N}_{\mathrm{skip}}(#1,#2)}
\newcommand{\Pnn}[2]{\mathcal{P}^{+}(#1,#2)}
\newcommand{\Ppos}[2]{\mathcal{P}^{++}(#1,#2)}
\newcommand{\Ppl}[2]{\mathcal{P}_{\mathrm{pl}}(#1,#2)}
\newcommand{\ip}[2]{\left\langle #1,#2\right\rangle}

\newcommand{\Z}{\mathbb{Z}}

\iclrfinalcopy

\begin{document}

\maketitle

\begin{abstract}
    Understanding loss landscapes is central to explaining neural-network training, yet their structure remains only partially understood even in simple models. We study the Gaussian population loss of shallow, bias-free ReLU networks in the teacher--student setting.
    This provides a simple model for studying essential aspects such as feature learning and overparameterization.
    For teacher networks with positive output weights and planar features, we show that including a learned linear skip removes all spurious local minima with non-negative student output weights once the student network is at least as wide as the teacher network. In contrast, without the skip, we construct a fixed teacher network with positive output weights and only three hidden neurons in input dimension two whose spurious local minima persist at every student width at least three. Thus, a learned linear skip can remove spurious minima that persist under arbitrary overparameterization.
    Furthermore, we show that a positive output weight student network always learns the subspace spanned by the teacher features: student features at local minima with non-negative student output weights lie in the span of the teacher features. For ReLU networks in two dimensions, even heavily overparameterized student networks have effective width controlled by the teacher width: every critical point with positive student output weights has at most twice as many distinct student feature directions as teacher neurons. Finally, we transfer the benignity result to empirical minima
    over parameter balls of any prescribed radius, with the required sampling accuracy depending on that radius.
\end{abstract}

\section{Introduction}
Training a neural network typically requires solving a highly non-convex
optimization problem, and understanding the corresponding loss landscape is
a central problem in machine learning. Overparameterization and adding residual connections are both associated with easier training.
Sufficient overparameterization enables convergence guarantees for gradient
descent from random initialization under suitable assumptions
\citep{DuZhaiPoczosSingh2019}.
Empirical studies show improved training with residual connections
\citep{HeZhangRenSun2016,HeZhangRenSunIdentity2016} and two-dimensional loss visualization suggest less rugged loss landscapes with residual connections \citep{LiEtAl2018}. We ask when increasing width or adding
a linear skip connection actually excludes spurious local minima.

Teacher--student models provide a standard setting for studying loss
landscapes and training dynamics
\citep{Tian2017,SafranShamir2018,ArjevaniField2021,ZhouGeJin2021}.
When the student network architecture contains the teacher network's, the problem is
realizable: zero population loss is attainable. Moreover, it is a simple model to study essential aspects such as feature learning and overparameterization. The latter can be modeled by allowing the student network to have more hidden neurons than the teacher network. For a shallow teacher network
without a skip, labels depend only on the input's projection onto the span
of its \emph{feature directions}, namely the normalized non-zero input-weight
vectors of its hidden neurons. It is therefore a particular multi-index model and provides a
controlled setting for asking whether training identifies the
target-relevant subspace.
Prior work studies such subspace learning through gradient-based training
\citep{DamianLeeSoltanolkotabi2022,MousaviHosseiniEtAl2023}; this leaves open to what extent subspace learning is a property of the optimization dynamics or of the loss landscape itself.

Overparameterization also raises the question whether additional neurons
learn distinct features or become redundant. We measure \emph{effective
width} by the number of distinct feature directions among neurons with
non-zero contribution, and ask whether criticality alone bounds it, even when
the student network does not fit the teacher network.

Realizability alone does not exclude spurious local minima: for shallow
ReLU networks with Gaussian inputs they exist even for teacher networks with orthogonal feature directions
\citep{SafranShamir2018}, in families whose Hessian spectra are known
\citep{ArjevaniField2021}, and increasing width turns only certain of them
into saddles \citep{SafranYehudaiShamir2021,ArjevaniField2022}.
It remains natural to ask whether an architectural change can remove
minima that survive arbitrary overparameterization.

We study Gaussian population squared loss for bias-free, scalar-output,
one-hidden-layer ReLU networks.
The population setting retains the non-convex
geometry of feature learning while separating it from finite-sample effects,
making it a useful model for isolating the roles of width and architecture.
At the same time, population-level landscape guarantees can inform the
finite-data setting.
We compare the respective loss landscapes of one-hidden-layer ReLU networks with and without a linear input-to-output skip connection. We refer to these models as
\emph{plain network} and \emph{skip network}.

\subsection{Our Contributions}
\label{sec:contributions}

We establish five results relating the loss landscape to the architecture,
the teacher geometry, and the student width.

\begin{itemize}
\item \emph{Main result:}
A linear skip eliminates spurious local minima under the following assumptions:
Suppose that the teacher network and the student network are both skip networks and the teacher network
has positive output weights and feature directions in a two-dimensional linear subspace.
Whenever the student network is at least as wide as the teacher network, every local minimum
of the Gaussian population loss with non-negative student output weights
has zero loss (\Cref{thm:headline}).

\item \emph{Limits of overparameterization:}
Increasing width alone need not eliminate spurious local minima.
More precisely, we construct as teacher network a plain network in input
dimension two with three hidden neurons and positive output weights.
At every student width at least three, its population loss admits a
spurious local minimum with positive student output weights
(\Cref{thm:plain-trap}).
Our main result (\Cref{thm:headline}) excludes such minima once a linear skip is added.

\item \emph{Feature learning:}
Local minimality confines student features to the teacher subspace.
More precisely, with or without a linear skip, at every local minimum with
non-negative student output weights, each student neuron with non-zero output
weight has its feature direction in the span of the teacher network's
feature directions (\Cref{prop:teacher-span-confinement}).

\item \emph{Effective width:}
At critical points, the teacher width limits the number of student features.
More precisely, for a plain network in input dimension two, every
critical point of the population loss with positive student output weights
has at most twice as many distinct feature directions as the teacher network
has hidden neurons.
This bound is sharp (\Cref{prop:plain-collision-ceiling}).

\item \emph{Finite sample consequence:}
We extend the main result from population loss to empirical loss given some additional local structure.
More precisely, assume that the same conditions are fulfilled as in the `main result' and that the teacher-generated samples are sufficiently accurate.
Then, a locally optimal student network has zero empirical loss if the locality depends on the accuracy
(\Cref{thm:fixed-radius-bridge}).
\end{itemize}

The key to our analysis of skip networks is the identity $[z]_+=\tfrac12|z|+\tfrac12z$:
a plain network splits into a \emph{centered} network with activation
$|z|/2$ and a linear map that a learned skip can absorb
\citep{ZhouGeJin2021}. Centering
removes exactly the first angular Fourier harmonic and leaves the even ones
unchanged, so the centered network is $\pi$-periodic in the direction angle; for non-negative masses
the split is a translation of zonotopes
(\Cref{fig:zonotope-first-harmonic} in \Cref{app:preliminary-proofs}).

\paragraph{Modern paper format.}
We decided on a novel structure for this paper to achieve various desirable aspects in the age of AI-generated mathematics.
The first 9 pages give an introduction and overview of the main results, explaining the fundamental concepts and some of the interesting proof ideas.
The appendix does not contain full proofs but all proof ideas supporting human understanding involved in the deduction of our main results extracted by the authors from LLM-generated proofs.
For proper independent verification, we link our repository containing an auto-formalization of the results;
these Lean certificates follow the same proof structure and calculations omitted from the exposition (\Cref{app:lean-certificates}).
We complement this by adding the formalization of the main results and relevant definitions in the appendix so that one can check that they agree with the natural language statements given in the first 9 pages.

\paragraph{Human--AI collaboration and proof certification.}
Generative AI played a substantial role in the mathematical discovery and proof search in Lean underlying this work.  For perspectives on AI in mathematics, see \cite{Tao2026ICM,Avigad2026,Commelin2026}.
The problem was particularly amenable to this approach: it is basic yet timely, has a natural hierarchy of subproblems and levels of generality, and involves substantial computation whose conceptual consequences are more informative than its details. Numerical exploration and formal proof search thus helped reveal low-level patterns that were progressively distilled into general statements.
\Cref{app:ai-use} details the discovery process, unsuccessful approaches, human and AI contributions, and verification procedures; the supplementary material also includes the final agent skills as Markdown (\texttt{.md}) files. Our contribution is mathematical rather than an evaluation of an agentic proof-search system.

\subsection{Related Work}

\paragraph{Teacher--student models.}
For shallow ReLU networks with Gaussian inputs and fixed output weights,
\citet{Tian2017} derives explicit population gradients and uses them
to analyze critical points and convergence.
\citet{SafranShamir2018} give a computer-assisted
construction of spurious local minima even for teacher networks with orthogonal teacher feature directions.
\citet{LiYuan2017} prove SGD convergence with a \emph{pre-activation} identity connection under Gaussian inputs and a prescribed random-initialization regime; unlike our learned input-to-output skip, it acts before the activation.
More recently, \citet{TinazXieSoltanolkotabi2026} study the learning
of a linear target by a ReLU network, showing that gradient descent from a moderately small random initialization converges globally at a linear rate despite the presence of non-strict saddle points.

\paragraph{Population loss with a learned linear term.}
\citet{ZhouGeJin2021} study Gaussian population regression for one-hidden-layer networks with absolute-value activation and non-negative output weights, and derive its equivalence to ReLU regression with an additional learned linear term. They prove gradient-descent convergence from a sufficiently small-loss initialization for teacher networks of any feature rank with sufficiently separated feature directions and any student width at least the teacher network's. In contrast, for rank-at-most-two teacher networks we characterize the full landscape: every local minimum with non-negative output weights is an exact fit, without a small-loss or teacher-separation assumption, thereby ruling out positive-loss local minima left open by their local convergence analysis. Their proof is perturbative around the teacher configuration, using low loss to aggregate student neurons around teacher neurons and construct a descent direction, whereas our residual formulation globally encodes the loss and its first derivatives and enables structural arguments from local minimality.
For other residual architectures, local-minimum guarantees compare risk with that of the best linear predictor under architectural and regularity assumptions \citep{Shamir2018,KawaguchiBengio2019,YunSraJadbabaieResNet2019}.
Such bounds need not imply exact fitting of a non-linear teacher network,
unlike our zero-loss conclusion under the stated assumptions.
\citet{LiangSunLeeSrikant2018} remove bad local minima for suitable
empirical classification losses using a regularized exponential
neuron; we instead use an unregularized linear skip.

\paragraph{Learning the relevant subspace.}
\citet{DamianLeeSoltanolkotabi2022} show that gradient descent can learn representations depending only on the relevant subspace of low-dimensional polynomial targets, yielding a sample-complexity advantage over kernel methods.
For Gaussian multi-index models, \citet{MousaviHosseiniEtAl2023} prove that online SGD with weight decay drives the first-layer weights toward the target-relevant subspace and use this to obtain width-independent generalization bounds.

\paragraph{Duplicated neurons.}
\citet{SimsekEtAl2021} analyze how neuron replication and permutation
symmetries generate connected sets of global minima and families of
critical points in overparameterized networks.
When a network is widened by duplicating neurons without changing
the function it represents, a local minimum of the original network
may remain a local minimum or become a saddle point in the wider
network \citep{FukumizuAmari2000,FukumizuYamaguchiMototakeTanaka2019,WuSimsekGed2025}.

\paragraph{Empirical loss landscapes.}

Sample-dependent width bounds can exclude spurious valleys without
excluding non-strict spurious minima \citep{VenturiBandeiraBruna2019}.
Convex reformulations characterize optima and stationary points of
regularized two-layer ReLU networks
\citep{PilanciErgen2020,WangLacottePilanci2022,KimMishkinPilanci2025}.
Our transfer uses fixed-radius minimality and uniform convergence
of loss values, rather than of gradients
\citep{FosterSekhariSridharan2018}.

\section{Preliminaries}
\label{sec:preliminaries}
Unless stated otherwise, $d\ge2$ is the input dimension; statements that also hold in dimension one, among them our main results, say so explicitly. Throughout, $X\sim\mathcal N(0,I_d)$, and
$q\in\{C,R\}$ labels the two \emph{features} $\varphi_C(z)=|z|/2$ (centered)
and $\varphi_R(z)=[z]_+$ (plain).

\begin{definition}[Units, networks, and parameter classes]
\label{def:networks}
A \emph{one-hidden-layer network of width $n\ge1$} has a weight vector $a_i\in\R^d$ and an output weight $o_i\in\R$
per hidden neuron and \emph{represents} the function
\begin{align}
 f_{q;o,a}(x)=\sum_{i=1}^{n}o_i\,\varphi_q(\ip{a_i}{x}),\qquad
 (o,a)\in\R^n\times(\R^d)^n .
 \label{eq:models-raw}
\end{align}
We write $\Nskip{d}{n}$ for the class of functions of the form
$F_{b,o,a}(x)=\ip{b}{x}+f_{R;o,a}(x)$ represented by a plain network together with a \emph{linear skip vector}
$b\in\R^d$.
\end{definition}

Both features are positively homogeneous, so
$o_i\varphi_q(\ip{a_i}{x})=o_i\|a_i\|\,\varphi_q(\ip{a_i/\|a_i\|}{x})$
whenever $a_i\ne0$: only the product $o_i\|a_i\|$ and the direction $a_i/\|a_i\|$ enter the represented function.
A \emph{unit} is accordingly a pair
$(s,w)\in\R\times\sphere^{d-1}$ of a \emph{mass} and a \emph{(feature) direction}, the vector $sw\in\R^d$ is its \emph{generator} (\Cref{fig:interlacing}a),
and the same classes are parameterized in mass--direction coordinates by
\begin{align}
 f_{q;s,w}(x)=\sum_{i=1}^{n}s_i\,\varphi_q(\ip{w_i}{x}),\qquad
 (s,w)\in\R^n\times(\sphere^{d-1})^n ,
 \label{eq:models}
\end{align}
and $F_{b,s,w}(x)=\ip{b}{x}+f_{R;s,w}(x)$.
We call parameters that are not restricted to form unit directions the \emph{raw weights}.
On the open set $\mathcal U_{\ne0}(d,n)$ of raw weights
$(o,a)$ with every $a_i\ne0$, the \emph{reparameterization map}
$\Pi(o,a)=(o_i\|a_i\|,\,a_i/\|a_i\|)_i$ satisfies $f_{q;o,a}=f_{q;\Pi(o,a)}$; it is a continuous surjection that removes the positive-scaling redundancy of the raw weights.

We fix a \emph{teacher network} by fixing parameters $(b^{\mathbf t},t,v)\in\R^d\times\R^m\times(\sphere^{d-1})^m$, representing the function $F_{b^{\mathbf t},t,v}\in\Nskip{d}{m}$. Its width $m$ may differ from the width $n$ of the trainable \emph{student network} with parameters $(b^{\mathbf s},s,w)\in\R^d\times\R^n\times(\sphere^{d-1})^n$. The weight tuples $(t_k,v_k)$ and $(s_i,w_i)$, are the \emph{teacher units} and the \emph{student units}.
A unit with $s_i=0$ is a
\emph{zero-mass unit} and its respective hidden neuron contributes nothing to the represented function.  Widths are \emph{underparameterized} ($n<m$), \emph{matched} ($n=m$), or \emph{overparameterized} ($n>m$).

We consider the following parameter regimes for the mass--direction vector pairs $(s,w)$, with the same notation for student and teacher parameters,
\begin{align}
 \Pnn{d}{n}=\{s_i\ge0\ \forall i\},\quad
 \Ppos{d}{n}=\{s_i>0\ \forall i\},\quad
 \Ppl{d}{n}=\{\dim\operatorname{span}\{w_i\}\le2\},
 \label{eq:parameter-classes}
\end{align}
comprising parameters with non-negative masses, with positive masses, and with coplanar directions.

To measure the distance between the teacher network and a student network, we introduce three loss functions, depending on the inclusion of linear skips and the choice of feature.
\begin{definition}[Losses]
\label{def:losses}
For a teacher network $(b^{\mathbf t},t,v)\in\R^d\times\R^m\times(\sphere^{d-1})^m$ and a student parameter $(b^{\mathbf s},s,w)\in\R^d\times\R^n\times(\sphere^{d-1})^n$ the \emph{skip loss} is
\begin{align}
 \mathscr{L}_{\mathrm{skip}}(b^{\mathbf s},s,w)&=\tfrac12\,\E\bigl[(F_{b^{\mathbf s},s,w}(X)-F_{b^{\mathbf t},t,v}(X))^2\bigr],
 \label{eq:losses}
\end{align}
and for a teacher network $(t,v)$ without linear skip and student parameter $(s,w)\in\R^n\times(\sphere^{d-1})^n$ the \emph{$q$-loss} is
\begin{align}
 \mathscr L_q(s,w)&=\tfrac12\,\E\bigl[(f_{q;s,w}(X)-f_{q;t,v}(X))^2\bigr],
 \label{eq:q-losses}
\end{align}
where we suppress the dependence of the loss on the teacher parameters in the notation to keep formulas light. The two instances of the $q$-loss are the \emph{centered loss} $\LC$ and the \emph{plain loss} $\LR$.
\end{definition}
\begin{definition}[Critical points and minima]
\label{def:minima}
Let $L$ be one of the losses above, with parameter space $M=\R^n\times(\sphere^{d-1})^n$
for the $q$-losses and $M=\R^d\times\R^n\times(\sphere^{d-1})^n$ for the skip loss, whose elements we write $\theta=(s,w)$, respectively $\theta=(b^{\mathbf s},s,w)$.
A parameter $\theta\in M$ is a
\emph{critical point} if $\nabla_{s}L=0$, every
spherical gradient
$\nabla_{\sphere,w_i}L=(I_d-w_iw_i^{\top})\nabla_{w_i}L$, i.e.\ the projection of the Euclidean gradient onto the tangent space $T_u\sphere^{d-1}$, vanishes and additionally for the skip loss, $\nabla_{b^{\mathbf s}}L=0$; a \emph{local minimum}
if no parameter of some neighborhood has smaller loss, \emph{strict} if every
other one has strictly larger loss and \emph{flat} otherwise; an \emph{exact fit} if $L(\theta)=0$; and \emph{spurious} if it is
a local minimum but not a global minimum. A local minimum \emph{in} a subset $P\subseteq M$ is a local minimum of $L$ on all of $M$ that lies in $P$, and the landscape of $L$ is \emph{benign} on $P$ if it has no spurious local minimum in $P$.
\end{definition}
For $n\ge m$, by padding teacher parameters with $n-m$ zeros one obtains parameters for an exact-fit student network, so each loss has global minimum $0$ and a local minimum is spurious exactly when its loss is positive.
For $n<m$, a global minimum may have positive loss, and we give an underparameterized example (\Cref{lem:nc-dead-counterexamples}).
We analyze the loss landscape in the mass--direction parameter space $\R^n \times (\sphere^{d-1})^n$. On raw weights the losses are the compositions with the reparameterization map: for $(o,a)\in\mathcal U_{\ne0}(d,n)$ we have $\tfrac12\E[(f_{q;o,a}(X)-f_{q;t,v}(X))^2]=\mathscr L_q(\Pi(o,a))$, and likewise for the skip loss. The following lemma transfers our results to the raw parameter space $\R^n\times(\R^d)^n$.

\begin{lemma}[\leandecl{Theorems.lean}{40}{PaperLeanFormalization.Theorems.lem_raw_transfer}, \leandecl{Theorems.lean}{59}{PaperLeanFormalization.Theorems.lem_raw_transfer_skip}]
\label{lem:raw-transfer}
    Let $(o,a)\in\mathcal U_{\ne0}(d,n)$. Then $(o,a)$ is a critical point (a local minimum, a spurious local minimum) of $\mathscr L_q\circ\Pi$ if and only if $\Pi(o,a)$ is a critical point (a local minimum, a spurious local minimum respectively) of $\mathscr L_q$; the same holds for the skip loss.
\end{lemma}
The \hyperref[pf:lem:raw-transfer]{proof} (\Cref{app:preliminary-proofs}) uses that $\Pi$ has a continuous local right inverse, so neighborhoods and loss values correspond, and applies the chain rule.

\paragraph{How the losses are connected.}
The identity $[z]_+=\tfrac12|z|+\tfrac12z$ splits a skip network into a
centered network plus a linear map,
\begin{align}
 F_{b^{\mathbf s},s,w}=\ip{b^{\mathbf s}+\tau(s,w)}{\,\cdot\,}+f_{C;s,w},
 \ \text{where} \
 \tau(s,w)=\tfrac12\sum_is_iw_i=\tfrac12\,s^\top w .
 \label{eq:skip-centered-split}
\end{align}
\Cref{fig:zonotope-first-harmonic} pictures this split for $d=2$. By central symmetry of the distribution of $X$, the random variables
$f_{C;s,w}(X)$ and $\ip{u}{X}$ are uncorrelated for any $u \in \R^d$: their
defining functions are even and odd, respectively.
Since $\operatorname{Cov}(X)=I_d$, the linear residual contributes
$\tfrac12\E[\ip{u}{X}^2]=\tfrac12\|u\|_2^2$ to the loss.
The \hyperref[pf:lem:equivalence_loss_skip_centered]{proof} of the following Lemma based on these ideas is given in \Cref{app:preliminary-proofs}.

\begin{lemma}[Loss decomposition, \leandecl{Theorems.lean}{83}{PaperLeanFormalization.Theorems.lem_equivalence_loss_skip_centered}]
\label{lem:equivalence_loss_skip_centered}
For every teacher parameter $(b^{\mathbf t},t,v) \in \R^d\times\R^m\times(\sphere^{d-1})^m$ and every student parameter $(b^{\mathbf s},s,w) \in \R^d\times\R^n\times(\sphere^{d-1})^n$ we have
\begin{align}
 \Lskip(b^{\mathbf s},s,w)&=\LC(s,w)+\tfrac12\|b^{\mathbf s}-b_{\rm match}(s,w)\|_2^2\label{eq:skip-connection}\\
 \LR(s,w)&=\LC(s,w)+\tfrac18\|s^\top w- t^\top v \|_2^2 ,
 \label{eq:loss-connection}
\end{align} where
$b_{\rm match}(s,w)=b^{\mathbf t}+\frac{1}{2}(t^\top v - s^\top w)$.
Moreover, $(b^{\mathbf s},s,w)$ is a (strict) local minimum (or critical point) of $\Lskip$ iff $b^{\mathbf s}=b_{\rm match}(s,w)$ and
$(s,w)$ is a (strict) local minimum (or critical point) of $\LC$.
\end{lemma}

\paragraph{From Gaussian loss to one residual.}
The following closed form of the pair moment goes back to \citet{ChoSaul2009} and underlies the computations of \citet{Tian2017} and \citet{SafranShamir2018}.
For unit vectors $\xi,\zeta\in\sphere^{d-1}$ we consider the \emph{pair moment} $M_q(\xi,\zeta)=\E[\varphi_q(\ip{\xi}{X})\varphi_q(\ip{\zeta}{X})]$.
\begin{lemma}[Pair moments,
\leandecl{Theorems.lean}{133}{PaperLeanFormalization.Theorems.lem_pair_moments}]
\label{lem:pair-moments}
Let $\xi,\zeta\in\sphere^{d-1}$ with $\rho=\ip{\xi}{\zeta}$ and $q\in\{C,R\}$. Then,
\begin{align}
 2\pi M_q(\xi,\zeta)=\widetilde\Phi_q(\rho), \ \text{where} \
 \widetilde\Phi_C(\rho)=\rho\arcsin\rho+\sqrt{1-\rho^2}, \ \text{and} \
 \widetilde\Phi_R(\rho)=\widetilde\Phi_C(\rho)+\frac\pi2\rho .
 \label{eq:kernels}
\end{align}
\end{lemma}

The \hyperref[pf:lem:pair-moments]{proof} (\Cref{app:preliminary-proofs}) reduces to the plane of $\xi,\zeta$ and integrates in polar coordinates over the arc on which both ReLUs are active; the centered case follows from the analysis of $\Phi_R$ via $\varphi_R=\varphi_C+z/2$. We call $\widetilde\Phi_C$ and $\widetilde\Phi_R$ the \emph{kernels}; they differ only by the degree-one harmonic $\frac\pi2\rho$ (\Cref{fig:angular-kernels} (c)).

The next proposition expresses the loss and its first derivatives through the \emph{residual} $\mathcal F_q \colon \sphere^{d-1} \to \R$ with student and teacher parameters $(s,w)$ and $(t,v)$ respectively,
\begin{align}
 \mathcal F_q(u)=\sum_i s_i\widetilde\Phi_q(\ip{u}{w_i})
 -\sum_k t_k\widetilde\Phi_q(\ip{u}{v_k}),
 \label{eq:ambient-residual-main}
\end{align}
the \emph{student potential} minus the \emph{teacher potential}. We suppress the dependence of the residual on the student network and the teacher network in the notation for light formulas.

\begin{proposition}[Residual representation of the loss, \leandecl{Theorems.lean}{146}{PaperLeanFormalization.Theorems.prop_loss_in_residual}]
\label{prop:loss_in_residual}
For student and teacher parameters $(s,w)$ and $(t,v)$ respectively, the loss and the directional derivatives of the loss satisfy
    \begin{align}
\mathscr L_q&=\frac{1}{ 4\pi}(\sum_i s_i\mathcal F_q(w_i)-\sum_k t_k\mathcal F_q(v_k)),
\label{eq:residual-loss}
\\
 \partial_{s_i}\mathscr L_q&=\frac{1}{2\pi}\mathcal F_q(w_i),
 \qquad
 \nabla_{\sphere, w_i}\mathscr L_q=\frac{1}{2\pi}s_i\,\nabla_{\sphere, w_i}\mathcal F_q(w_i).
 \label{eq:residual-mass-gradient}
 \end{align}
\end{proposition}
The \hyperref[pf:prop:loss_in_residual]{proof} (\Cref{app:preliminary-proofs}) expands the square with \Cref{lem:pair-moments} into a quadratic form in the masses with the kernels as coefficients and regroups it by units. This gives the loss representation; the derivative representations follow from simple calculations.

\section{Properties of Local Minima}
\label{sec:local-minima}
Two properties of every local minimum with non-negative masses drive the analysis of \Cref{sec:benign}: each student direction of positive mass lies in the span of the teacher directions, and a student unit of zero mass forces an exact fit.

\begin{proposition}[Feature Learning,
\leandecl{Theorems.lean}{189}{PaperLeanFormalization.Theorems.prop_teacher_span_confinement}]
\label{prop:teacher-span-confinement}
Let $d\ge2$. For any teacher parameter $(t,v)$, let
$\mathcal T=\operatorname{span}\{v_k\}$ be the linear span of the teacher
directions. For $q\in\{C,R\}$, let $(s,w) \in \Pnn{d}{n}$ be a local minimum of $\mathscr L_q$.  Then $w_i\in\mathcal T$
for every $i \in [n]$ with $s_i>0$.
\end{proposition}

The
\hyperref[pf:prop:teacher-span-confinement]{proof}
(\Cref{app:local-minima-proofs}) decomposes each student direction
as $w_i=p_i+q_i$, with $p_i\in\mathcal T$ and
$q_i\perp\mathcal T$, and shows that $s_iq_i=0$.
We conclude this by analyzing the variation of $\mathscr L_q$ under two specific motions, that is infinitesimal changes of student generators $s_i w_i$. The
\emph{normal motion}
$z_N=(s_1q_1,\ldots,s_nq_n)$ stretches the normal components,
while the \emph{radial motion}
$z_A=(s_1w_1,\ldots,s_nw_n)$ scales all generators uniformly.
For two motions $z, z'$, write $z^\top Hz'$ for the mixed second derivative of $\mathscr L_q$ in directions $z, z'$.
First, we show that
local minimality of $(s, w)$ implies that the normal motion causes no change in $\mathscr L_q$ up to second order, $z_N^\top Hz_N=0$.
Consequently, Cauchy--Schwarz yields $z_A^\top Hz_N=0$.
Second, we
interpret
this mixed
derivative
in terms of variation of student's second moment under the normal motion,
$z_A^\top Hz_N
=\frac12\,\frac{d}{d\tau}\big|_{\tau=0}
\E[f_{z_A+\tau z_N}(X)^2]$,
where $f_V$ denotes the student output with generators $V$.
An argument using convexity of this second moment in the
generators shows that the mixed
derivative $z_A^\top Hz_N$
would be positive
if some $s_iq_i\ne0$, contradicting the first step.

A zero-mass unit also contradicts being a spurious local minimum.
\begin{proposition}[Zero student mass implies exact fit,
\leandecl{Theorems.lean}{199}{PaperLeanFormalization.Theorems.prop_zero_mass_exact_fit},
\leandecl{Theorems.lean}{211}{PaperLeanFormalization.Theorems.prop_nc_dead_student}]
\label{prop:zero_mass_exact_fit}
\label{prop:nc-dead-student}
Let $d\ge1$ and assume only that the teacher masses are non-negative when $d=2$.
If $(s,w)$ with unrestricted masses is a local minimum of $\LC$ (for $d\ge1$) or $\LR$ (for $d\ge3$) with a zero-mass unit, then it is an exact fit.
\end{proposition}

The \hyperref[pf:prop:zero_mass_exact_fit]{proof} (\Cref{app:local-minima-proofs}) uses the representation of the loss via the residual in
Proposition~7: to show an exact fit, it suffices to show that the residual
vanishes everywhere. Analyzing how the loss changes when changing the direction
and mass of the zero-mass unit $(s_i,w_i)$,
using \cref{eq:ambient-residual-main}, we show that
there would be a descent direction if the residual would not vanish on an open
set around $w_i$. We prove that the residual is real analytic away from finitely
many breakpoints at the student and teacher directions and their reflections
w.r.t.\ the origin. Depending on the dimension, different techniques propagate
this zero set of the residual to prove the residual vanishes everywhere, with
the same proof for the plain and centered residual in case $d\ge3$.

Surprisingly, the $d=2$ non-negative-teacher hypothesis cannot be dropped, as
the example in Lemma~29(b) demonstrates.

\section{Benign Loss Landscape for Coplanar Teacher Networks}
\label{sec:benign}
This section presents the ideas behind our main result and the role of the residual \cref{eq:ambient-residual-main} in our analysis.

\begin{theorem}[Main Result,
\leandecl{Theorems.lean}{410}{PaperLeanFormalization.Theorems.thm_headline}]
\label{thm:headline}
Let $d\ge1$, let the teacher network satisfy $(t,v)\in\Ppos{d}{m}\cap\Ppl{d}{m}$ with any linear skip $b^{\mathbf t}\in\R^d$, and let $n\ge m$. Then every local minimum $(b^{\mathbf s},s,w)$ of $\Lskip$ on $\R^d\times\R^n\times(\sphere^{d-1})^n$ with $(s,w)\in\Pnn{d}{n}$ is an exact fit.
\end{theorem}
The \hyperref[pf:thm:headline]{proof} first reduces the statement to the centered model: the skip loss is the centered loss plus a term in the linear skip alone (\Cref{lem:equivalence_loss_skip_centered}), so a spurious local minimum with non-negative masses of the skip model yields one of the centered model, and the main result follows from its centered formulation, \Cref{thm:centered} in \Cref{app:benign-proofs}. The case $d=1$ is a direct argument in the masses. Let $d\ge2$ and let $(s,w)$ be a local minimum of $\LC$. A zero student mass gives an exact fit (\Cref{prop:zero_mass_exact_fit}); otherwise feature learning (\Cref{prop:teacher-span-confinement}) places student units and teacher units in a common plane $P$, and since centered features of directions in $P$ see only the projection of $X$ onto $P$, $(s,w)$ is a local minimum of the planar loss with the same value. For convenience let us identify $P = \R^2$.

\begin{figure}[t]
 \centering
 \edef\unitfigureheight{\the\dimexpr.27\linewidth\relax}
 \begin{minipage}[b]{.36\linewidth}
  \centering
  \includegraphics[height=\unitfigureheight]{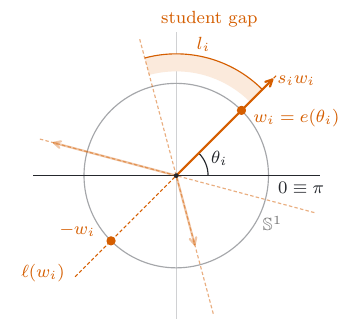}
  \par\small (a) Mass, direction, and line
 \end{minipage}\hfill
 \begin{minipage}[b]{.63\linewidth}
  \centering
  \includegraphics[height=\unitfigureheight]{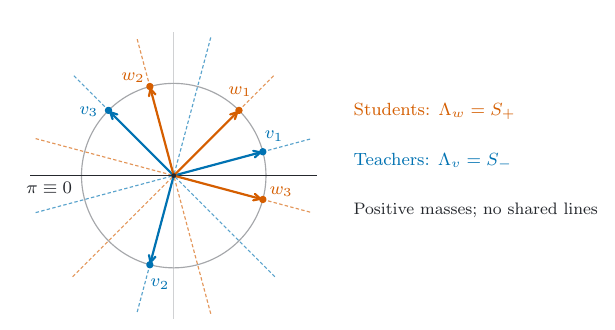}
  \par\small (b) Strict interlacing
 \end{minipage}
 \caption{Unit notation and strict interlacing in the centered model.
 (a)~A student unit has mass $s_i$, direction $w_i=e(\theta_i)$ and generator
 $s_iw_i$; the antipodes $\pm w_i$ determine one line $\ell(w_i)\in\R/\pi\Z$,
 and the shaded arc marks a student gap of length $l_i$.
 (b)~Every gap between student lines (orange) contains exactly one teacher
 line (blue); with positive masses and distinct lines the student lines are
 $S_+$ and the teacher lines $S_-$. The configuration illustrates the order
 condition of interlacing, not stationarity.}
 \label{fig:interlacing}
\end{figure}

In the plane, write a direction as $e(\gamma)=(\cos\gamma,\sin\gamma)$, with student and teacher angles $\theta_i,\beta_k$. The centered feature $|\ip{u}{x}|$ does not distinguish $u$ from $-u$, so only the \emph{line} $\ell(u)=\gamma\bmod\pi\in\R/\pi\Z$ of $u=e(\gamma)$ matters (\Cref{fig:interlacing}a); collecting the units on a line $\alpha$ gives its \emph{net mass} and the lines of positive and negative net mass,
\begin{align}
 \omega(\alpha)=\sum_{\ell(w_i)=\alpha}s_i-\sum_{\ell(v_k)=\alpha}t_k, \qquad S_\pm=\{\alpha\in\R/\pi\Z:\pm\omega(\alpha)>0\}.
 \label{eq:net-weight-main}
\end{align}

A spurious local minimum with positive masses is excluded in two steps. First,
we demonstrate (\Cref{lem:line-count}, \Cref{prop:locamin_are_line_separated}) that
its lines of positive and negative net mass alternate around the circle: every \emph{gap} of $S_+$, a connected component of $(\R/\pi\Z)\setminus S_+$, contains exactly one point of $S_-$; we call this \emph{interlacing} (\Cref{fig:interlacing}b); moreover $S_+$ is the set of the $n$ student lines and $S_-$ that of the $m=n$ teacher lines, all distinct (\Cref{prop:locamin_are_line_separated}), so for $n>m$ the first step already closes the theorem by contradiction, while the following second step is only needed for the matched case. At an interlacing critical point we construct a direction of strictly negative second variation of the centered loss (\Cref{prop:beam-descent}), so it is not a local minimum. Both steps rest on the residual of \Cref{prop:loss_in_residual} read as a function of the angle, and together they prove \Cref{thm:centered} and with it the main result.

\begin{figure}[t] \centering
 \begin{minipage}[t]{.445506\linewidth}
  \vspace{0pt}
  \centering
  \includegraphics[width=\linewidth]{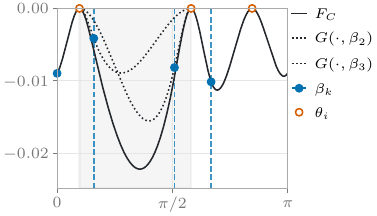}
  \par\small (a) Centered angular residual $\mathcal F_C(\gamma)$
 \end{minipage}\hfill
 \begin{minipage}[t]{.256762\linewidth}
  \vspace{0pt}
  \centering
  \includegraphics[width=\linewidth]{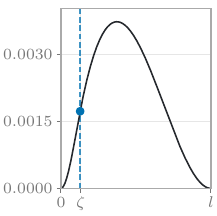}
  \par\small (b) Green function $G_l(\cdot;\zeta)$
 \end{minipage}\hfill
 \begin{minipage}[t]{.287732\linewidth}
  \vspace{0pt}
  \centering
  \includegraphics[width=\linewidth]{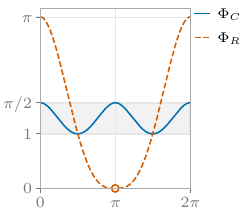}
  \par\small (c) Angular kernels $\Phi_q(\tau)$
 \end{minipage}
 \caption{Two representations of the residual: kernel sums and Green-function
 solutions (\Cref{lem:gap-sign}).
 (a)~Centered angular residual \eqref{eq:residual-potential-paper} for three student units at numerical double zeros (orange
 markers; uncertified Newton approximation) and the teacher units
 $\beta=(0,0.5,1.6,2.1)$, $t=(1,0.6,0.8,1.2)$ (blue markers). On the shaded
 student gap, which contains two teacher units, the dotted Green functions
 $-2.4G_1$, $-3.2G_2$ of the two loads sum to the solid residual as in
 \eqref{eq:gap-green-representation}.
 (b)~The Green function $G(\,\cdot\,;\zeta)$: non-negative, with zero value and
 slope at both ends and a unit third-derivative jump at $\zeta$.
 (c)~The angular kernels $\Phi_q(\tau)=\widetilde\Phi_q(\cos\tau)$ of \eqref{eq:angular-kernel}; $\Phi_C$ is
 $\pi$-periodic, $\Phi_R$ is $2\pi$-periodic and vanishes at $\tau=\pi$.}
 \label{fig:residual-green}
 \label{fig:angular-kernels}
\end{figure}

\emph{The residual as a bent ring.}
The residual \eqref{eq:ambient-residual-main} is a weighted sum of shifted copies of the kernel.
We observe that between student and teacher lines, it solves a certain ODE (\Cref{lem:gap-sign}), whose analysis is the key to our main result. Surprisingly, the same formulas were studied in the physical discipline of elasticity: the residual follows the shape of a ring bent by one pair of opposite forces per line, of strength the net mass (\Cref{rem:ring}; compare to \cite{Love1927}).
The response to a single unit load at $\zeta$ is the Green function $G(\,\cdot\,;\zeta)$ (\Cref{fig:residual-green}b), and the centered residual is also a sum of Green functions between the lines of $S_+$ (\Cref{fig:residual-green}a), from which the interlacing \Cref{lem:line-count} follows combinatorially.

\section{Spurious local minima for ReLU}

In this section, we construct a spurious local minimum for plain networks with positive second-layer weights that persists under arbitrary overparameterization.

\label{sec:plain}
\begin{theorem}[Plain spurious local minima at arbitrary overparameterization, \leandecl{Theorems.lean}{504}{PaperLeanFormalization.Theorems.thm_plain_trap}] \label{thm:plain-trap}
  Let $d=2$ and take the teacher network with unit masses at the angles $(0,5\pi/6,4\pi/3)$. For every $n\ge3$ the plain loss $\LR$ has a spurious local minimum with all student masses positive. For $n=3$, the local minimum is strict and for $n>3$ it is flat.
\end{theorem}

Spurious local minima of this loss were first established by \citet{SafranShamir2018}, for orthonormal teacher networks of width at least six, by a computer-assisted certificate around a candidate proposed by gradient descent: a small gradient, a Hessian floor, and a bound on the third derivative confine a local minimum to a small ball. We follow this paradigm with three changes (\hyperref[pf:thm:plain-trap]{full proof ideas} in \Cref{app:plain-proofs}). Strong convexity on a whole box around a rational candidate, together with a small gradient at its center, yields a unique strict local minimum inside the box (\Cref{lem:box-certificate}). Positive loss follows from a conceptual argument analyzing the kinks of the student and teacher networks. This yields the strict spurious local minimum for $n=3$. At it, the third student unit has positive residual curvature, $\mathcal F_R''(\theta_3)>0$, and positive residual curvature lets a student unit be split into several units with the same direction on which the mass is distributed without losing local minimality (\Cref{lem:positive-curvature-splitting}).

The construction achieves persistence through collisions of student features. Are such collisions necessary at large student width? \Cref{prop:plain-collision-ceiling} bounds the number of distinct student directions at critical points.
\begin{proposition}[Effective width at overparameterization, \leandecl{Theorems.lean}{557}{PaperLeanFormalization.Theorems.prop_plain_collision_ceiling}]\label{prop:plain-collision-ceiling}
  Let $d=2$. A critical point of $\LR$ with all masses positive with respect to a teacher of width $m$ and unrestricted masses has at most $2m$ pairwise distinct student directions, and for every odd $m\ge3$ this bound is attained.
\end{proposition}
The \hyperref[pf:prop:plain-collision-ceiling]{proof} (\Cref{app:plain-proofs}) studies the sign pattern of $\mathcal W = \tfrac12(\mathcal F_R''+\mathcal F_R)$, a piecewise-sinusoid function
with slope jumps at
the student and teacher angles.
We assign each student direction an interval on which $\mathcal W > 0$ and a combinatorial argument gives the desired bound. The bound is sharp for every odd $m\ge3$, as witnessed by an explicit critical point with $2m$ distinct student directions.

\section{Finite sample consequence}
\label{sec:boundaries}

We leverage the main result \Cref{thm:headline} in the empirical loss setting of a finite dataset $\mathcal D=\{(x_j,y_j)\}_{j=1}^N$ with inputs $x_j\in\R^d$ and \emph{teacher-generated} labels $y_j=F_{b^{\mathbf t},t,v}(x_j)$, where $F_{b,s,w}(x)=\ip{b}{x}+f_{R;s,w}(x)$ is the skip network: every empirical local minimum that persists on a parameter ball of fixed radius has zero empirical loss once the inputs approximate the Gaussian distribution well enough. The empirical loss is the unweighted sample average
\begin{align}
 \widehat L_{\mathcal D}(b^{\mathbf s},s,w)=\frac1{2N}\sum_{j=1}^N\bigl[F_{b^{\mathbf s},s,w}(x_j)-y_j\bigr]^2 ,
 \label{eq:empirical-loss}
\end{align}
and the approximation is measured on the inputs alone, through their average squared radius $R_2(\mathcal D)$: the dataset has \emph{accuracy $\delta$} if for every continuous $H\colon\R^d\to\R$ with $H(\tau x)=\tau^2H(x)$ for $\tau\ge0$ and every Lipschitz constant $L$ of $H$ on the closed unit ball, with $G\sim\mathcal N(0,I_d)$,
\begin{align}
 R_2(\mathcal D)=\tfrac1N\textstyle\sum_{j}\|x_j\|^2,
 \qquad
 \Bigl|\tfrac dN\textstyle\sum_{j}H(x_j)-R_2(\mathcal D)\,\E H(G)\Bigr|\le d\,R_2(\mathcal D)\,L\delta .
 \label{eq:modulus}
\end{align}
For independent, identically distributed inputs from an orthogonally invariant distribution with finite positive second moment, accuracy $\delta$ holds with high probability for all sufficiently large $N$ (\Cref{rem:finite-data-accuracy} in \Cref{app:transfer-proofs}). Minima are compared on balls of a fixed radius: let $d_J(p,p')$ be the maximum of the distances of the masses, of the directions, and of the linear skips measured from the matching skip $b_{\rm match}$ of \Cref{lem:equivalence_loss_skip_centered} (\eqref{eq:max-product-distance} in \Cref{app:transfer-proofs}); a local minimum $p$ is \emph{$r$-stable} if it minimizes $\widehat L_{\mathcal D}$ on the closed $d_J$-ball of radius $r$ about itself.

\begin{theorem}[Fixed-radius stability transfer, \leandecl{Theorems.lean}{593}{PaperLeanFormalization.Theorems.thm_fixed_radius_bridge}]
\label{thm:fixed-radius-bridge}
Let $d\ge1$, let the teacher network be positive and coplanar, and let the student width satisfy $n\ge m$, as in \Cref{thm:headline}. For every $r>0$ there is a $\delta>0$ such that, for every finite dataset with teacher-generated labels and accuracy $\delta$, every $r$-stable local minimum with non-negative masses has zero empirical loss. For $d=1$ the conclusion holds for every finite dataset, with no accuracy requirement.
\end{theorem}

We emphasize that this result only concerns $r$-stable minima and is therefore an application, not a one-to-one transfer of the main result to the empirical loss.
The \hyperref[pf:thm:fixed-radius-bridge]{proof ideas} (\Cref{app:transfer-proofs}) handle $d=1$ by a direct scaling argument, while for $d\ge2$ we argue by contradiction. Suppose there are datasets $\mathcal D_j$ of accuracy $1/j$ and $r$-stable minima $p_j$ of positive empirical loss. The minimizer stability theorem of \citet{BastinCirilloToint2006}, \Cref{thm:minimizer-stability}, states that cluster points of fixed-radius minima under uniform convergence are local minima of the limit. A scaling identity and the accuracy bound place all $p_j$ in a common compact set $K$, while on its fixed-radius enlargement the rescaled empirical losses converge uniformly to $\Lskip$. Hence a cluster point $p$ is a population local minimum and thus an exact fit by \Cref{thm:headline}. The fit is pointwise, so $\widehat L_{\mathcal D_j}(p)=0$ for every $j$, while for large $j$, $p$ lies in the $r$-ball about $p_j$, contradicting $0<\widehat L_{\mathcal D_j}(p_j)\le\widehat L_{\mathcal D_j}(p)=0$.

\section{Conclusion and Limitations}
\label{sec:conclusion}
We expect the planarity assumption to be removable: it enters only through the reduction to the circle in \Cref{sec:benign}, while confinement (\Cref{prop:teacher-span-confinement}) and the zero-mass argument (\Cref{prop:zero_mass_exact_fit}) hold in every dimension.
\begin{conjecture}
\label{conj:higher-rank}
Let $d\ge1$, let the teacher network satisfy $(t,v)\in\Ppos{d}{m}$ with any linear skip $b^{\mathbf t}\in\R^d$ and let $n\ge m$. Then every local minimum $(b^{\mathbf s},s,w)$ of $\Lskip$ on $\R^d\times\R^n\times(\sphere^{d-1})^n$ with $(s,w)\in\Pnn{d}{n}$ is an exact fit.
\end{conjecture}
What is missing is a replacement for the angular residual analysis: the interlacing and descent arguments
rely heavily on
the one-dimensional order of lines on the circle,
and no way to extend these techniques to higher dimensions
is known to us.
Like most of the loss landscape theory it builds on, our analysis is confined to one hidden layer: the residual representation of \Cref{prop:loss_in_residual} rests on the closed form of the pair moments, which has no counterpart for deeper networks, so the effect of a learned skip on the landscape of deep networks remains open.
Moreover, our result on the empirical loss landscape concerns minima that persist on a prescribed parameter ball. Whether one can rule out ordinary positive-loss empirical local minima, without a fixed-radius stability assumption, remains open. Understanding when population-landscape benignity transfers quantitatively to finite samples is an important direction for future work.

Our results are landscape statements and do not imply convergence of a particular optimization algorithm. A natural next question is whether our insights into the geometry of the loss landscape can be used to analyze gradient-based dynamics.

\clearpage

\paragraph{Acknowledgements.}
We gratefully acknowledge Iliad for supporting this project with API credits and subscription support. Jakob Paul Zimmermann thanks Wojciech Samek for his support and for providing the freedom and flexibility to pursue this research.

\bibliographystyle{iclr2027_conference}

\bibliography{references}

\begin{thebibliography}{38}
\providecommand{\natexlab}[1]{#1}
\providecommand{\url}[1]{\texttt{#1}}
\expandafter\ifx\csname urlstyle\endcsname\relax
  \providecommand{\doi}[1]{doi: #1}\else
  \providecommand{\doi}{doi: \begingroup \urlstyle{rm}\Url}\fi

\bibitem[Arjevani \& Field(2021)Arjevani and Field]{ArjevaniField2021}
Yossi Arjevani and Michael Field.
\newblock Analytic study of families of spurious minima in two-layer {ReLU}
  neural networks: A tale of symmetry {II}.
\newblock In \emph{Advances in Neural Information Processing Systems},
  volume~34, 2021.
\newblock URL
  \url{https://proceedings.neurips.cc/paper/2021/hash/806d926414ce19d907700e23177ab4ff-Abstract.html}.

\bibitem[Arjevani \& Field(2022)Arjevani and Field]{ArjevaniField2022}
Yossi Arjevani and Michael Field.
\newblock Annihilation of spurious minima in two-layer {ReLU} networks.
\newblock In \emph{Advances in Neural Information Processing Systems},
  volume~35, 2022.
\newblock URL
  \url{https://proceedings.neurips.cc/paper_files/paper/2022/hash/f3da4165893c2465fd7e8df453c41ffa-Abstract-Conference.html}.

\bibitem[Avigad(2026)]{Avigad2026}
Jeremy Avigad.
\newblock Mathematicians in the age of {AI}.
\newblock arXiv preprint arXiv:2603.03684, 2026.
\newblock URL \url{https://arxiv.org/abs/2603.03684}.

\bibitem[Bastin et~al.(2006)Bastin, Cirillo, and Toint]{BastinCirilloToint2006}
Fabian Bastin, Cinzia Cirillo, and Philippe~L. Toint.
\newblock Convergence theory for nonconvex stochastic programming with an
  application to mixed logit.
\newblock \emph{Mathematical Programming}, 108\penalty0 (2--3), 2006.
\newblock \doi{10.1007/s10107-006-0708-6}.

\bibitem[Cho \& Saul(2009)Cho and Saul]{ChoSaul2009}
Youngmin Cho and Lawrence~K. Saul.
\newblock Kernel methods for deep learning.
\newblock In \emph{Advances in Neural Information Processing Systems},
  volume~22, 2009.
\newblock URL
  \url{https://proceedings.neurips.cc/paper/2009/hash/5751ec3e9a4feab575962e78e006250d-Abstract.html}.

\bibitem[Commelin et~al.(2026)Commelin, Jamnik, Ochigame, Taelman, and
  Venkatesh]{Commelin2026}
Johan Commelin, Mateja Jamnik, Rodrigo Ochigame, Lenny Taelman, and Akshay
  Venkatesh.
\newblock Shaping the future of mathematics in the age of {AI}.
\newblock \emph{Notices of the American Mathematical Society}, 73\penalty0 (6),
  2026.
\newblock URL \url{https://arxiv.org/abs/2603.24914}.

\bibitem[Damian et~al.(2022)Damian, Lee, and
  Soltanolkotabi]{DamianLeeSoltanolkotabi2022}
Alexandru Damian, Jason~D. Lee, and Mahdi Soltanolkotabi.
\newblock Neural networks can learn representations with gradient descent.
\newblock In \emph{Proceedings of the 35th Conference on Learning Theory},
  volume 178 of \emph{Proceedings of Machine Learning Research}, 2022.
\newblock URL \url{https://proceedings.mlr.press/v178/damian22a.html}.

\bibitem[de~Moura \& Ullrich(2021)de~Moura and Ullrich]{deMouraUllrich2021}
Leonardo de~Moura and Sebastian Ullrich.
\newblock The {Lean} 4 theorem prover and programming language.
\newblock In \emph{Automated Deduction -- CADE 28}, volume 12699 of
  \emph{Lecture Notes in Computer Science}. Springer, 2021.
\newblock \doi{10.1007/978-3-030-79876-5_37}.

\bibitem[Du et~al.(2019)Du, Zhai, P{\'o}czos, and Singh]{DuZhaiPoczosSingh2019}
Simon~S. Du, Xiyu Zhai, Barnab{\'a}s P{\'o}czos, and Aarti Singh.
\newblock Gradient descent provably optimizes over-parameterized neural
  networks.
\newblock In \emph{International Conference on Learning Representations}, 2019.
\newblock URL \url{https://openreview.net/forum?id=S1eK3i09YQ}.

\bibitem[Foster et~al.(2018)Foster, Sekhari, and
  Sridharan]{FosterSekhariSridharan2018}
Dylan~J. Foster, Ayush Sekhari, and Karthik Sridharan.
\newblock Uniform convergence of gradients for non-convex learning and
  optimization.
\newblock In \emph{Advances in Neural Information Processing Systems},
  volume~31, 2018.
\newblock URL
  \url{https://proceedings.neurips.cc/paper/2018/hash/59ab3ba90ae4b4ab84fe69de7b8e3f5f-Abstract.html}.

\bibitem[Fukumizu \& Amari(2000)Fukumizu and Amari]{FukumizuAmari2000}
Kenji Fukumizu and Shun-ichi Amari.
\newblock Local minima and plateaus in hierarchical structures of multilayer
  perceptrons.
\newblock \emph{Neural Networks}, 13\penalty0 (3), 2000.
\newblock \doi{10.1016/S0893-6080(00)00009-5}.

\bibitem[Fukumizu et~al.(2019)Fukumizu, Yamaguchi, Mototake, and
  Tanaka]{FukumizuYamaguchiMototakeTanaka2019}
Kenji Fukumizu, Shoichiro Yamaguchi, Yoh-ichi Mototake, and Mirai Tanaka.
\newblock Semi-flat minima and saddle points by embedding neural networks to
  overparameterization.
\newblock In \emph{Advances in Neural Information Processing Systems},
  volume~32, 2019.
\newblock URL
  \url{https://proceedings.neurips.cc/paper/2019/hash/a4ee59dd868ba016ed2de90d330acb6a-Abstract.html}.

\bibitem[He et~al.(2016{\natexlab{a}})He, Zhang, Ren, and
  Sun]{HeZhangRenSun2016}
Kaiming He, Xiangyu Zhang, Shaoqing Ren, and Jian Sun.
\newblock Deep residual learning for image recognition.
\newblock In \emph{Proceedings of the IEEE Conference on Computer Vision and
  Pattern Recognition}, 2016{\natexlab{a}}.
\newblock \doi{10.1109/CVPR.2016.90}.

\bibitem[He et~al.(2016{\natexlab{b}})He, Zhang, Ren, and
  Sun]{HeZhangRenSunIdentity2016}
Kaiming He, Xiangyu Zhang, Shaoqing Ren, and Jian Sun.
\newblock Identity mappings in deep residual networks.
\newblock In \emph{Computer Vision -- ECCV 2016}, volume 9908 of \emph{Lecture
  Notes in Computer Science}. Springer, 2016{\natexlab{b}}.
\newblock \doi{10.1007/978-3-319-46493-0_38}.

\bibitem[Kawaguchi \& Bengio(2019)Kawaguchi and Bengio]{KawaguchiBengio2019}
Kenji Kawaguchi and Yoshua Bengio.
\newblock Depth with nonlinearity creates no bad local minima in {ResNets}.
\newblock \emph{Neural Networks}, 118, 2019.
\newblock \doi{10.1016/j.neunet.2019.06.009}.

\bibitem[Kim et~al.(2025)Kim, Mishkin, and Pilanci]{KimMishkinPilanci2025}
Sungyoon Kim, Aaron Mishkin, and Mert Pilanci.
\newblock Exploring the loss landscape of regularized neural networks via
  convex duality.
\newblock In \emph{International Conference on Learning Representations}, 2025.
\newblock URL \url{https://openreview.net/forum?id=4xWQS2z77v}.

\bibitem[Li et~al.(2018)Li, Xu, Taylor, Studer, and Goldstein]{LiEtAl2018}
Hao Li, Zheng Xu, Gavin Taylor, Christoph Studer, and Tom Goldstein.
\newblock Visualizing the loss landscape of neural nets.
\newblock In \emph{Advances in Neural Information Processing Systems},
  volume~31, 2018.
\newblock URL
  \url{https://proceedings.neurips.cc/paper_files/paper/2018/hash/a41b3bb3e6b050b6c9067c67f663b915-Abstract.html}.

\bibitem[Li \& Yuan(2017)Li and Yuan]{LiYuan2017}
Yuanzhi Li and Yang Yuan.
\newblock Convergence analysis of two-layer neural networks with {ReLU}
  activation.
\newblock In \emph{Advances in Neural Information Processing Systems},
  volume~30, 2017.
\newblock URL
  \url{https://papers.nips.cc/paper/2017/hash/a96b65a721e561e1e3de768ac819ffbb-Abstract.html}.

\bibitem[Liang et~al.(2018)Liang, Sun, Lee, and
  Srikant]{LiangSunLeeSrikant2018}
Shiyu Liang, Ruoyu Sun, Jason~D. Lee, and R.~Srikant.
\newblock Adding one neuron can eliminate all bad local minima.
\newblock In \emph{Advances in Neural Information Processing Systems},
  volume~31, 2018.
\newblock URL
  \url{https://proceedings.neurips.cc/paper/2018/hash/a012869311d64a44b5a0d567cd20de04-Abstract.html}.

\bibitem[Love(1927)]{Love1927}
A.~E.~H. Love.
\newblock \emph{A Treatise on the Mathematical Theory of Elasticity}.
\newblock Cambridge University Press, fourth edition, 1927.
\newblock URL \url{https://archive.org/details/in.ernet.dli.2015.462644}.
\newblock Unabridged Dover reprint, 1944.

\bibitem[Mei et~al.(2018)Mei, Bai, and Montanari]{MeiBaiMontanari2018}
Song Mei, Yu~Bai, and Andrea Montanari.
\newblock The landscape of empirical risk for nonconvex losses.
\newblock \emph{The Annals of Statistics}, 46\penalty0 (6A), 2018.
\newblock \doi{10.1214/17-AOS1637}.

\bibitem[Mousavi-Hosseini et~al.(2023)Mousavi-Hosseini, Park, Girotti,
  Mitliagkas, and Erdogdu]{MousaviHosseiniEtAl2023}
Alireza Mousavi-Hosseini, Sejun Park, Manuela Girotti, Ioannis Mitliagkas, and
  Murat~A. Erdogdu.
\newblock Neural networks efficiently learn low-dimensional representations
  with {SGD}.
\newblock In \emph{International Conference on Learning Representations}, 2023.
\newblock URL \url{https://openreview.net/forum?id=6taykzqcPD}.

\bibitem[Pilanci \& Ergen(2020)Pilanci and Ergen]{PilanciErgen2020}
Mert Pilanci and Tolga Ergen.
\newblock Neural networks are convex regularizers: Exact polynomial-time convex
  optimization formulations for two-layer networks.
\newblock In \emph{Proceedings of the 37th International Conference on Machine
  Learning}, volume 119 of \emph{Proceedings of Machine Learning Research},
  2020.
\newblock URL \url{https://proceedings.mlr.press/v119/pilanci20a.html}.

\bibitem[Polak(1993)]{Polak1993}
Elijah Polak.
\newblock On the use of consistent approximations in the solution of
  semi-infinite optimization and optimal control problems.
\newblock \emph{Mathematical Programming}, 62, 1993.
\newblock \doi{10.1007/BF01585175}.

\bibitem[Safran \& Shamir(2018)Safran and Shamir]{SafranShamir2018}
Itay Safran and Ohad Shamir.
\newblock Spurious local minima are common in two-layer {ReLU} neural networks.
\newblock In \emph{Proceedings of the 35th International Conference on Machine
  Learning}, volume~80 of \emph{Proceedings of Machine Learning Research},
  2018.
\newblock URL \url{https://proceedings.mlr.press/v80/safran18a.html}.

\bibitem[Safran et~al.(2021)Safran, Yehudai, and
  Shamir]{SafranYehudaiShamir2021}
Itay Safran, Gilad Yehudai, and Ohad Shamir.
\newblock The effects of mild over-parameterization on the optimization
  landscape of shallow {ReLU} neural networks.
\newblock In \emph{Proceedings of the 34th Conference on Learning Theory},
  volume 134 of \emph{Proceedings of Machine Learning Research}, 2021.
\newblock URL \url{https://proceedings.mlr.press/v134/safran21a.html}.

\bibitem[Shamir(2018)]{Shamir2018}
Ohad Shamir.
\newblock Are {ResNets} provably better than linear predictors?
\newblock In \emph{Advances in Neural Information Processing Systems},
  volume~31, 2018.
\newblock URL
  \url{https://proceedings.neurips.cc/paper/2018/hash/26e359e83860db1d11b6acca57d8ea88-Abstract.html}.

\bibitem[{\c S}im{\c s}ek et~al.(2021){\c S}im{\c s}ek, Ged, Jacot, Spadaro,
  Hongler, Gerstner, and Brea]{SimsekEtAl2021}
Berfin {\c S}im{\c s}ek, Fran{\c c}ois Ged, Arthur Jacot, Francesco Spadaro,
  Cl{\'e}ment Hongler, Wulfram Gerstner, and Johanni Brea.
\newblock Geometry of the loss landscape in overparameterized neural networks:
  Symmetries and invariances.
\newblock In \emph{Proceedings of the 38th International Conference on Machine
  Learning}, volume 139 of \emph{Proceedings of Machine Learning Research},
  2021.
\newblock URL \url{https://proceedings.mlr.press/v139/simsek21a.html}.

\bibitem[Tao(2026)]{Tao2026ICM}
Terence Tao.
\newblock Mathematics in the age of {AI}.
\newblock arXiv preprint arXiv:2608.16753, 2026.
\newblock URL \url{https://arxiv.org/abs/2608.16753}.
\newblock Submitted to the Proceedings of the International Congress of
  Mathematicians 2026.

\bibitem[{The mathlib Community}(2020)]{mathlib2020}
{The mathlib Community}.
\newblock The {Lean} mathematical library.
\newblock In \emph{Proceedings of the 9th ACM SIGPLAN International Conference
  on Certified Programs and Proofs}, 2020.
\newblock \doi{10.1145/3372885.3373824}.

\bibitem[Tian(2017)]{Tian2017}
Yuandong Tian.
\newblock An analytical formula of population gradient for two-layered {ReLU}
  network and its applications in convergence and critical point analysis.
\newblock In \emph{Proceedings of the 34th International Conference on Machine
  Learning}, volume~70 of \emph{Proceedings of Machine Learning Research},
  2017.
\newblock URL \url{https://proceedings.mlr.press/v70/tian17a.html}.

\bibitem[Tinaz et~al.(2026)Tinaz, Xie, and
  Soltanolkotabi]{TinazXieSoltanolkotabi2026}
Berk Tinaz, Changzhi Xie, and Mahdi Soltanolkotabi.
\newblock When both layers learn: Training dynamics of representing linear
  models via {ReLU} networks.
\newblock In \emph{Proceedings of the 39th Conference on Learning Theory},
  volume 336 of \emph{Proceedings of Machine Learning Research}, 2026.
\newblock URL \url{https://proceedings.mlr.press/v336/tinaz26a.html}.

\bibitem[van~der Vaart(1998)]{vanderVaart1998}
Aad~W. van~der Vaart.
\newblock \emph{Asymptotic Statistics}, volume~3 of \emph{Cambridge Series in
  Statistical and Probabilistic Mathematics}.
\newblock Cambridge University Press, 1998.
\newblock \doi{10.1017/CBO9780511802256}.

\bibitem[Venturi et~al.(2019)Venturi, Bandeira, and
  Bruna]{VenturiBandeiraBruna2019}
Luca Venturi, Afonso~S. Bandeira, and Joan Bruna.
\newblock Spurious valleys in one-hidden-layer neural network optimization
  landscapes.
\newblock \emph{Journal of Machine Learning Research}, 20\penalty0 (133), 2019.
\newblock URL \url{https://www.jmlr.org/papers/v20/18-674.html}.

\bibitem[Wang et~al.(2022)Wang, Lacotte, and Pilanci]{WangLacottePilanci2022}
Yifei Wang, Jonathan Lacotte, and Mert Pilanci.
\newblock The hidden convex optimization landscape of regularized two-layer
  {ReLU} networks: An exact characterization of optimal solutions.
\newblock In \emph{International Conference on Learning Representations}, 2022.
\newblock URL \url{https://openreview.net/forum?id=Z7Lk2cQEG8a}.

\bibitem[Wu et~al.(2025)Wu, {\c S}im{\c s}ek, and Ged]{WuSimsekGed2025}
Frank~Zhengqing Wu, Berfin {\c S}im{\c s}ek, and Fran{\c c}ois Ged.
\newblock Loss landscape of shallow {ReLU}-like neural networks: Stationary
  points, saddle escape, and network embedding.
\newblock In \emph{International Conference on Learning Representations}, 2025.
\newblock URL \url{https://openreview.net/forum?id=ogKE7LcvW6}.

\bibitem[Yun et~al.(2019)Yun, Sra, and Jadbabaie]{YunSraJadbabaieResNet2019}
Chulhee Yun, Suvrit Sra, and Ali Jadbabaie.
\newblock Are deep {ResNets} provably better than linear predictors?
\newblock In \emph{Advances in Neural Information Processing Systems},
  volume~32, 2019.
\newblock URL
  \url{https://proceedings.neurips.cc/paper/2019/hash/661c1c090ff5831a647202397c61d73c-Abstract.html}.

\bibitem[Zhou et~al.(2021)Zhou, Ge, and Jin]{ZhouGeJin2021}
Mo~Zhou, Rong Ge, and Chi Jin.
\newblock A local convergence theory for mildly over-parameterized two-layer
  neural network.
\newblock In \emph{Proceedings of the 34th Conference on Learning Theory},
  volume 134 of \emph{Proceedings of Machine Learning Research}, 2021.
\newblock URL \url{https://proceedings.mlr.press/v134/zhou21b.html}.

\end{thebibliography}

\clearpage
\appendix

\section{Proofs}

In this appendix we give the proof ideas and underlying theory of the results and formulas presented in our main part. The following subsections reflect the structure of the main paper:
\begin{itemize}
 \item \Cref{app:preliminary-proofs} (\emph{Auxiliary Lemmas}) proves the results of \Cref{sec:preliminaries}: the critical-point transfer under the reparameterization map, the loss decomposition, the centered and plain kernel, and the residual representation of the loss.
 \item \Cref{app:local-minima-proofs} (\emph{Properties of Local Minima}) proves the results of \Cref{sec:local-minima}: feature learning (teacher-span confinement) and the zero-mass exact-fit proposition.
 \item \Cref{app:benign-proofs} (\emph{Benign Loss Landscape for Coplanar Teacher Networks}) proves the main result, \Cref{thm:headline}, through its centered formulation \Cref{thm:centered}, as presented in \Cref{sec:benign}.
 \item \Cref{app:plain-proofs} (\emph{Spurious Local Minima for ReLU}) proves the results of \Cref{sec:plain}: the trap for plain networks at arbitrary overparameterization and the effective-width bound, via the box certificate and the splitting lemma.
 \item \Cref{app:transfer-proofs} (\emph{Finite sample consequence}) proves the result of \Cref{sec:boundaries}: the transfer of the main result to empirical loss, via the minimizer stability theorem.
\end{itemize}

\subsection{Auxiliary Lemmas}
\label{app:preliminary-proofs}
Recall from \Cref{def:minima} that a \emph{saddle} is a critical point that is neither a local minimum nor a local maximum.
In this subsection we give the proofs and proof ideas of \Cref{sec:preliminaries}, namely of the critical-point transfer \Cref{lem:raw-transfer}, the loss decomposition \Cref{lem:equivalence_loss_skip_centered}, the pair-moment \Cref{lem:pair-moments}, and the residual representation \Cref{prop:loss_in_residual}.

\begin{ProofSketch}{Lemma}{lem:raw-transfer}
The map $\sigma(s',w')=(s'_i/\|a_i\|,\,\|a_i\|w'_i)_i$ is continuous, satisfies $\Pi\circ\sigma=\mathrm{id}$ and $\sigma(\Pi(o,a))=(o,a)$. Hence $\Pi$ maps every neighborhood of $(o,a)$ onto a set containing a neighborhood of $\Pi(o,a)$, and since $\Pi$ is continuous, the preimage of a neighborhood of $\Pi(o,a)$ is a neighborhood of $(o,a)$. As $\mathscr L_q\circ\Pi$ and $\mathscr L_q$ take the same values on corresponding parameters, local minimality and global minimality transfer in both directions. For critical points, the chain rule with $s_i=o_i\|a_i\|$ and $w_i=a_i/\|a_i\|$ gives
\begin{align}
    \partial_{o_i}(\mathscr L_q\circ\Pi)=\|a_i\|\,\partial_{s_i}\mathscr L_q,\qquad
    \nabla_{a_i}(\mathscr L_q\circ\Pi)=o_i\,\partial_{s_i}\mathscr L_q\;w_i+\frac{1}{\|a_i\|}\nabla_{\sphere, w_i}\mathscr L_q ,
    \label{eq:raw-chain-rule}
\end{align}
evaluated at $\Pi(o,a)$, where the two summands of the second gradient are orthogonal. Since $a_i\ne0$, the raw gradient vanishes if and only if $\partial_{s_i}\mathscr L_q=0$ and also the spherical gradient $\nabla_{\sphere, w_i}\mathscr L_q=0$ for all $i$.
\end{ProofSketch}

\begin{Proof}{Lemma}{lem:equivalence_loss_skip_centered}
Expanding the square in the definition of the skip loss \eqref{eq:losses} after substituting \eqref{eq:skip-centered-split}, the cross term between the difference of centered model outputs and the linear residual has zero expectation: the difference of centered model outputs is even in the input and the linear residual is odd, so their product is odd and hence has zero expectation under the centrally symmetric Gaussian law. The squared linear residual has expectation equal to the squared Euclidean norm of its coefficient vector, since the Gaussian has identity covariance. This gives \eqref{eq:skip-connection}; setting both skip vectors to zero gives \eqref{eq:loss-connection}.

The smooth coordinate change $(b^{\mathbf s},s,w)\leftrightarrow(u,s,w)$ reduces the loss to $\LC(s,w)+\|u\|_2^2/2$. Differentiation gives the critical-point equivalence. Local minimality forces $u=0$ because moving non-zero $u$ toward zero lowers the loss; restriction to $u=0$ gives a centered minimum and preserves strictness. Conversely, adding the non-negative quadratic preserves a centered minimum. If that minimum is strict, any different nearby parameter increases either the centered term or the quadratic, proving strictness for the skip loss.
\end{Proof}

\begin{figure}[!htbp]
 \centering
 \begin{minipage}[t]{.285970\linewidth}
  \vspace{0pt}
  \centering
  \includegraphics[width=\linewidth]{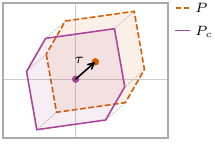}
  \par\small (a) Polytope translation
 \end{minipage}\hfill
 \begin{minipage}[t]{.359075\linewidth}
  \vspace{0pt}
  \centering
  \includegraphics[width=\linewidth]{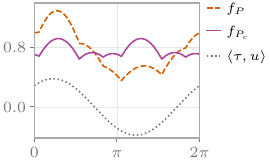}
  \par\small (b) Support functions
 \end{minipage}\hfill
 \begin{minipage}[t]{.344955\linewidth}
  \vspace{0pt}
  \centering
  \includegraphics[width=\linewidth]{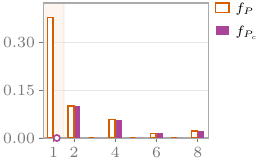}
  \par\small (c) Fourier amplitudes
 \end{minipage}
 \caption{The split \eqref{eq:skip-centered-split} as a translation of a
 zonotope, for one planar student network $(s,w)$ with all $s_i>0$. A network with
 non-negative masses is the \emph{support function}
 $u\mapsto\max_{x\in K}\ip{u}{x}$ of a zonotope $K$: the plain network of
 $P=\sum_is_i[0,w_i]$, the centered network of
 $P_c=\sum_is_i[-\tfrac12w_i,\tfrac12w_i]$.
 (a)~The two zonotopes and the shift
 $\tau=\tau(s,w)=\tfrac12\sum_is_iw_i$, the \emph{Steiner point}, with
 $P=P_c+\tau$ (arrow).
 (b)~The support functions $f_P=f_{R;s,w}$ and $f_{P_c}=f_{C;s,w}$ on the unit
 circle $u=(\cos\theta,\sin\theta)$; their difference is the odd term
 $\ip{\tau}{u}$, and $f_{P_c}$ is $\pi$-periodic. (c)~Angular Fourier
 amplitudes of $\theta\mapsto f_P(u)$ and $\theta\mapsto f_{P_c}(u)$:
 centering removes the degree-one coefficient and leaves the even degrees
 unchanged. Support-function language needs $s_i\ge0$;
 \eqref{eq:skip-centered-split} and \Cref{lem:equivalence_loss_skip_centered}
 hold for potentially negative masses.}
 \label{fig:zonotope-first-harmonic}
\end{figure}

\Cref{fig:zonotope-first-harmonic} shows the split \eqref{eq:skip-centered-split} behind this proof for a planar student network with non-negative masses, where both networks represent support functions of zonotopes and the linear term is the translation by the Steiner point.

\begin{Proof}{Lemma}{lem:pair-moments}
As only the projection of the Gaussian $X$ on $\operatorname{span}(\xi, \zeta)$ enters, we may assume $d=2$ and by rotational invariance we may assume $\xi=(1,0)$ and $\zeta=(\cos\vartheta,\sin\vartheta)$. The two ReLU factors $[R\cos a]_+$ and $[R\cos(a-\vartheta)]_+$ are both positive exactly for $a\in(\vartheta-\pi/2,\pi/2)$, so, writing the projected input in polar coordinates $(R\cos a,R\sin a)$ with $a$ uniform on $[0,2\pi)$ independent of $R$ and $\E R^2=2$,
\begin{align}
 M_R(\xi,\zeta)
 &=\frac{\E R^2}{2\pi}\int_{\vartheta-\pi/2}^{\pi/2}\cos a\cos(a-\vartheta)\,da
 =\frac{\sin\vartheta+(\pi-\vartheta)\cos\vartheta}{2\pi},
 \label{eq:relu-pair-angular}
\end{align}
where $2\cos a\cos(a-\vartheta)=\cos\vartheta+\cos(2a-\vartheta)$ has the antiderivative $a\cos\vartheta+\tfrac12\sin(2a-\vartheta)$, and $\sin(\pi-\vartheta)-\sin(\vartheta-\pi)=2\sin\vartheta$, yielding the derived equation. Substituting $\sin\vartheta=\sqrt{1-\rho^2}$ and $\pi-\vartheta=\pi/2+\arcsin\rho$ gives $2\pi M_R=\widetilde\Phi_C(\rho)+\frac\pi2\rho$. For the centered feature, $\varphi_R(z)=\varphi_C(z)+z/2$ and the cancellation of the even--odd cross terms under $X\mapsto-X$ give $M_R=M_C+\tfrac14\E[\ip\xi X\ip\zeta X]=M_C+\rho/4$, yielding $2\pi M_C=\widetilde\Phi_C(\rho)$.
\end{Proof}

\begin{Proof}{Proposition}{prop:loss_in_residual}
Expanding the square in \eqref{eq:q-losses} and applying \Cref{lem:pair-moments} to every pair of unit directions gives
\begin{align}
 4\pi\mathscr L_q
 =\sum_{i,j}s_is_j\widetilde\Phi_q(\ip{w_i}{w_j})
 -2\sum_{i,k}s_it_k\widetilde\Phi_q(\ip{w_i}{v_k})
 +\sum_{k,l}t_kt_l\widetilde\Phi_q(\ip{v_k}{v_l}).
 \label{eq:loss-quadratic}
\end{align}
Regrouping the right side as
\begin{align}
    \sum_is_i\Bigl[\sum_js_j\widetilde\Phi_q(\ip{w_i}{w_j})-\sum_kt_k\widetilde\Phi_q(\ip{w_i}{v_k})\Bigr]
    \notag\\
    \qquad-\sum_kt_k\Bigl[\sum_js_j\widetilde\Phi_q(\ip{v_k}{w_j})-\sum_lt_l\widetilde\Phi_q(\ip{v_k}{v_l})\Bigr]
    \label{eq:loss-quadratic-regrouped}
\end{align}
and using the definition of the residual \eqref{eq:ambient-residual-main} gives \eqref{eq:residual-loss}. The right side of \eqref{eq:loss-quadratic} is a quadratic polynomial in the student masses; differentiating it in $s_i$ collects the two symmetric student sums $\sum_js_j\widetilde\Phi_q(\ip{w_i}{w_j})$ and the teacher sum $2\sum_k t_k\widetilde\Phi_q(\ip{w_i}{v_k})$ into $2\mathcal F_q(w_i)$, which is \eqref{eq:residual-mass-gradient}.

For the angle identity of \eqref{eq:residual-mass-gradient}, move $w_i$ along the sphere, $w_i(\tau)=w_i+\tau h$ with $h\perp w_i$. The diagonal term $s_i^2\widetilde\Phi_q(1)$ is constant, and every other term containing $w_i$ appears twice in the symmetric student sum and once in the teacher sum, so
\begin{align}
 4\pi\,\frac{d}{d\tau}\Big|_{\tau=0}\mathscr L_q
 &=2s_i\Bigl[\sum_{j\ne i}s_j\widetilde\Phi_q'(\ip{w_i}{w_j})\ip{h}{w_j}-\sum_kt_k\widetilde\Phi_q'(\ip{w_i}{v_k})\ip{h}{v_k}\Bigr]
 \notag\\
 &=2s_i\,D\mathcal F_q(w_i)[h],
 \label{eq:residual-angle-derivative}
\end{align}
where $D\mathcal F_q(w_i)[h]$ is the derivative of the probe $u\mapsto\mathcal F_q(u)$ at $u=w_i$ in the tangential direction $h$. This is the stated spherical-gradient identity.
\end{Proof}

\subsection{Properties of Local Minima}
\label{app:local-minima-proofs}
In this subsection we give the proof ideas of \Cref{sec:local-minima}, namely of the feature-learning \Cref{prop:teacher-span-confinement} and the zero-mass \Cref{prop:zero_mass_exact_fit,prop:nc-dead-student}.

\begin{ProofSketch}{Proposition}{prop:teacher-span-confinement}
Fix $q\in\{C,R\}$, the input dimension $d\ge2$, teacher parameters
$(t,v)$ with arbitrary masses, and student parameters $(s,w)$
with non-negative masses $s_i\ge0$ constituting a local minimum
of the $q$-loss $\mathscr L_q$.
Let us denote the linear span of the teacher directions,
the \emph{teacher span}, by
$\mathcal T=\operatorname{span}\{v_k\}$.
Decompose each student direction as $w_i=p_i+q_i$, with
$p_i\in\mathcal T$ and $q_i\perp\mathcal T$.
We aim to show that $q_i=0$ for every positive-mass student unit.

We may assume all student masses to be positive: zero-mass
units may be pruned without changing the represented function
or the local minimality assumption.
If no units remain, the claim is immediate.
Absorb the masses into the student weights and work with the
generators $V_i=s_iw_i$, interpreted as raw first-layer weights
with fixed unit output weights.
This preserves the network function and local minimality by
\Cref{lem:raw-transfer}.
In these coordinates, write
\begin{align}
 V&=(s_1w_1,\ldots,s_nw_n),&
 f_V(x)&=\sum_i\varphi_q(\ip{V_i}{x}).
 \notag
\end{align}
The idea of the proof is to compare the local curvature of
the loss along two motions:
the \emph{normal motion} stretches each $s_iq_i$ while keeping
the projection of the generator onto $\mathcal T$ fixed,
whereas the \emph{radial motion} scales all generators uniformly.
Their velocities and paths are
\begin{align}
 z_N&=(s_1q_1,\ldots,s_nq_n),&
 V+\tau z_N&=(s_ip_i+(1+\tau)s_iq_i)_i,\notag\\
 z_A&=V,&
 V+az_A&=(1+a)V.
 \notag
\end{align}
We call the first and second variations along these motions
the \emph{normal and radial derivatives and curvatures}, respectively.

We take special care to justify the second and mixed
derivatives used below, without assuming that the full loss
admits a Hessian everywhere.
The proof shows that its restriction to
$V+\mathcal W$, where
$\mathcal W=\operatorname{span}\{z_A,z_N,z_{e,b}\}$
and the shear velocities $z_{e,b}$ are defined below,
is $C^2$ near $V$.
Write $H$ for the Hessian of this restriction at $V$, acting
on $\mathcal W$.
Thus, for $z,z'\in\mathcal W$,
\begin{align}
 \partial_z\mathscr L_q(V)
 &:=\frac{d}{d\tau}\Big|_{\tau=0}
       \mathscr L_q(V+\tau z),\notag\\
 z^\top Hz'
 &=\partial_a\partial_\tau\Big|_{(a,\tau)=(0,0)}
       \mathscr L_q(V+az+\tau z').
 \label{eq:directional-hessian-notation}
\end{align}
Local minimality implies
$\partial_z\mathscr L_q(V)=0$ and $z^\top Hz\ge0$
for every $z\in\mathcal W$.

The first step shows that the normal curvature is zero.
The argument evaluates its sum with the curvatures along
\emph{shear motions}: these move each generator within
$\mathcal T$, with velocity determined by its normal component.
For $e$ and $b$ ranging over orthonormal bases of
$\mathcal T$ and $\R^d$, respectively, define
\begin{align}
 z_{e,b}
 &=\bigl(s_1\ip{b}{q_1}e,\ldots,
         s_n\ip{b}{q_n}e\bigr).
 \notag
\end{align}
For $q=C$, Parseval's identity and cancellations in the explicit
curvature formulas show that the sum of the normal and shear
curvatures equals $(\dim\mathcal T+1)$ times the normal
derivative of the loss:
\begin{align}
 z_N^\top Hz_N+\sum_{e,b}z_{e,b}^\top Hz_{e,b}
 &=(\dim\mathcal T+1)\,\partial_{z_N}\LC(V).
 \label{eq:global-trace}
\end{align}
For $q=R$, the additional first-moment quadratic satisfies
the same identity:
\begin{align}
 \Delta(V)
 &:=\LR(V)-\LC(V)
   =\frac18\left\|\sum_iV_i-\sum_kt_kv_k\right\|_2^2,
   \notag\\
 \partial_{z_N}^2\Delta(V)
 +\sum_{e,b}\partial_{z_{e,b}}^2\Delta(V)
 &=(\dim\mathcal T+1)\,\partial_{z_N}\Delta(V).
 \label{eq:plain-correction-trace}
\end{align}
Adding this identity proves \eqref{eq:global-trace} with
$\LC$ replaced by $\LR$.
In either case, the normal derivative vanishes at the local
minimum, while all curvatures on the left are non-negative.
Consequently, $z_N^\top Hz_N=0$.

This also forces the mixed normal--radial curvature to vanish:
by Cauchy--Schwarz for the positive-semidefinite restricted
Hessian, its square is bounded by the product of the normal
and radial curvatures,
\begin{align}
 (z_A^\top Hz_N)^2
 &\le(z_A^\top Hz_A)(z_N^\top Hz_N)=0.
 \notag
\end{align}

Next, we identify the mixed curvature with half the normal
derivative of the \emph{second student moment}
$\E[f_V(X)^2]$, a quantity independent of the teacher network.
To see this, first differentiate the loss along the normal
motion: this gives the expected product of the student--teacher
difference and the normal derivative of the student output,
\begin{align}
 \frac{d}{d\tau}\mathscr L_q(V+\tau z_N)
 &=\E\left[
   \bigl(f_{V+\tau z_N}-f_{q;t,v}\bigr)(X)
   \frac{d}{d\tau}f_{V+\tau z_N}(X)
   \right],
 \label{eq:loss-first-derivative}\\
 \frac{d}{d\tau}f_{V+\tau z_N}(x)
 &=\sum_i
   \varphi_q'\bigl(\ip{V_i+\tau(z_N)_i}{x}\bigr)
   \ip{(z_N)_i}{x}.
 \label{eq:student-first-derivative}
\end{align}
These feature derivatives exist almost everywhere.
Now differentiate this product along the radial motion,
keeping the normal displacement $z_N$ fixed:
positive homogeneity makes the radial derivative of the
student--teacher difference equal to $f_V$, while the student network's
normal response remains unchanged almost everywhere under
positive radial scaling.
Dominated convergence justifies differentiating under the
expectation.
The remaining product therefore gives
\begin{align}
 z_A^\top Hz_N
    =\E\left[
      f_V(X)\,
      \frac{d}{d\tau}\Big|_{\tau=0}
      f_{V+\tau z_N}(X)
     \right]
    =\frac12\,\frac{d}{d\tau}\Big|_{\tau=0}
      \E\bigl[f_{V+\tau z_N}(X)^2\bigr].
 \label{eq:mixed-second-moment}
\end{align}

Let us now assume that some $q_i\ne0$.
Write $P=V-z_N=(s_ip_i)_i$ for the projected generators and
parameterize the second student moment by their normal scale:
\begin{align}
 M(c)&:=\E\bigl[f_{P+cz_N}(X)^2\bigr],&
 V+\tau z_N&=P+(1+\tau)z_N.
 \notag
\end{align}
Thus $M(0)$ is the projected student network's second moment,
$M(1)$ is the original student network's second moment, and
$z_A^\top Hz_N=\tfrac12M'(1)$.

To show that this normal derivative is positive, first observe
that the second moment is convex along the normal motion.
Indeed, for fixed input the pre-activations depend affinely
on the normal scale:
\begin{align}
 f_{P+cz_N}(x)
 &=\sum_i\varphi_q\bigl(
      \ip{P_i}{x}+c\ip{(z_N)_i}{x}
    \bigr).
 \notag
\end{align}
Since both activations are convex and non-negative, this output
and its square are convex in $c$.
As integration preserves convexity, so $c\longmapsto M(c)$ is a convex function.
Geometrically, the derivative at the original generators is
at least the secant slope from the projected generators:
$M'(1)\ge M(1)-M(0)$.
It therefore suffices to show that the normal parts of the generators increase the second student moment, this is $M(1) \ge M(0)$.

For this comparison, reflect each input $x$ across the teacher
span $\mathcal T$, and write $Rx$ for its reflection.
Their midpoint is the projection of $x$ onto $\mathcal T$.
Since the student output as a function of the network input is convex and non-negative, its squared
output at this midpoint is at most the average of its squared
outputs at $x$ and $Rx$.
Evaluating the student network at the projected input is equivalent to
evaluating the student network with projected generators $P$ at the
original input.

If some $s_iq_i\ne0$, this comparison is strict at $x_0=s_iq_i$:
the projected student output vanishes there, whereas the
$i$-th original student unit contributes positively.
By continuity, strictness persists on an open neighborhood,
which has positive Gaussian probability by full support.
Since reflection preserves the Gaussian law, integrating the
comparison shows that the projected second moment $M(0)$
is strictly smaller than the original second moment $M(1)$.
As this leads to $z_A^T H z_N = \frac{1}{2} M'(1) > 0$, a contradiction to the first step, this closes the proof of \Cref{prop:teacher-span-confinement}.
\end{ProofSketch}

\begin{ProofSketch}{Proposition}{prop:zero_mass_exact_fit}
Let the input dimension $d\ge1$ and let the student masses and directions $(s,w)$ constitute a local minimum of the centered loss $\LC$ (for $d\ge1$) or the plain loss $\LR$ (for $d\ge3$), and let us assume that the teacher masses are non-negative in case $d=2$ (for $\LC$).
Moreover, let us assume that there is a zero-mass unit $(s_i, w_i)$. We want to prove that the student network is an exact fit.

In the first step we show for both $q\in\{C,R\}$ that the residual $\mathcal F_q$, see \eqref{eq:ambient-residual-main}, vanishes on an open set $U \subseteq \sphere^{d-1}$, see \leanlinkdecl{Geometry.lean}{4738}{PaperLeanFormalization.Geometry.stp_zm_revival}{Lean}. The argument analyzes the change of loss when altering the mass $s_i$ and direction $w_i$ of the zero-mass unit. In case the residual does not vanish on an open set around the direction $w_i$, we would be able to find student parameters arbitrarily close to $(s,w)$ with smaller loss, a contradiction. The argument works using the representation of the loss as a quadratic polynomial in the student mass coordinates with the values of the residual at the respective student and teacher directions as the coefficients (see \Cref{prop:loss_in_residual}, \leanlinkdecl{Geometry.lean}{4636}{PaperLeanFormalization.Geometry.eq_revival_signed}{Lean}).

This representation also comes in handy to show the final statement of the theorem, namely that the loss is zero: by \Cref{prop:loss_in_residual} it suffices to show that the residual vanishes at the student and teacher directions, see \leanlinkdecl{Geometry.lean}{4763}{PaperLeanFormalization.Geometry.stp_zm_pairing}{Lean}.
Hence, in our second step we show that the residual vanishes on the whole sphere, see \leanlinkdecl{Geometry.lean}{4517}{PaperLeanFormalization.Geometry.stp_zm_propagation}{Lean}. For $d\ge3$ this follows from an elegant argument using the fact that an analytic function vanishing on an open set vanishes everywhere. The centered kernel \eqref{eq:kernels} is analytic on $(-1,0)\cup(0,1)$ and the plain kernel is analytic on $(-1,1)$; in either case it is easy to see that the residual is analytic away from the finitely many points $\{ \pm w_i, \pm v_k \ | \ i,k  \}$, which we call its \emph{atoms}.
Hence, it is also analytic restricted to \emph{great circles}, the intersection of the sphere with a hyperplane.
We observe that for any point $x$ on the sphere $\sphere^{d-1}$, there is a great circle through it that intersects the zero set $U$ in an open set and avoids the finitely many residual atoms. As the residual restricted to this great circle has to vanish, it vanishes at $x$. As the residual vanishes at all points on the sphere except the atoms, its continuity implies it also vanishes at the atoms. This concludes the proof for the case $d \ge 3$.

For $d=1$, the residual is centrally symmetric, so the zero set from the first step extends to the whole sphere. For $d=2$, where the unit circle $\sphere^{d-1}$ is chopped into pieces by the residual atoms, we use the machinery developed in \Cref{app:benign-proofs} and non-negative teacher weights to propagate the zero set inductively around the circle.

\finishes{Proposition}{prop:zero_mass_exact_fit}
\end{ProofSketch}

\subsection{Benign Loss Landscape for Coplanar Teacher Networks}
\label{app:benign-proofs}
In this subsection we give the proof ideas of the main result, \Cref{thm:headline} through its centered formulation, \Cref{thm:centered}.

Denote by $(t,v)\in\Ppos{d}{m}\cap\Ppl{d}{m}$ the teacher network's mass--direction vector pair throughout, and let the student width satisfy $n\ge m$. Whenever we call a student parameter $(s,w)$ a local minimum or critical point of $\LC$, it is with respect to these fixed teacher parameters $(t,v)$. The positivity of the teacher masses enters through one fact only: a line of positive net mass carries a student unit of positive mass, compare to the definition of the net mass $\omega$ in \eqref{eq:net-weight-main}. \Cref{lem:critical-points-are-double-zeros,lem:beam-kernel,lem:gap-sign,lem:line-count,lem:opposed-rotation} do not use teacher mass non-negativity and hold for unrestricted teacher masses.

A hypothetical spurious local minimum is progressively restricted by four facts. A zero-mass student unit forces an exact fit (\Cref{prop:zero_mass_exact_fit}). Otherwise, confinement (\Cref{prop:teacher-span-confinement}) places every student direction in the teacher span, which reduces the problem to the plane. There, the necessary positive-mass configuration of matched width and strictly interlacing student and teacher lines already resolves the overparameterized case (\Cref{prop:locamin_are_line_separated}). For the matched width case, we need to take one step further: in the given setting \Cref{prop:beam-descent} constructs a direction in parameter space certified to decrease the loss, so also this last case cannot be a local minimum. Finally, \Cref{lem:equivalence_loss_skip_centered} carries the result about the centered to the skip network (\Cref{thm:headline}).
For the analysis below take $d\ge2$; the case $d=1$ is treated directly in the proof of \Cref{thm:centered}.
By the assumption, the teacher directions $\mathcal T=\operatorname{span}\{v_k\}$ are contained in a two-dimensional linear subspace that we identify with
$\R^2$.

With student and teacher directions in $\R^2$, it is convenient to work with angles $e(\gamma)$, $\theta_i,\beta_k$ and lines $\ell(u)$ of \Cref{sec:benign} (\Cref{fig:interlacing}a); every direction $u\in\sphere^1$ is $e(\gamma)$ for a unique angle $\gamma=:\theta(u)\in\R/2\pi\Z$, and the line of an angle $\gamma$ is $\ell(\gamma)=\gamma\bmod\pi$, so that $\ell(u)=\ell(\theta(u))$. Throughout, $\alpha$ denotes a line and $\gamma,\theta,\beta$ denote angles.

We call $(t,\beta)$ and $(s,\theta)$ the \emph{angular representation} of the teacher network and the student network, respectively, and write $\mathscr L_q(s,\theta)$ for the loss in these coordinates. Let $\Lambda_{v} \coloneqq \{ \ell(v_k)\mid k \in [m] \}$ and $\Lambda_w \coloneqq \{ \ell(w_i) \mid  i \in [n] \}\subset\R/\pi\Z$ be the
finite sets of teacher and student lines and their union $\Lambda_{v,w} =
\Lambda_v \cup \Lambda_w$. The net mass $\omega$ \eqref{eq:net-weight-main} vanishes off $\Lambda_{v,w}$, so $S_\pm\subseteq\Lambda_{v,w}$.

Our planar theorems mostly rely on the connection of the loss with the residual presented in \Cref{prop:loss_in_residual}; the key property we exploit is the residual's relation to the \emph{angular kernel}, the kernel \eqref{eq:kernels} evaluated at the angle $\tau$ between two directions. As $\arcsin(\cos\tau)=\frac\pi2-|\tau|$ and $\sqrt{1-\cos^2\tau}=|\sin\tau|$ for $|\tau|\le\pi$, it is explicitly given by
\begin{align}
    \Phi_q(\tau)&=\widetilde\Phi_q(\cos\tau), \\
    \Phi_C(\tau)&=\Bigl(\frac\pi2-|\tau|\Bigr)\cos\tau+|\sin\tau|,\\
    \Phi_R(\tau)&=(\pi-|\tau|)\cos\tau+|\sin\tau|
    \label{eq:angular-kernel}
\end{align}
extended $2\pi$-periodically. The key identity is

\begin{align}
    \Phi_q''(\tau)+\Phi_q(\tau)=2|\sin \tau| \quad\text{ for }\quad q \in \{ C, R \},
    \label{eq:angular-kernel-d2+1}
\end{align}
which we will use to determine other representations of the angular residual. Before this work, let us observe that the angular kernel is even and $2\pi$-periodic, and $\Phi_C$ is moreover $\pi$-periodic with $1\le\Phi_C\le\pi/2$, while $\Phi_R=\Phi_C+\frac\pi2\cos \tau$ satisfies $\Phi_R(\pi)=0$.

\begin{figure}[t]
 \centering
 \begin{minipage}[b]{.49\linewidth}
  \centering
  \includegraphics[width=\linewidth]{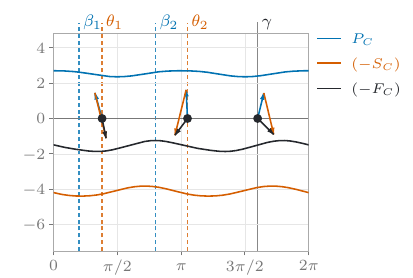}
  \par\small (a) Centered: mass rate $2\pi\dot s$
 \end{minipage}\hfill
 \begin{minipage}[b]{.49\linewidth}
  \centering
  \includegraphics[width=\linewidth]{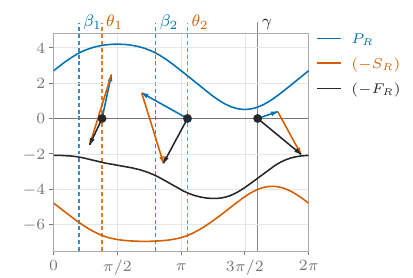}
  \par\small (b) Plain: mass rate $2\pi\dot s$
 \end{minipage}
 \caption{Pull, push, and descent for an added probe of mass $s=1$.
 Background teacher units lie at $\beta=(0.2\pi,0.8\pi)$ (blue dashed guides)
 and student units at $\theta=(0.38\pi,1.05\pi)$ (orange dashed guides), all with
 mass $1$. Solid curves show $P_q$ (blue), $-S_q$ (orange), and
 $-\mathcal F_q=P_q-S_q$ (black), evaluated at $\gamma$ for the configuration
 with the probe adjoined there (see \eqref{eq:residual-potential-paper} for the definition of $\mathcal F_q$). By \Cref{rem:planar_residual_equivalence} the value of the residual gives the gradient of the student masses while the slope gives the gradient for the student angles. Thus angular pull climbs the teacher potential, while angular push descends the student potential. The three dots at the student angles and a third angle each consider one probe separately; blue and orange arrows add head to tail to the black descent arrow.}
 \label{fig:pull-push-step}
\end{figure}

The \emph{student potential} $S_q(\gamma)=\sum_i s_i\Phi_q(\gamma-\theta_i)$ and
\emph{teacher potential} $P_q(\gamma)=\sum_k t_k\Phi_q(\gamma-\beta_k)$
give the \emph{angular residual} $\mathcal F_q\colon \R/2\pi\Z\to \R$, namely $\mathcal F_q=S_q-P_q$:
\begin{align}
 \mathcal F_q(\gamma)
 =\sum_i s_i\Phi_q(\gamma-\theta_i)-\sum_k t_k\Phi_q(\gamma-\beta_k).
 \label{eq:residual-potential-paper}
\end{align}
Compare this to the non-angular residual definition in \eqref{eq:ambient-residual-main} and take a look at \Cref{fig:pull-push-step} for intuition about the connection of the mass and angular gradient of the student units to the angular residual.
For $q=C$ the angular residual is $\pi$-periodic, hence a function of lines, and collecting the units on each line gives $\mathcal F_C(\gamma)=\sum_{\alpha\in\Lambda_{v,w}}\omega(\alpha)\Phi_C(\gamma-\alpha)$, where $\Phi_C(\gamma-\alpha)$ is well defined because $\Phi_C$ is $\pi$-periodic.
\begin{remark}[\leandecl{Theorems.lean}{170}{PaperLeanFormalization.Theorems.rem_planar_residual_equivalence}]
\label{rem:planar_residual_equivalence}
 For $\gamma \in [0,2\pi]$ with $u=e(\gamma)$,
the angular residual satisfies $\mathcal F_q(\gamma)=\mathcal F_q(u)$ as well as $\partial_{s_i}\mathscr L_q=\frac{1}{2\pi}\mathcal F_q(\theta_i)$ and $\partial_{\theta_i}\mathscr L_q= \frac{1}{2\pi}s_i\mathcal F_q'(\theta_i)$.
\end{remark}

A point $\alpha\in\R/\pi\Z$ is a \emph{double zero} of $\mathcal F_C$ if
$\mathcal F_C(\alpha)=\mathcal F_C'(\alpha)=0$.

\begin{lemma}[Double zeros of $\mathcal F_C$ at critical points, \leandecl{Theorems.lean}{258}{PaperLeanFormalization.Theorems.lem_critical_points_are_double_zeros}]
\label{lem:critical-points-are-double-zeros}
If $(s,w)$ is a critical point of $\LC$, then $\ell(w_i)$ is a double zero of $\mathcal F_C$ for all $i$ with $s_i \neq 0$.
\end{lemma}

Let us define the beam operator
\begin{align}
    L = (D^2+1)^2 \label{eq:definition_beam_operator}.
\end{align}
In the following lemma we prove that the kernel of $L$ on every interval is the following four-dimensional space of functions on $\R$.
\begin{align}
    \mathcal V = \operatorname{span}\{\cos x,\sin x,x\cos x,x\sin x\}
    \label{eq:definition_V}
\end{align}

\begin{lemma}[The kernel of the beam operator, \leandecl{Theorems.lean}{268}{PaperLeanFormalization.Theorems.lem_beam_kernel}]
\label{lem:beam-kernel}
Let $I\subseteq\R$ be a non-empty open interval. A function $f\in C^4(I)$ satisfies $Lf=0$ on $I$ if and only if $f$ is the restriction to $I$ of an element of $\mathcal V$.
\end{lemma}

\begin{ProofSketch}{Lemma}{lem:beam-kernel}
Each function of $\mathcal V$ solves $Lf=0$ by direct differentiation; the four generators correspond to the double roots $\pm i$ of the characteristic polynomial $(\lambda^2+1)^2$. They are linearly independent on $I$: the combination $g = a\cos x+b\sin x+cx\cos x+dx\sin x$ has the jet $(g,g',g'',g''')(0)=(a,\,b+c,\,-a+2d,\,-b-3c)$ at $0$, which vanishes only for $a=b=c=d=0$, and a real-analytic function vanishing on an interval vanishes on $\R$. Conversely, $L$ is a linear differential operator of order four with continuous coefficients, so by the uniqueness theorem for linear ordinary differential equations a solution on $I$ is determined by its values $(f,f',f'',f''')$ at one point of $I$; hence the kernel has dimension at most four and is spanned by the four restricted generators.
\end{ProofSketch}

\begin{lemma}[The angular residual on a gap, \leandecl{Theorems.lean}{275}{PaperLeanFormalization.Theorems.lem_gap_sign}]
\label{lem:gap-sign}
Let $(a,b)$ be a gap of the set $S_+$ of lines of positive net mass \eqref{eq:net-weight-main}, of length $l=b-a\in(0,\pi]$, whose endpoints $a,b$ are double zeros of the centered angular residual $\mathcal F_C$ \eqref{eq:residual-potential-paper}. Then there is a function $G(\cdot\,;\zeta)\colon[0,l]\to [0, \infty)$, depending on $l$ and on a parameter $\zeta\in(0,l)$, such that the residual on $(a,b)$ is a sum of contributions of the points of $(a,b)\cap S_-$ only,
\begin{align}
     \mathcal F_C(a+z)=4\sum_{p\in S_-\cap(a,b)}\omega(p)\,G(z;\,p-a)\qquad(0<z<l).
     \label{eq:gap-green-representation}
\end{align}
Consequently, since $\omega(p)<0$ for $p\in S_-$,
\begin{enumerate}
    \item $\mathcal F_C\equiv0$ on $[a,b]$ if and only if $(a,b)\cap S_-=\varnothing$, and otherwise $\mathcal F_C<0$ on $(a,b)$;
    \item if $(a,b)\cap S_-\neq\varnothing$, then $\mathcal F_C''(a)<0$ and $\mathcal F_C''(b)<0$.
\end{enumerate}
\end{lemma}
\Cref{fig:residual-green} illustrates the double zeros and the Green superposition.

\begin{ProofSketch}{Lemma}{lem:gap-sign}
The key observation is that $\mathcal F_C$ is a piecewise-$\mathcal V$ function whose breakpoints are given by $S_+ \cup S_-$.
Furthermore, since $\mathcal F_C \in C^2$, the left- and the right-hand pieces agree up to second order at each breakpoint;
only the third derivative $\mathcal F_C'''$ exhibits a jump discontinuity.
Consequently, the main idea is to construct a $C^2$, piecewise-$\mathcal V$ function $G(\,\cdot\,; \zeta)$ for each breakpoint $\zeta$ of $\mathcal F_C(a + \,\cdot\,)\big|_{[0, l]}$.
This function $G(\,\cdot\,; \zeta)$ is designed to have a unique breakpoint at $\zeta$ with a specified jump in $G'''(\zeta)$, while simultaneously satisfying the boundary conditions at the endpoints of $[0, l]$.
A dimension count argument for the underlying function space then guarantees that $\mathcal F_C(a + \,\cdot\,)\big|_{[0, l]}$ can be uniquely represented as a weighted sum~(\ref{eq:gap-green-representation}) of these functions $G(\,\cdot\,; \zeta)$.
Finally, claims 1.--2.\ reduce to verifying two key properties of $G(\,\cdot\,; \zeta)$: its non-negativity on $[0, l]$ and its strict convexity at the endpoints $0$ and $l$.

To demonstrate that $\mathcal F_C$ is piecewise-$\mathcal V$, recall from~(\ref{eq:net-weight-main}) that $\mathcal F_C(\gamma)=\sum_{\alpha \in S_+ \cup S_-}\omega(\alpha)\Phi_C(\gamma-\alpha)$.
According to (\ref{eq:angular-kernel-d2+1}), the angular kernel $\Phi_C$ satisfies $(D^2 + 1)\Phi_C(\gamma) = 2 |\sin(\gamma)|$.
Applying this differential operator to $\mathcal{F}_C$ yields $(D^2 + 1)\mathcal{F}_C(\gamma) = 2 \cdot \sum_{\alpha \in S_+ \cup S_-} \omega(\alpha) |\sin(\gamma - \alpha)|$.
Because functions of the form $\pm \sin(\,\cdot\, - \alpha)$ and their linear combinations form the kernel of $D^2 + 1$, it follows that $(D^2 + 1)^2\mathcal{F}_C\big|_I \equiv 0$ on any interval $I \subset \R / \pi\Z$ with $I \cap (S_+ \cup S_-) = \emptyset$.
By \Cref{lem:beam-kernel}, the restriction of $\mathcal{F}_C$ to $I$ coincides with that of some function from $\mathcal{V}$.
The breakpoints of $\mathcal{F}_C$ are therefore given by $S_+ \cup S_-$; they are characterized by jump of the third derivative:
an explicit calculation shows that $\Phi_C'''(0^+) - \Phi_C'''(0^-) = 4$, hence at each breakpoint $\alpha \in S_+ \cup S_-$ we have $\mathcal{F}_C'''(\alpha^+) - \mathcal{F}_C'''(\alpha^-) = 4\omega(\alpha)$.

Formula~(\ref{eq:gap-green-representation}) follows from a more general decomposition that we establish below.
Specifically, any piecewise~$\cal V$ function $f \in C^2([0, l])$ satisfying the boundary conditions $0 = f(0) = f'(0) = f(l) = f'(l)$ admits the representation
\begin{align}
    f(\,\cdot\,) =
    \sum_{\zeta \text{ is a breakpoint of }f}
    c_{\zeta, f} G(\,\cdot\,; \zeta),
    \quad\text{ where }\quad
    c_{\zeta, f} := f'''(\zeta^+) - f'''(\zeta^-).
    \label{eq:green-decomposition}
\end{align}
Here, a breakpoint refers to a point of discontinuity in the third derivative of $f$.
The basis functions $G(\,\cdot\,; \zeta) \in C^2([0, l])$ are constructed below as the unique functions satisfying the double-zero boundary conditions at $0$ and $l$, while possessing a unique breakpoint at $\zeta$ with a unit third-derivative jump, i.e.\ $c_{\zeta, G(\,\cdot\,; \zeta)} = 1$, and belonging to $\mathcal V$ on each side of $\zeta$.
For any partition $0 =: \zeta_0 < \zeta_1 < \ldots < \zeta_k < \zeta_{k + 1} := l$, these piecewise-$\cal V$ functions with specified breakpoints form a linear space.
Explicitly, this space consists of all functions $f \in C^2([0, l])$ satisfying $0 = f(0) = f'(0) = f(l) = f'(l)$ such that for each $i = 0, \ldots, k$ there exists $f_i \in \cal V$ with $f_i\big|_{[\zeta_i, \zeta_{i + 1}]} \equiv f\big|_{[\zeta_i, \zeta_{i + 1}]}$.
We prove that this space is $k$-dimensional and $G(\,\cdot\,; \zeta_i)$ indeed form its basis.

First, we show that if $c_{\zeta_i, f} = 0$ for all $i = 1, \ldots, k$, then $f \equiv 0$.
Indeed, the assumption implies $f_{i - 1}'''(\zeta_i) = f'''(\zeta_i^-) = f'''(\zeta_i^+) = f_i'''(\zeta_i)$.
Since $f_{i - 1}$ and $f_i$ both lie in the kernel of the differential operator $(D^2 + 1)^2$ and coincide up to third order at $\zeta_i$, the existence and uniqueness theorem
yields $f_{i - 1} \equiv f_i$.
Consequently, $f$ globally belongs to $\cal V$; it remains to show that double zeros of $f$ at $0$ and $l$ imply $f \equiv 0$.
The conditions $f(0) = f'(0) = 0$ leave the two-parameter family $f(x) = \lambda_1 (x \cos x - \sin x) + \lambda_2 x \sin x$, while the double zero at $l$ is equivalent to the linear system:
\begin{align}
    M_l
    \begin{pmatrix}
        \lambda_1 \\ \lambda_2
    \end{pmatrix}
    =
    \begin{pmatrix}
        \kappa_1 \\ \kappa_2
    \end{pmatrix}
    \quad\text{ with }\quad
    M_l :=
    \begin{pmatrix}
        l \cos l - \sin l & l \sin l \\
        -l \sin l & l \cos l + \sin l
    \end{pmatrix},
    \label{eq:green-system}
\end{align}
where $\kappa_1 = \kappa_2 = 0$.
Since $\det M_l = l^2 - \sin^2 l > 0$ for all $l \neq 0$, the matrix $M_l$ is invertible, thus $\lambda_1 = \lambda_2 = 0$ is the unique solution.

Second, we demonstrate that for each $i = 1, \ldots, k$ there exists a piecewise-$\cal V$ function $G(\,\cdot\,; \zeta_i)$ satisfying the boundary conditions at $0$ and $l$, having its unique breakpoint at $\zeta_i$ with a unit jump $c_{\zeta_i, G(\,\cdot\,; \zeta_i)} = 1$.
Consider the auxiliary function $\psi(\gamma) := \sin \gamma - \gamma \cos \gamma$.
A direct calculation verifies that $(\psi, \psi', \psi'', \psi''')(0) = (0, 0, 0, 2)$.
Consequently, we construct
\begin{align}
    G(\gamma; \zeta_i) &:= \frac{1}{2}\psi(\gamma - \zeta _i)\mathbf{1}_{\gamma > \zeta_i} + g(\gamma)
    \quad\text{ with } \notag\\
    g(\gamma) &:=
    \begin{pmatrix}
        \gamma \cos \gamma - \sin \gamma \\
        \gamma \sin \gamma
    \end{pmatrix}^\top
    M_l^{- 1}
    \begin{pmatrix}
        -\psi(l - \zeta_i) / 2 \\
        -\psi'(l - \zeta_i) / 2
    \end{pmatrix}.
    \label{eq:green-construction}
\end{align}
Here $g \in \cal V$ is determined by the unique solution to the linear system~(\ref{eq:green-system}) with appropriate parameters $\kappa_1, \kappa_2$ chosen so that $G(\,\cdot\,; \zeta_i)$ satisfies the boundary conditions.

Applying the above construction~(\ref{eq:green-decomposition}) to a gap $(a, b)$ and the piecewise-$\cal V$ function $\mathcal{F}_C$ yields~(\ref{eq:gap-green-representation}).
An explicit expression for the function $G(\,\cdot\,; \zeta)$ illustrated in \Cref{fig:residual-green}(b) comes from~(\ref{eq:green-construction}).
Direct calculations show that $G(\gamma; \zeta) > 0$ for all $\gamma \in (0, l)$, and that the second derivatives satisfy $G''(0; \zeta), G''(l; \zeta) > 0$ for any $l \in (0,\pi]$.
\finishes{Lemma}{lem:gap-sign}
\end{ProofSketch}

\begin{remark}[The angular residual as a bent ring]
\label{rem:ring}
The beam operator $L$ \eqref{eq:definition_beam_operator} has a classical interpretation in elasticity.
The equations for in-plane bending of a thin elastic circular ring under a
radial load $X$ lead to the fourth-order displacement equation
\citep[Art.~292]{Love1927}
\begin{align}
    (D^2+1)^2u \propto X+C,
    \label{eq:ring-operator}
\end{align}
where $u$ is the radial displacement and $C$ is a constant corresponding
to uniform radial loading and $\propto$ denotes equality up to constants. Thus, up to physical scaling and this uniform
load $C$ which we assume to be zero, the point forces
\begin{align}
    X=\sum_\alpha \omega(\alpha)\delta_\alpha \notag
\end{align}
produce the same differential equation as our residual,
\begin{align}
    (D^2+1)^2\mathcal F_C
    =4\sum_\alpha\omega(\alpha)\delta_\alpha.
    \label{eq:LFC}
\end{align}
Accordingly, $\mathcal F_C$ may be viewed as the radial displacement of a circular
ring under point forces at the student and teacher lines, with strengths
proportional to their net masses.
\end{remark}

\begin{lemma}[Interlacing, \leandecl{Theorems.lean}{299}{PaperLeanFormalization.Theorems.lem_line_count}]
\label{lem:line-count}
Suppose that $|S_+|\ge2$ and that every point of $S_+$ is a double zero
of $\mathcal F_C$. Then every gap of $S_+$ contains a point of $S_-$. Consequently,
\(
|S_+|\le |S_-|.
\)

\end{lemma}
\begin{ProofSketch}{Lemma}{lem:line-count}
Let $(a,b)$ be a gap of $S_+$. Its endpoints are double zeros, so \Cref{lem:gap-sign} gives
\begin{align}
\label{eq:gap-derivative}
    (a,b) \cap S_-= \varnothing \implies F''_C(a)=F''_C(b)=0 \text{ and }    (a,b) \cap S_-\neq \varnothing \implies  F''_C(a),F''_C(b)<0.
\end{align}
Then, since $\mathcal F_C''$ is continuous and neighboring gaps share an endpoint, it follows that either all gaps must contain a point of $S_-$ or none does. The latter option is impossible, since $S_- = \emptyset$ would imply that $\mathcal F_C>0$, contradicting the assumption that the points in $S_+$ are double zeros. Hence all gaps must contain a point of $S_-$. The $|S_+|$ gaps are pairwise disjoint, so $|S_+|\le|S_-|$.
\end{ProofSketch}

\begin{lemma}[Colliding student units provide descent in the centered case, \leandecl{Theorems.lean}{314}{PaperLeanFormalization.Theorems.lem_opposed_rotation}]
\label{lem:opposed-rotation}
Let $(s,\theta)$ be a planar student parameter in which two units $i\neq j$ lie on a common line
$\ell(\theta_i)=\ell(\theta_j)=:\alpha$, and define $z=(\delta s,\delta\theta)$ by
$\delta s=0$,
$\delta\theta_i=s_j$, $\delta\theta_j=-s_i$,
and $\delta\theta_r=0$ for $r\notin\{i,j\}$. Then
\begin{align}
D^2\LC(s,\theta)[z,z]
=
\frac{s_is_j(s_i+s_j)}{2\pi}\,\mathcal F_C''(\alpha).
\label{eq:opposed-rotation-curvature}
\end{align}
In particular, if $s_i,s_j>0$ and $\mathcal F_C''(\alpha)<0$, then $z$ is a direction of strictly negative curvature, so $(s,w)$ is not a local minimum.
\end{lemma}

\begin{ProofSketch}{Lemma}{lem:opposed-rotation}
Since $\Phi_C$ is $\pi$-periodic we may assume $\theta_i=\theta_j$, and then $\mathcal F_C''(\alpha)=\mathcal F_C''(\theta_i)$. By \Cref{prop:loss_in_residual} and \Cref{rem:planar_residual_equivalence} the angle gradient is $2\pi\,\partial_{\theta_k}\LC=s_k\mathcal F_C'(\theta_k)$, where $\mathcal F_C'(\gamma)=\sum_l s_l\Phi_C'(\gamma-\theta_l)-\sum_m t_m\Phi_C'(\gamma-\beta_m)$ depends on $\theta_l$ only through its $l$-th summand. Differentiating once more, for $k\ne l$ the only $\theta_l$-dependent summand is $s_l\Phi_C'(\theta_k-\theta_l)$, so $2\pi\,\partial^2_{\theta_k\theta_l}\LC=-s_ks_l\,\Phi_C''(\theta_k-\theta_l)$, while for $k=l$ the chain rule gives
\begin{align}
 2\pi\,\partial^2_{\theta_k\theta_k}\LC
    &=\frac{d}{d\theta_k}\Bigl[s_k\,\mathcal F_C'(\theta_k)\Bigr] \notag\\
    &=s_k\,\frac{d}{d\theta_k}\Bigl[\sum_{l\ne k}s_l\,\Phi_C'(\theta_k-\theta_l)+s_k\,\Phi_C'(0)-\sum_m t_m\,\Phi_C'(\theta_k-\beta_m)\Bigr]\notag\\
    &=s_k\Bigl[\sum_{l\ne k}s_l\,\Phi_C''(\theta_k-\theta_l)\;-\;\sum_m t_m\,\Phi_C''(\theta_k-\beta_m)\Bigr]\notag\\
    &=s_k\bigl[\mathcal F_C''(\theta_k)-s_k\,\Phi_C''(0)\bigr].
    \label{eq:diagonal-angle-hessian}
\end{align}
Using this, let us calculate
\begin{align}
    2\pi\,D^2\LC(s,\theta)[z,z]
        &=2\pi\sum_{k,l}\partial^2_{\theta_k\theta_l}\LC\;\delta\theta_k\,\delta\theta_l \notag\\
        &=2\pi\Bigl[\partial^2_{\theta_i\theta_i}\LC\,(\delta\theta_i)^2+2\,\partial^2_{\theta_i\theta_j}\LC\,\delta\theta_i\,\delta\theta_j+\partial^2_{\theta_j\theta_j}\LC\,(\delta\theta_j)^2\Bigr] \notag\\
        &=s_i\bigl[\mathcal F_C''(\alpha)-s_i\Phi_C''(0)\bigr]\,s_j^2 \;+\;2\bigl(-s_is_j\Phi_C''(0)\bigr)\,s_j(-s_i)\notag\\
        &\qquad+\;s_j\bigl[\mathcal F_C''(\alpha)-s_j\Phi_C''(0)\bigr]\,s_i^2 \notag\\
        &=s_is_j(s_i+s_j)\,\mathcal F_C''(\alpha),
    \label{eq:opposed-rotation-computation}
\end{align}
which closes the proof.
\end{ProofSketch}

\begin{lemma}[At most one positive line implies exact fit, \leandecl{Theorems.lean}{347}{PaperLeanFormalization.Theorems.lem_S_is_small}]
\label{lem:S-is-small}
 If $(s,w)\in\Pnn{d}{n}$ with $n\ge m$ is a local minimum of $\LC$ and $|S_+|\le1$, then
it is an exact fit.
\end{lemma}
\begin{Proof}{Lemma}{lem:S-is-small}
If $n=0$, then $m=0$ by $n\ge m$, and the loss is zero. We may therefore assume $n\ge1$. If some student mass is zero, \Cref{prop:zero_mass_exact_fit} gives an exact fit; hence we may also assume positive student masses.

First suppose that there are no lines of positive net mass $S_+=\varnothing$, this is $\omega(\alpha)\le0$ for every
line $\alpha$. Due to \Cref{prop:loss_in_residual} to show zero loss it suffices to show that the centered angular residual \eqref{eq:residual-potential-paper} vanishes everywhere. As by its definition $\mathcal F_C(x)=\sum_{\alpha}\omega(\alpha)\Phi_C(x-\alpha)$ and $\Phi_C>0$ it is easy to see that the residual is negative if $S_- \neq \varnothing$ and otherwise vanishes everywhere. However, as student angles are double zeros by \Cref{lem:critical-points-are-double-zeros}, the former cannot be the case and the residual vanishes everywhere.

Now, suppose there is exactly one line of positive mass $S_+=\{\alpha_0\}$.
Since the teacher masses are positive, and the net mass calculates as aggregated student masses minus aggregated teacher masses, there must be a student direction $w_i$ on the given line $\ell(w_i)=\alpha_0$. By \Cref{lem:critical-points-are-double-zeros}, $\alpha_0$ is a double zero of $\mathcal F_C$.
Moreover, $S_-\neq\varnothing$, because
otherwise
$
\mathcal F_C(\alpha_0)=\omega(\alpha_0)\Phi_C(0)>0,
$
which contradicts criticality. Thus the unique gap of $S_+$ contains a point
of $S_-$. By \Cref{lem:gap-sign}(1),
$
\mathcal F_C(\alpha)<0$ for every line $\alpha\neq\alpha_0$ and since $\mathcal F_C$ vanishes at every student line, all student units lie on $\alpha_0$.

Moreover, \Cref{lem:gap-sign}(2) gives $\mathcal F_C''(\alpha_0)<0$. If $n\ge2$, two positive-mass student units therefore lie on $\alpha_0$, and \Cref{lem:opposed-rotation} supplies a direction of strictly negative curvature, contradicting local minimality.

It remains to consider the case of exactly one student unit $(s_1, \theta_1)$. Since $n\ge m$, we have at most one teacher unit $m\le1$. If $m=0$, the target function is the zero function and the loss is a convex function in the single student unit mass $s_1$ and the only local minimum is where the student unit has mass zero $s_1=0$.

The last case is that of one teacher unit $(t_1, \beta_1)$. On a high level angular criticality forces angle collision or orthogonal angles, but in the latter case masses cannot be optimal. Hence the angles collide and the problem is a convex problem in the masses and any local minimum is an exact fit:
The representation of the loss in \Cref{prop:loss_in_residual} evaluates to
\begin{align}
 \LC(s_1,\theta_1)
 =\frac{s_1^2+t_1^2}{8}-\frac{s_1t_1}{2\pi}\Phi_C(\theta_1 - \beta_1).
 \label{eq:width-one-centered-loss}
\end{align}
Now criticality in the angle and $s_1,t_1>0$ requires $\Phi_C'(\theta_1 - \beta_1) = 0$. Calculating $\Phi_C'(\eta)=-\sin\eta\,\arcsin(\cos\eta)$ implies $\sin\eta=0$ or $\cos\eta=0$. In the latter case, $\Phi_C''(\eta)=1$, so $\partial_{\theta_1}^2\LC=-s_1t_1/(2\pi)<0$, contradicting local minimality. Hence teacher and student unit directions lie on the same line and $\Phi_C(\theta_1 - \beta_1)=\pi/2$.
Mass criticality now gives $0=\partial_{s_1}\LC=(s_1-t_1)/4$, so $s_1=t_1$. We conclude, that the student network is an exact fit.
\end{Proof}

\begin{proposition}[Positive lines are student lines, \leandecl{Theorems.lean}{325}{PaperLeanFormalization.Theorems.prop_S_plus_eq_student_lines}]
\label{prop:S-plus-eq-student-lines}
      If
$(s,w)\in\Ppos{d}{n}$ is a critical point of $\LC$ for the positive teacher network $(t,v)$ and $|S_+| \ge 2$, then $S_+ = \Lambda_w$.
\end{proposition}
\begin{Proof}{Proposition}{prop:S-plus-eq-student-lines}
   Since the student network is a critical point, it follows by \Cref{lem:critical-points-are-double-zeros} that every student line is a double zero of $\mathcal F_C$. Since the masses of the teacher network are positive we have that $S_+ \subseteq \Lambda_w$. Hence, by \Cref{lem:line-count}, every gap of $S_+$ contains a point of $S_-$, and \Cref{lem:gap-sign}(1) gives $
\mathcal F_C(x)<0$
on every gap of $S_+$. This in turn, again by \Cref{lem:critical-points-are-double-zeros} implies that every student line must be contained in $S_+$ and hence $S_+ = \Lambda_w$.
\end{Proof}

\begin{lemma}[Student line count at centered spurious local minima, \leandecl{Theorems.lean}{336}{PaperLeanFormalization.Theorems.lem_no_collisions}]
\label{lem:no-collisions}
If
$(s,w)\in\Ppos{d}{n}$ is a spurious local minimum of $\LC$, then
$
|\Lambda_w|=n.
$
\end{lemma}
\begin{Proof}{Lemma}{lem:no-collisions}
Since $(s,w)$ is a spurious local minimum,
\Cref{lem:S-is-small} gives $|S_+|\ge2$. Hence, by \Cref{lem:line-count}, every gap of $S_+$ contains a point of $S_-$, so \Cref{lem:gap-sign} applies to every gap. Since the masses are positive, by \Cref{lem:gap-sign}(1),
$S_+=\Lambda_w,
$
and therefore by \Cref{lem:gap-sign}(2)
$
\mathcal F_C''(\alpha)<0$ for every $\alpha\in\Lambda_w$.

Assume for the sake of contradiction that there are $w_i,w_j$ with $\ell(w_j)=\ell(w_i)$. Then, since the masses are positive, \Cref{lem:opposed-rotation} applies.
Since $
\mathcal F_C''(\alpha)<0$, the direction constructed in \Cref{lem:opposed-rotation} contradicts the positive semidefiniteness of the Hessian at a local minimum. Hence, all student directions lie on different lines, equivalently, $|\Lambda_w|=n$.
\end{Proof}

We call the student network and the teacher network \emph{line-separated} if the lines are pairwise distinct, equivalently, $n=|\Lambda_w|,m=|\Lambda_v|$ and $\Lambda_w \cap \Lambda_v = \emptyset.$
We say that $S_+$ and $S_-$ \emph{strictly interlace} if every gap of
$S_+$ contains exactly one point of $S_-$. When the student network and the teacher network
are line-separated and have positive masses, we say that they strictly
interlace if $\Lambda_w=S_+$ and $\Lambda_v=S_-$ strictly interlace.
\Cref{fig:interlacing}b shows three teacher and three student lines that
strictly interlace.

\begin{proposition}[Interlacing and Line-separation at centered spurious local minima, \leandecl{Theorems.lean}{358}{PaperLeanFormalization.Theorems.prop_locamin_are_line_separated}]
\label{prop:locamin_are_line_separated}
      If $(s,w) \in\Ppos{d}{n}$ is a spurious local minimum of $\LC$, then $n=m$ and the student network and the teacher network are line-separated and strictly interlace.
\end{proposition}

\begin{Proof}{Proposition}{prop:locamin_are_line_separated}

Since the local minimum is spurious, by \Cref{lem:S-is-small}, we have that $|S_+| \ge 2$. By \Cref{lem:no-collisions}, we have that $|S_+| = |\Lambda_w|=n$ and consequently by \Cref{lem:line-count}, $n =|S_+| \leq |S_-| \leq |\Lambda_v| \leq m$. Since also $n\geq m$, it follows that $n=m$ and that equality holds everywhere in the preceding chain of inequalities. In particular, the student network and the teacher network are line-separated. Equality in \Cref{lem:line-count} implies that every gap
of $S_+$ contains exactly one point of $S_-$, so they strictly
interlace.
\end{Proof}

\begin{proposition}[Descent direction in the matched width case,
\leandecl{Theorems.lean}{370}{PaperLeanFormalization.Theorems.prop_beam_descent}]
\label{prop:beam-descent}
Let the student network and the teacher network be line-separated
and strictly interlacing,  with $n \ge 2$. If $(s,w)\in\Ppos{d}{n}$ is
a critical point of $\LC$, then there exists
$z\in\R^n\times\R^n$ such that
\begin{align}
D^2\LC(s,\theta)[z,z]<0.
\label{eq:beam-descent-direction}
\end{align}
\end{proposition}

\begin{ProofSketch}{Proposition}{prop:beam-descent}
We give explicit rational expressions involving sines and cosines for the construction of the negative-curvature direction, and outline the identities certifying it.
In what follows we focus on first giving the precise definition of the descent direction and give the conceptual ideas and steps used to show negative loss curvature along the constructed direction.

Fix teacher parameters $(t,\beta)$ and student parameters $(s,\theta)$,
which we assume to constitute a critical point of $\LC$, with student
and teacher width $n=m\ge2$, positive masses, and pairwise distinct,
strictly interlacing student and teacher lines.
We may assume that
$\theta_1<\cdots<\theta_n<\theta_{n+1}:=\theta_1+\pi$;
let us index the teacher unit in student gap $(\theta_g,\theta_{g+1})$,
let us call it \emph{gap $g$}, by $\beta_g=\theta_g+p_g$,
and denote the gap length
$l_g:=\theta_{g+1}-\theta_g\in(0,\pi)$, so $p_g\in(0,l_g)$.
Write $z=(\delta s,\delta\theta)$ for a \emph{mass--angle velocity} and
$x_g=s_g(\delta\theta)_g$ for its \emph{mass-weighted angular components}.
Define
\begin{align}
 \Lambda(l)&=l^2-\sin^2l,
 & \psi(l)&=\sin l-l\cos l,\notag\\
 u_g&=-\frac4{\mathcal F_C''(\theta_g)},
 & \varrho_g&=-\frac{\mathcal F_C''(\theta_g)}{s_g}.
 \notag
\end{align}
These divisions are justified. The Green function representation of the centered angular residual in \Cref{lem:gap-sign} and our interlacing assumption show $\mathcal F_C''(\theta_g)<0$ at every student unit.
In particular, $u_g,\varrho_g>0$, while $\Lambda(l_g)>0$ and $s_g>0$
justify the remaining divisions.

Define coefficients
\begin{align}
 a_g&=\frac{2(l_g-\sin l_g\cos l_g)}{\Lambda(l_g)},
 & b_g&=\frac{2\psi(l_g)}{\Lambda(l_g)},\notag\\
 d_g&=\frac{8\sin^2l_g}{\Lambda(l_g)},
 & e_g&=\frac{8l_g\sin l_g}{\Lambda(l_g)}.
 \label{eq:beam-constants}
\end{align}
Now we are ready to give the formulas for the descent direction
$(\delta s,\delta\theta)$:
\begin{align}
 (\delta\theta)_g&=\frac{x_g}{s_g}, \ \text{where} \ x_g=(a_g+a_{g-1})u_g+b_gu_{g+1}+b_{g-1}u_{g-1}
 \label{eq:beam-direction-angle}\\
 (\delta s)_g
 &=\frac14\bigl[(d_g-d_{g-1})u_g
               +e_gu_{g+1}-e_{g-1}u_{g-1}\bigr].
 \label{eq:beam-direction}
\end{align}
The explicit representation of the centered angular residual in \eqref{eq:gap-green-representation}
makes these rational expressions in the gap
lengths, teacher offsets, masses, and their trigonometric data: only the two neighboring gap geometries and student unit masses enter
the descent direction for each student unit.
It can be verified that $a_g,b_g,u_g,s_g>0$-we remark that indeed all student units rotate in the same direction,
$(\delta\theta)_g>0$.

To understand the construction, consider first an arbitrary velocity
$z=(\delta s,\delta\theta)$, and let $\xi(y)$ be the almost-everywhere
first variation of twice the student network output for unit direction input at angle $y$ along $z$.
At criticality, the curvature along $z$ is the non-negative feature energy
minus a term for each student unit, containing mass-weighted angular components and mass-divided curvature of the residual at the specific student angle:
\begin{align}
 2\pi D^2\LC(s,\theta)[z,z]
 &=E-\sum_g\varrho_gx_g^2,
 \label{eq:beam-second-variation}\\
 E&=\int_0^\pi\xi(y)^2\,dy\ge0.
 \label{eq:definition_E}
\end{align}
We show that its shifted restriction
$S_g(y):=2\xi(\theta_g+y+\pi/2)$ is a sinusoid, that is,
an amplitude-scaled and phase-shifted sine function of unit frequency,
and its squared integral gives the gap's contribution to the feature
energy:
\begin{align}
 E&=\int_0^\pi\xi(y)^2\,dy
   =\frac14\sum_g\int_0^{l_g}S_g(y)^2\,dy.
 \label{eq:beam-energy-gapwise}
\end{align}
We choose $z$ as in
\eqref{eq:beam-direction-angle}--\eqref{eq:beam-direction}
so that these squared-sinusoid integrals admit an explicit evaluation
suited to proving negative curvature.
The proof progresses by showing that the chosen descent direction $z$
makes $S_g$ equal to an auxiliary sinusoid $B_g$:
their value and slope jumps agree, so their difference extends to one
global sinusoid.
It is shown that the difference is $\pi$-periodic, whereas every
unit-frequency sinusoid is $\pi$-antiperiodic, and the proof concludes
equality (\leanlink{Beam.lean}{1253}{periodicity argument}).
Evaluating the integrals in \eqref{eq:beam-energy-gapwise} using the formulation of the auxiliary sinusoid $B_g$, the calculation arrives at
\begin{align}
 E&=4\sum_g u_gx_g.
 \label{eq:beam-energy}
\end{align}

Substituting this energy identity into
\eqref{eq:beam-second-variation}, for the specifically constructed descent direction $z$, the proof arrives at
\begin{align}
 2\pi D^2\LC(s,\theta)[z,z]
 &=4\sum_g u_gx_g-\sum_g\varrho_gx_g^2\notag\\
 &=4\left[\sum_g\frac{4u_g^2}{\varrho_g}-\sum_g u_gx_g\right]
   -\sum_g\frac{(\varrho_gx_g-4u_g)^2}{\varrho_g}.
 \label{eq:beam-descent-identity}
\end{align}
The rest of the proof certifies negativity of the given formula and since $\varrho_g>0$, it suffices to prove the
\leanlink{Beam.lean}{3154}{summed inequality}
\begin{align}
 \sum_g\frac{4u_g^2}{\varrho_g}
 &<\sum_g u_gx_g.
 \label{eq:beam-target}
\end{align}
Lean establishes this by
\leanlink{Beam.lean}{2690}{regrouping into gap contributions}
and proving each \leanlink{Beam.lean}{2625}{gap inequality}
through an \leanlink{Beam.lean}{508}{exact trigonometric factorization}
and monotonicity of $\psi/\Lambda$.
The \leanlink{Beam.lean}{3188}{completed-square identity}
then gives strictly negative curvature, closing the proof of \Cref{prop:beam-descent}.
\end{ProofSketch}

\begin{remark}[Scope and presentation of the matched-width certificate]
Proposition~26 is needed only for the matched-width case $n=m$. The
overparameterized case $n>m$ is already excluded by Proposition~25: any
hypothetical positive-loss local minimum must satisfy
\[
n = |\Lambda_w|
  = |S_+|
  \le |S_-|
  \le |\Lambda_v|
  \le m,
\]
which contradicts $n>m$. Thus neither the descent construction of
Proposition~26 nor its gap inequality enters the proof in the
overparameterized regime.

The remaining matched-width completion uses an exact but technically
intricate trigonometric certificate specific to the constructed direction.
This certificate arose from AI-assisted formal proof search rather than from
an argument originally designed for concise human exposition. We subsequently
worked through the algebra, reconstructed its logical dependencies, and made
repeated attempts to simplify and translate it into a conventional proof.
Those attempts did not yield an exposition that we considered reasonably
readable or independently checkable at conference-paper length.

We therefore separate exposition from complete verification. The paper gives
the exact negative-curvature direction, the curvature decomposition, the
energy identity, and the precise summed inequality whose strictness implies
negative curvature. The accompanying Lean development supplies the full
derivation of that inequality, including the auxiliary identities and exact
trigonometric calculations omitted from the prose. Every displayed equation
and named proof step has a corresponding Lean declaration.
Our aim is to make the theorem and the structural ideas behind it accessible
to a broad audience without replacing the exposition by a long transcription
of a machine-discovered algebraic certificate.
\end{remark}

\begin{theorem}[Main Result, centered formulation,
\leandecl{Theorems.lean}{397}{PaperLeanFormalization.Theorems.thm_centered}]
\label{thm:centered}
Let $d\ge1$, let the teacher network satisfy $(t,v)\in\Ppos{d}{m}\cap\Ppl{d}{m}$, and let $n\ge m$. Then every $(s,w)\in\Pnn{d}{n}$ that is a local minimum of $\LC$ on $\R^n\times(\sphere^{d-1})^n$ is an exact fit.
\end{theorem}

\begin{Proof}{Theorem}{thm:centered}
For $d=1$, every unit direction is $\pm1$, so every centered unit function equals $|x|/2$. Put $D=\sum_i s_i-\sum_k t_k$. The loss is $\kappa D^2$ with the fixed coefficient $\kappa=\E[|X|^2]/8$; varying one mass by $\tau$ changes it to $\kappa(D+\tau)^2$. Stationarity gives $2\kappa D=0$, hence $\LC=0$. This argument needs no mass-sign assumption (\leanlinkdecl{Geometry.lean}{3898}{PaperLeanFormalization.Geometry.centered_local_min_dimension_one_zero}{Lean}). If empty networks are allowed, $n=0$ forces $m=0$ and the centered loss vanishes identically.

Now let $d\ge2$ and let $(s,w)\in\Pnn{d}{n}$ be a local minimum. If some student mass is zero, the parameter is an exact fit by \Cref{prop:zero_mass_exact_fit}, whose planar case applies since the teacher masses are positive. Otherwise all masses are positive, and \Cref{prop:teacher-span-confinement} places every student direction in the teacher span $\mathcal T$, which has dimension at most two. Hence all student and teacher directions lie in a common two-dimensional subspace $P\subseteq\R^d$. Restricting the perturbations to parameters with directions in $P$ preserves local minimality, and since every feature $\varphi_C(\ip{w}{x})$ with $w\in P$ depends on $x$ only through its orthogonal projection onto $P$, which is standard Gaussian in $P$, the loss of such parameters equals the planar loss of the same teacher network and student network in $P\cong\R^2$. This reduces the claim to $d=2$. Assume, for the sake of contradiction, that this planar local minimum is spurious. By \Cref{prop:locamin_are_line_separated}, the student network and the teacher network are line-separated with $n=m$ and strictly interlace, so \Cref{prop:beam-descent} supplies a direction of strictly negative second variation, contradicting local minimality.
\end{Proof}
The loss decomposition in \Cref{lem:equivalence_loss_skip_centered} gives the main result for $d\ge2$; a direct residual-scaling argument covers $d=1$.

\begin{ProofSketch}{Theorem}{thm:headline}
For $d\ge2$, apply \Cref{lem:equivalence_loss_skip_centered} and \Cref{thm:centered}. For $d=1$, keep the student directions fixed and choose masses with total $\sum_k t_k$ and their matching linear skip. This is a pointwise exact fit, since all centered unit functions equal $|x|/2$. Interpolating the masses and linear skip linearly from the candidate to this fit multiplies the residual by $1-\tau$, and the loss by $(1-\tau)^2$. Stationarity at $\tau=0$ forces the original loss to vanish (\leanlinkdecl{Geometry.lean}{5174}{PaperLeanFormalization.Geometry.skip_local_min_dimension_one_zero}{Lean}). For $n=m=0$, interpolate just the linear skip to $b^{\mathbf t}$.
\end{ProofSketch}

The hypothesis $d\ge3$ of \Cref{prop:nc-dead-student} cannot be dropped: \Cref{lem:nc-dead-counterexamples} exhibits spurious local minima with a zero-mass unit in dimensions $1$ and $2$, the planar one for a teacher network with negative masses. Whether \Cref{prop:nc-dead-student} holds for $d=2$ and non-negative teacher masses, as \Cref{prop:zero_mass_exact_fit} does for the centered loss, is open.

\begin{lemma}[Zero-mass spurious minima in $d\le2$, \leandecl{Theorems.lean}{228}{PaperLeanFormalization.Theorems.lem_nc_dead_counterexamples}]
\label{lem:nc-dead-counterexamples}
There are spurious local minima of $\LR$ with a positive-mass teacher unit and a zero-mass student unit in $d=1$, and spurious local minima of $\LR$ and $\LC$ for a teacher network with negative-mass units and a student network consisting of a single zero-mass unit in $d=2$.
\begin{enumerate}
\item[(a)] In $d=1$ for the one-unit teacher network $(t,v)=(1,v)$ and one student unit at $w=-v$ with mass $s$, $\LR(s,-v)=\tfrac14(s^2+1)$, so $(0,-v)$ is a local minimum of $\LR$ with loss $\tfrac14$, whereas $(1,v)$ has loss $0$.
\item[(b)] In $d=2$ take the teacher network with angles $\beta=(\pi/6,\pi/3,\pi/2,2\pi/3,5\pi/6)$ and masses $t=(1,-2\sqrt3,5,-2\sqrt3,1)$, and one student unit of mass $0$ at angle $\tau$. For every $|\tau|<\pi/6$ the parameter $(0,e(\tau))$ is a local minimum of $\LR$ and of $\LC$ with positive loss.
\end{enumerate}
\end{lemma}

\begin{ProofSketch}{Lemma}{lem:nc-dead-counterexamples}
(a) In $d=1$ let us observe $\ip{\pm v}{X}=\pm X$, $[X]_+[-X]_+=0$ and $\E[X]_+^2=\E[-X]_+^2=\tfrac12$. This gives $\LR(s,-v)=\tfrac12\E[(s[-X]_+-[X]_+)^2]=\tfrac14(s^2+1)$. Since $\sphere^0=\{\pm v\}$ is discrete, a neighborhood of $(0,-v)$ only varies $s$, and $s=0$ is the minimizer.

(b) For one student unit of mass $s$ at angle $\theta$, \Cref{prop:loss_in_residual} gives the exact quadratic
\begin{align}
 \mathscr L_q(s,\theta)-\mathscr L_q(0,\theta)=\kappa_q s^2-\frac{s}{2\pi}P_q(\theta),\qquad
 P_q(\theta)=\sum_kt_k\Phi_q(\theta-\beta_k)
 \label{eq:one-student-quadratic}
\end{align}
with $P_q$ the teacher potential, and $\kappa_q=\widetilde\Phi_q(1)/(4\pi)$ resulting in $\kappa_C=\tfrac18$ and $\kappa_R=\tfrac14$. Since a zero-mass student unit contributes nothing, $\mathscr L_q(0,\theta)=\tfrac12\E[f_{q;t,v}(X)^2]$ does not depend on $\theta$. Let us show that for all $s$ and all $|\theta|<\pi/6$ the loss is
\begin{align}
    \mathscr L_q(s,\theta) = \mathscr L_q(0,\tau)+\kappa_qs^2,
    \label{eq:one-student-arc}
\end{align}
so that on the neighborhood $\{|\theta|<\pi/6\}$ of $(0,\tau)$ the loss is at least $\mathscr L_q(0,\tau)$, with equality exactly on $s=0$: $(0,e(\tau))$ is a non-strict local minimum in mass and angle together. By the quadratic, the claim is $P_q\equiv0$ on $|\theta|<\pi/6$. Let us first show this for $q=R$. There every $\theta-\beta_k$ lies in $(-\pi,0)$, where $\arcsin(\cos z)=\frac\pi2+z$ and $\sqrt{1-\cos^2z}=-\sin z$ give $\Phi_R(z)=(\pi+z)\cos z-\sin z$. Expanding $\cos(\theta-\beta_k)$ and $\sin(\theta-\beta_k)$ by the addition theorems,
\begin{align}
 P_R(\theta)=(\pi+\theta)\bigl(A\cos\theta+B\sin\theta\bigr)-\bigl(C\cos\theta+D\sin\theta\bigr)-\bigl(A\sin\theta-B\cos\theta\bigr),
 \label{eq:teacher-potential-expansion}
\end{align}
with $A=\sum_kt_k\cos\beta_k$, $B=\sum_kt_k\sin\beta_k$, $C=\sum_kt_k\beta_k\cos\beta_k$, $D=\sum_kt_k\beta_k\sin\beta_k$. For the displayed teacher network all four vanish for the given teacher angles and masses---the simple calculation uses $\cos\frac\pi6=\frac{\sqrt3}2$, $\cos\frac\pi3=\frac12$ and their companions.
To show $P_q\equiv0$ on the arc for $q=C$ observe that $\Phi_C=\Phi_R-\frac\pi2\cos$ and the argument runs through analogously.

The minimum is not global: as $\mathscr L_q(0,\tau)=\tfrac12\E[f_{q;t,v}(X)^2]>0$, the teacher function is non-zero, and since $P_q(\theta)=2\pi\E[f_{q;t,v}(X)\varphi_q(\ip{e(\theta)}{X})]$ by \Cref{lem:pair-moments}, $P_q\equiv0$ would give $\E[f_{q;t,v}(X)^2]=\frac1{2\pi}\sum_kt_kP_q(\beta_k)=0$. Hence $P_q(\theta)\ne0$ for some angle $\theta$, and by \eqref{eq:one-student-quadratic} a small mass $s$ of the sign of $P_q(\theta)$ at this angle lowers the loss below $\mathscr L_q(0,\tau)$.
\end{ProofSketch}

\subsection{Spurious Local Minima for ReLU}
\label{app:plain-proofs}
In this subsection we give the proof ideas of \Cref{thm:plain-trap} and \Cref{prop:plain-collision-ceiling} of \Cref{sec:plain}, preceded by the box certificate and the splitting lemma on which the trap rests.

Spurious local minima of this population loss were first established by \citet{SafranShamir2018}, for teacher networks with orthonormal directions of width $k\ge6$ at matched student width, by a computer-assisted proof of the following shape. Gradient descent proposes a candidate $w$; certificates using variable-precision floating-point arithmetic for a small gradient $\|\nabla F(w)\|\le\varepsilon$, a Hessian floor $\lambda_{\min}(\nabla^2F(w))\ge\lambda_{\min}>0$, and a uniform bound $B$ on the third-order Taylor remainder are combined to $F(w+\tau u)\ge F(w)+\tau\bigl(\tfrac{\lambda_{\min}}2\tau-\tfrac B6\tau^2-\varepsilon\bigr)>F(w)$ for unit $u$ and $\tau$ in an interval: a small closed ball around $w$ has values above $F(w)$ and contains an interior local minimum; a Lipschitz bound on $F$ and $F(w)>0$ show it is not global. We follow the same paradigm with three changes. First, the Hessian is bounded below on a whole box, so no third-derivative remainder enters: strong convexity on the box gives a single and strict interior local minimum. Second, every enclosure is exact rational interval arithmetic checked in Lean, the Hessian floor is certified as a sum-of-squares identity, rather than floating-point arithmetic. Third, positive loss follows from a conceptual rather than numerical argument analyzing the differentiability of the student network and the teacher network on a certain input line.

\begin{lemma}[Box certificate for a strict minimum; \leandecl{Theorems.lean}{428}{PaperLeanFormalization.Theorems.lem_box_certificate}]
\label{lem:box-certificate}
Let $\ell:\R^D\to\R$ be $C^2$, let $r,\lambda>0$ and $g\ge0$, and put $B=\{z:\|z-z_0\|_\infty\le r\}$. Suppose
\begin{align}
 D^2\ell(z)[h,h]\ge\lambda\|h\|_2^2\quad(z\in B,\ h\in\R^D),
 \qquad \|\nabla\ell(z_0)\|_2\le g,
 \qquad 2g<\lambda r.
 \label{eq:box-hypotheses}
\end{align}
Then $\ell$ has a unique minimizer $z_\star$ on $B$, this point lies in the interior of $B$, and it is a strict local minimum.
\end{lemma}

\begin{ProofSketch}{Lemma}{lem:box-certificate}
For $z,z+v\in B$, the segment function $\varphi(\tau)=\ell(z+\tau v)$ satisfies $\varphi''(\tau)\ge\lambda\|v\|_2^2$ on $[0,1]$. Integrating twice gives
\begin{align}
 \ell(z+v)-\ell(z)-D\ell(z)[v]
 &=\int_0^1\!\bigl(\varphi'(\tau)-\varphi'(0)\bigr)\,d\tau \notag\\
 &=\int_0^1\!\!\int_0^\tau\varphi''(\sigma)\,d\sigma\,d\tau \notag\\
 &\ge\lambda\|v\|_2^2\int_0^1\tau\,d\tau \notag\\
 &=\frac\lambda2\|v\|_2^2.
 \label{eq:trap-taylor}
\end{align}
Pulling $D \ell(z)[v]$ to the right and applying Cauchy--Schwarz yields
\begin{align}
 \ell(z_0+v)-\ell(z_0)
 \ge\|v\|_2\left(\frac\lambda2\|v\|_2-g\right).
 \label{eq:box-boundary-gap}
\end{align}
At a boundary point $z_0+v$ with $\|v\|_2\ge r$ this is positive by the assumption $2g < \lambda r$.
Compactness therefore gives an interior minimizer $z_\star$, where $D\ell(z_\star)=0$. Applying \eqref{eq:trap-taylor} there gives strict comparison with every other point of $B$, proving uniqueness and strict local minimality.
\end{ProofSketch}

\begin{lemma}[Positive residual curvature permits finite splitting; \leandecl{Theorems.lean}{465}{PaperLeanFormalization.Theorems.lem_positive_curvature_splitting}]
\label{lem:positive-curvature-splitting}
Fix a planar plain-ReLU teacher network. Let $(s,\theta)$ be a local minimum of $\LR$, and suppose that some student unit $i$ satisfies $s_i>0$ and $\mathcal F_R''(\theta_i)>0$. For any $k\ge1$ and any positive partition
\begin{align}
 s_i=\sum_{j=1}^k s'_j,\qquad s'_j>0,
 \label{eq:positive-partition}
\end{align}
replace that unit by $k$ children with masses $s'_j$, all at angle $\theta_i$, leaving every other unit unchanged. The resulting parameter is a local minimum with the same represented function and loss. For $k\ge2$ it is non-strict: transferring mass between two children gives a non-constant constant-loss curve through it.
\end{lemma}

\begin{ProofSketch}{Lemma}{lem:positive-curvature-splitting}
Merging nearby children will give a loss lower bound through Jensen's inequality. For a fixed teacher network, set the normalized kernel loss $\ell_R(s, \theta)$ to be
\begin{align}
 \ell_R(s,\theta)
 &=\frac12\sum_{i,j}s_is_j\Phi_R(\theta_i-\theta_j)
   -\sum_{i,k}s_it_k\Phi_R(\theta_i-\beta_k)\notag\\
 &=2\pi\LR(s,\theta)-\frac12\sum_{k,l}t_kt_l\Phi_R(\beta_k-\beta_l).
 \label{eq:plain-loss-normalization}
\end{align}
Write $q_\star=(s,\theta)$ for the given width-$n$ local minimum and $p_\star$ for its split: unit $i$ is replaced by $k$ children of masses $s'_1,\dots,s'_k$, all at the angle $\theta_i$, and the other units are copied. The children together are the old unit, so $p_\star$ represents the same function and $\LR(p_\star)=\LR(q_\star)$. We must show $\LR(p)\ge\LR(p_\star)$ for every width-$(n+k-1)$ parameter $p$ near $p_\star$. Near $p_\star$ the child masses $s'_j(p)$ remain positive, and we read the child angles $\vartheta_j(p)$ as real numbers near $\theta_i$. The merge $M(p)$ is the width-$n$ parameter that replaces the children by a single unit of mass $A(p)=\sum_j s'_j(p)$ at the barycentric angle
\begin{align}
 \bar\vartheta(p)=\frac{\sum_j s'_j(p)\vartheta_j(p)}{A(p)},
 \label{eq:child-barycenter}
\end{align}
and copies the other units. It is continuous with $M(p_\star)=q_\star$, so local minimality of $q_\star$ gives $\LR(M(p))\ge\LR(q_\star)$ for $p$ near $p_\star$. It remains to compare $p$ with its merge.

Since $\partial_\gamma^2\mathcal F_R(q_\star,\theta_i)>0$, there are a neighborhood $U$ of $q_\star$ and an open interval $I\ni\theta_i$ with $\partial_\gamma^2\mathcal F_R(q,\gamma)>0$ for all $q\in U$ and $\gamma\in I$; that is, for each fixed $q\in U$ the function $\gamma\mapsto \mathcal F_R(q,\gamma)$ is convex on the same interval $I$. For $p$ near $p_\star$, continuity of $M$ puts $M(p)\in U$, and the child angles $\vartheta_j(p)$, hence their weighted mean $\bar\vartheta(p)$, lie in $I$. Thus $F:=\mathcal F_R(M(p),\cdot)$ is convex on $I$, an interval containing every child angle of $p$.

Expanding the loss square, $\ell_R(p)-\ell_R(M(p))$ equals the non-negative term $\pi\,\E[(f_{R;p}(X)-f_{R;M(p)}(X))^2]$ plus
\begin{align}
 \sum_{j=1}^k s'_j(p)F(\vartheta_j(p))-A(p)F(\bar\vartheta(p))\ge0,
 \label{eq:merge-cross-term}
\end{align}
by Jensen with weights $s'_j(p)/A(p)$. Consequently,
\begin{align}
 \LR(p)\ge\LR(M(p))\ge\LR(q_\star)=\LR(p_\star),
 \label{eq:split-local-comparison}
\end{align}
where the equality follows because the split children represent the original unit. This proves local minimality and preserves the function and loss for any finite partition. For $k\ge2$, the masses $s'_1+\tau,s'_2-\tau$ remain positive for small $|\tau|$ and define a non-constant constant-loss curve, proving non-strictness.
\end{ProofSketch}

\begin{ProofSketch}{Theorem}{thm:plain-trap}
The box certificate gives a strict seed; positive residual curvature then permits splitting by \Cref{lem:positive-curvature-splitting}.

\pfstep{stp:plain-box}{Verify the box certificate}
Use $\ell_R$ from the splitting proof for the stated teacher network, in coordinate order $(s_1,s_2,s_3,\theta_1,\theta_2,\theta_3)$. Angles and box radii are measured in radians, not in units of $\pi$. The center $x_0=(\bar s,\bar\theta)$ has exact rational masses and angle ratios:
\begin{align}
    \bar s &= \TrapMasses \label{eq:trap-student-masses}\\
    \frac{\bar\theta}{\pi} &= \TrapAnglesOverPi.
    \label{eq:trap-student-angles}
\end{align}
On $B=\{x_0+v:\|v\|_\infty\le r\}$ with $r=\TrapBoxRadius$, we certify curvature, gradient, and positive loss.

(i) \emph{Curvature floor.} Let $H(z)=D^2\ell_R(z)$ and use the upper-triangular rational matrix
\begin{align}
 C=\TrapGramScale\TrapGramMatrix.
 \label{eq:seed-gram-matrix}
\end{align}
For this \leanlinkdecl{Plain.lean}{8167}{PaperLeanFormalization.SeedNumerics.seedGramWitness}{exact matrix}, the \leanlinkdecl{Plain.lean}{8267}{PaperLeanFormalization.SeedNumerics.seed_hessian_gram_enclosure}{entrywise enclosure} is
\begin{align}
 \left|H_{ij}(z)-(C^{\mathsf T}C)_{ij}-\TrapHessianShift\delta_{ij}\right|
 \le\TrapHessianBound\qquad(z\in B,\ 1\le i,j\le6),
 \label{eq:seed-hessian-enclosure}
\end{align}
with the certified form identified as the actual Hessian by \leanlinkdecl{Plain.lean}{8874}{PaperLeanFormalization.Plain.seed_second_eq_fderiv}{the derivative identification}. For $E=H-C^{\mathsf T}C-\frac1{10}I$, Cauchy--Schwarz gives $|v^{\mathsf T}Ev|\le\|v\|_1^2/250\le6\|v\|_2^2/250$. Hence $H\succeq Q:=C^{\mathsf T}C+\TrapShiftMinusSixBound I$, and
\begin{align}
 v^{\mathsf T}\left(Q-\frac1{16}I\right)v
 =\sum_{j=1}^6(Cv)_j^2+\TrapSosRemainder\sum_{i=1}^6v_i^2\ge0
 \label{eq:seed-hessian-sos}
\end{align}
certifies $D^2\ell_R(z)[v,v]\ge\underline\lambda\|v\|_2^2$ with $\underline\lambda=\TrapCurvatureFloorInline$.

(ii) \emph{Gradient bound.} The \leanlinkdecl{Plain.lean}{8203}{PaperLeanFormalization.SeedNumerics.seed_center_gradient_sq}{center-gradient enclosures} bound the six absolute coordinates by $\TrapGradientScale\TrapGradientBounds$; thus $g^2:=\|\nabla\ell_R(x_0)\|_2^2\le873046800\cdot10^{-28}<10^{-19}$.
For (i) and (ii), rational center--radius enclosures propagate through sums, products, and absolute values. Trigonometric argument reduction, degree-$12$ cosine and degree-$13$ sine \leanlinkdecl{Plain.lean}{3382}{PaperLeanFormalization.SeedNumerics.trig_taylor_bound}{Taylor remainder bounds} on $|x|\le1$, and $3.141592653589<\pi<3.14159265359$ enclose the inputs. The \leanlinkdecl{Plain.lean}{3586}{PaperLeanFormalization.SeedNumerics.Certificate.sound}{enclosure soundness proof} turns these finite rational checks into inequalities for the real derivatives.

(iii) \emph{Positive loss.} For every parameter in $B$ the student angles lie in $(\pi,2\pi)$, $(\pi,2\pi)$, and $(0,\pi)$, so $\sin\theta_i\ne0$ for all $i$. On the input line $x=(\xi,1)$ a unit at angle $\gamma$ contributes $[\sin\gamma+\xi\cos\gamma]_+$. The three student units and the teacher units at $5\pi/6$ and $4\pi/3$ have $\sin\gamma\ne0$ and are therefore differentiable in $\xi$ at $\xi=0$, whereas the teacher unit at angle $0$ contributes the kink $[\xi]_+$. Hence the two networks differ at some point of this line (\leanlinkdecl{Plain.lean}{93}{PaperLeanFormalization.SeedCertificate.finite_relu_cannot_equal_kink}{no finite differentiable sum equals a kink}); both are continuous and the Gaussian density is positive, so their squared difference is positive on a non-empty open set of positive measure. Thus
\begin{align}
    \LR(z)>0 \qquad (z\in B).
    \label{eq:trap-positive-loss}
\end{align}

\emph{Conclusion.} The finite kernel expression $\ell_R$ is $C^2$ because $\Phi_R$ is, and $2g<\underline\lambda r$ since $4g^2<4\cdot10^{-19}<(1/800000)^2=(\underline\lambda r)^2$. \Cref{lem:box-certificate} therefore gives a strict local minimum $z_\star$ of $\ell_R$ in the interior of $B$. In angle coordinates $\LR=\ell_R/(2\pi)+\text{const}$, and $\theta\mapsto e(\theta)$ is a local chart of the circle, so $z_\star$ is a strict local minimum of $\LR$ in mass--direction coordinates. Its masses are positive, since every point of $B$ has masses above $0.6$, and it is spurious, since $\LR(z_\star)>0$ by (iii) while the teacher network itself has loss zero.

\pfstep{stp:plain-curvature}{Certify positive residual curvature}
For splitting the third student unit of $z_\star=(s_\star,\theta_\star)$, we certify $\mathcal F_R''(\theta_{\star,3})>1/8$, although only positivity is needed. The identity $\Phi_R''+\Phi_R=2|\sin\xi|$ gives
\begin{align}
    \mathcal F_R''(\gamma)=2\Bigl[\sum_is_i\,|\sin(\gamma-\theta_i)|-\sum_kt_k\,|\sin(\gamma-\beta_k)|\Bigr]-\mathcal F_R(\gamma).
    \label{eq:sine-load-identity}
\end{align}
At $\gamma=\theta_{\star,3}$ the third student term vanishes, as does $\mathcal F_R(\theta_{\star,3})$ by criticality. Throughout $B$, the angle sectors give $|\sin(\theta_3-\theta_1)|\ge24/25$, $|\sin(\theta_3-\theta_2)|\ge163/200$, $|\sin\theta_3|\le37/40$, and $|\sin(\theta_3-4\pi/3)|\le1/10$. With $s_1\ge999/1000$, $s_2\ge1389/1000$, and $|\sin(\theta_3-5\pi/6)|\le1$, the lower bound is
\begin{align}
 2\left(\frac{999}{1000}\frac{24}{25}
       +\frac{1389}{1000}\frac{163}{200}
       -\frac{37}{40}-1-\frac1{10}\right)
 =\frac{2643}{20000}>\frac18,
 \label{eq:seed-curvature-margin}
\end{align}
so
\begin{align}
    \mathcal F_R''(\theta_{\star,3})>\frac18.
    \label{eq:trap-seed-curvature}
\end{align}

\pfstep{stp:plain-splitting}{Apply finite splitting}
For $n>3$, \Cref{lem:positive-curvature-splitting} splits the third student unit into $n-2$ children of mass $s_{\star,3}/(n-2)$ at its angle, leaving the other two unchanged. This preserves positive masses and positive loss and supplies a constant-loss mass-transfer curve. The padded teacher copy fits exactly, so the minimum is spurious and non-strict. For $n=3$, use the strict seed from Step~\ref{stp:plain-box}.
\end{ProofSketch}

\begin{ProofSketch}{Proposition}{prop:plain-collision-ceiling}

As we work in input dimension $d=2$, we are working with angles instead of directions. Fix teacher and student weight-angle parameter pairs $(t,\beta)$ and $(s, \theta)$ of widths $m$ and $n$ respectively such that all student masses are positive and assume this constitutes a critical point of the plain loss $\LR$. First, we aim to show that there are at most $2m$ pairwise distinct student angles.

Let us merge student units that share a direction; this changes neither the represented function nor criticality, so we may assume the $n$ student angles to be pairwise distinct and have to show $n\le2m$. For $n=1$ this is immediate, as the double zero $\mathcal F_R(\theta_1)=0$ forces a teacher unit, so let $n\ge2$. Unlike in the centered case, plain units with opposite directions compute different functions, so we order the student angles on the full circle $\R/2\pi\Z$, $\theta_1<\dots<\theta_n<\theta_{n+1}:=\theta_1+2\pi$, and call $[\theta_j,\theta_{j+1})$ the \emph{gap} of student unit $j$; a gap is shorter than $2\pi$ but may be longer than $\pi$. The \emph{test interval} of student unit $j$ is the closed gap if the gap has length at most $\pi$, and otherwise its middle piece $[\theta_{j+1}-\pi,\theta_j+\pi]$, which has length less than $\pi$ and lies strictly inside the gap.

The proof studies the \emph{sine load} $\mathcal W=\tfrac12(\mathcal F_R''+\mathcal F_R)$, where $\mathcal F_R$ is the plain angular residual \eqref{eq:residual-potential-paper}, which by the kernel identity \eqref{eq:angular-kernel-d2+1} is simply $\sum_is_i|\sin(\cdot-\theta_i)|-\sum_kt_k|\sin(\cdot-\beta_k)|$. We call the student and teacher angles together with their antipodes the \emph{atoms}, and those on a teacher line the \emph{teacher atoms}; there are at most $2m$ teacher atoms. A \emph{sinusoid} is a function of the form $\gamma\mapsto\alpha\cos\gamma+\beta\sin\gamma$; sinusoids form a two-dimensional space, and a sinusoid is determined by its value and its slope at any single point. The sine load is $\pi$-periodic and a sinusoid between consecutive atoms, and at an atom $\tau$ its slope jumps by
\begin{align}
    \mathcal W'(\tau^+)-\mathcal W'(\tau^-)=2\omega(\ell(\tau)),
    \label{eq:sine-load-jump}
\end{align}
with the net mass $\omega$ of \eqref{eq:net-weight-main}: upward where the student mass on the line dominates, downward where the teacher mass does.

The goal is the following count that we achieve via a combinatorial argument about \emph{positive humps} of $\mathcal W$, the maximal intervals where $\mathcal W>0$. We show that in every student gap $[\theta_i, \theta_{i+1})$ either a positive hump ends, and every such hump contains a teacher atom with $\mathcal W>0$, or the student angle $\theta_i$ itself is a teacher atom with $\mathcal W=0$. As distinct humps are disjoint and each hump ends in exactly one gap, this gives
\begin{align}
    n&=\#\{\text{student units in whose gap a hump ends}\}+\#\{\text{student units in whose gap no hump ends}\}\notag\\
        &\le\#\{\text{teacher atoms with }\mathcal W>0\}+\#\{\text{teacher atoms with }\mathcal W=0\}\le2m .
    \label{eq:ceiling-count}
\end{align}
It remains to explain the two alternatives.

Criticality in the mass and angle of each student unit makes every student angle a double zero of $\mathcal F_R$, as $s_j>0$ (\Cref{rem:planar_residual_equivalence}). Since $2\mathcal W=\mathcal F_R''+\mathcal F_R$, integrating by parts twice over a gap $[\theta_j,\theta_{j+1}]$ between consecutive student angles leaves only boundary terms in $\mathcal F_R$ and $\mathcal F_R'$, which vanish by our criticality assumption, so $\mathcal W$ is orthogonal to every sinusoid on the gap in the following sense
\begin{align}
 \int_{\theta_j}^{\theta_{j+1}}\mathcal W(\gamma)\sin(\gamma-a)\,d\gamma=0\qquad\text{for every phase }a .
 \label{eq:gap-orthogonality}
\end{align}
For a gap longer than $\pi$, its two end pieces $[\theta_j,\theta_{j+1}-\pi]$ and $[\theta_j+\pi,\theta_{j+1}]$ are translates of each other by $\pi$, so their contributions cancel by $\pi$-periodicity of $\mathcal W$. Hence $\mathcal W$ is orthogonal to every sinusoid on each test interval. On an interval of length at most $\pi$, such a function either vanishes or changes sign at least twice: if it changed sign only once, at $\xi_0$, then $\mathcal W(\gamma)\sin(\gamma-\xi_0)$ would have one sign on the interval and a non-zero integral. Hence, for every student unit, either $\mathcal W$ vanishes on its test interval or a positive hump of $\mathcal W$ ends inside the test interval, and thus inside the gap.

Both alternatives now come from the way $\mathcal W$ bends at atoms: as student masses are positive, only teacher atoms can bend it downward. Positive humps are shorter than $\pi$, since a hump of length at least $\pi$ would, by $\pi$-periodicity, make $\mathcal W$ positive everywhere up to isolated zeros, while $\mathcal W$ vanishes or changes sign on every test interval. A positive hump must bend downward somewhere to return to zero, since a sinusoid that only bends upward stays positive for half a period, longer than any hump; hence every hump contains a teacher atom with $\mathcal W>0$. If instead no hump ends in the gap of student unit $j$, then $\mathcal W$ vanishes on its test interval and cannot bend upward at $\theta_j$, so the teacher mass on this line is at least the student mass there, which is at least $s_j>0$. Here positivity of $s_j$ is what forces a teacher onto the line, and $\theta_j$ is itself a teacher atom, with $\mathcal W(\theta_j)=0$.

The bound is attained for odd $m\ge3$: teacher units of mass $2$ at the vertices of a regular $m$-gon and student units of mass $1$ at these vertices and their antipodes fit exactly, because $2[z]_+=[z]_++[-z]_++z$ and the vertices sum to zero.
\finishes{Proposition}{prop:plain-collision-ceiling}
\end{ProofSketch}

\subsection{Finite sample consequence}
\label{app:transfer-proofs}
In this subsection we give the proof of \Cref{thm:fixed-radius-bridge} of \Cref{sec:boundaries} with the definitions, the sampling remark, and the minimizer stability theorem it uses.

In the mismatch coordinate $u=b^{\mathbf s}-b_{\mathrm{match}}(s,w)$ of \Cref{lem:equivalence_loss_skip_centered}, the identity $[z]_+=(|z|+z)/2$ and teacher-generated labels write the empirical loss \eqref{eq:empirical-loss} as
\begin{align}
 \widehat L_{\mathcal D}(b^{\mathbf s},s,w)=\frac1{2N}\sum_{j=1}^N\bigl[f_{C;s,w}(x_j)-f_{C;t,v}(x_j)+\ip u{x_j}\bigr]^2 .
 \label{eq:empirical-loss-mismatch}
\end{align}
For $p=(b^{\mathbf s},s,w)$ and $p'=(b^{\mathbf s'},s',w')$, with the corresponding mismatch coordinates $u,u'$, the max-product distance of \Cref{sec:boundaries} is
\begin{align}
 d_J(p,p')=\max\left\{\|s-s'\|_\infty,\ \max_{i\in[n]}\|w_i-w_i'\|_2,\ \|u-u'\|_2\right\}.
 \label{eq:max-product-distance}
\end{align}
and the closed ball of an $r$-stable local minimum is $B_J(p,r)=\{p'\in\R^d\times\R^n\times(\sphere^{d-1})^n:d_J(p,p')\le r\}$.

\begin{remark}[\leandecl{Theorems.lean}{613}{PaperLeanFormalization.Theorems.rem_finite_data_accuracy}]
\label{rem:finite-data-accuracy}
Let $d\ge2$. For independent, identically distributed inputs from an orthogonally invariant law, i.e.\ a distribution invariant under orthogonal transformations of $\R^d$, with finite positive second moment, the accuracy condition \eqref{eq:modulus} holds with arbitrarily high probability for all sufficiently large sample sizes (pairwise independence suffices, as the Lean statement records). Indeed, after normalization by its Lipschitz constant, every test function in \eqref{eq:modulus} belongs to the class $\mathcal G$ of functions $x\mapsto\|x\|^2h(x/\|x\|)$ with $h$ ranging over the $1$-Lipschitz functions on the sphere bounded by one. This index set is compact in the uniform norm by Arzel\`a--Ascoli, $h\mapsto\|x\|^2h(x/\|x\|)$ is continuous for every $x$, and $\|x\|^2$ is an integrable envelope, so the bracketing numbers of $\mathcal G$, the number of pairs $l\le u$ with $\E[u(X)-l(X)]\le\varepsilon$ needed to sandwich every member of $\mathcal G$ between some $l$ and $u$, are finite for every $\varepsilon>0$, and the Glivenko--Cantelli theorem under bracketing gives almost sure convergence of the sample averages, uniformly over $\mathcal G$ \citep[Theorem~19.4 and Example~19.8]{vanderVaart1998}. Since the distribution is orthogonally invariant, the weighted direction $Z/\|Z\|$ under the weight $\|Z\|^2$ is uniform on the sphere, so $\E[\|Z\|^2h(Z/\|Z\|)]=\E\|Z\|^2\int h\,d\sigma$, and for the Gaussian $G$ likewise $\E H(G)=d\int h\,d\sigma$ with $h=H|_{\sphere^{d-1}}$; as also $R_2(\mathcal D)\to\E\|Z\|^2$, the uniform convergence yields $\frac dN\sum_jH(Z_j)-R_2(\mathcal D)\,\E H(G)\to0$ uniformly over the normalized tests, which is \eqref{eq:modulus}. The accuracy statement needs $d\ge2$; for every $d\ge1$ the same event makes every qualifying fixed-radius candidate an exact fit of the sampled data (\leanlinkdecl{Theorems.lean}{631}{PaperLeanFormalization.Theorems.rem_finite_data_accuracy_high_probability}{Lean}), the event being the whole sample space when $d=1$.
\end{remark}

By \Cref{thm:headline}, every population local minimum with non-negative masses is a pointwise exact fit. To transfer this conclusion to accurate datasets, we use the following stability fact.

\begin{theorem}[{\citealp[Assumption~A.4 and Theorem~4.1]{BastinCirilloToint2006}}, \leandecl{Theorems.lean}{581}{PaperLeanFormalization.Theorems.thm_minimizer_stability}]
\label{thm:minimizer-stability}
Let $\ell$ be a continuous function on a metric space, let $K$ be a compact subset, let $r>0$, and let $\ell_j\to\ell$ uniformly on the union $K_r$ of the closed balls of radius $r$ about the points of $K$. If each member of a sequence in $K$ minimizes the corresponding $\ell_j$ on the closed ball of radius $r$ about itself, then every cluster point of the sequence, that is, every limit of a subsequence, is a local minimum of $\ell$.
\end{theorem}
\begin{Proof}{Theorem}{thm:minimizer-stability}
Let $p$ be a cluster point; pass to a subsequence $p_j\to p$ and relabel the sequence $l_j$ accordingly. Since $K$ is closed, $p\in K$. Let $y$ with $d(y,p)<r/2$. For all large $j$, $d(p_j,p)<r/2$, hence $y$ lies in the closed $r$-ball about $p_j$ and $\ell_j(p_j)\le\ell_j(y)$. Both $p_j$ and $y$ lie in $K_r$, so uniform convergence on $K_r$ and the continuity of $\ell$ give $\ell(p)\le\ell(y)$ in the limit. Thus $p$ minimizes $\ell$ on the open $r/2$-ball about itself.
\end{Proof}

The radius must be fixed before the approximation is refined: the counterexample in \citet{BastinCirilloToint2006} of minima with shrinking neighborhoods converging to a saddle shows that this is not an artifact. \citet[Theorem~2.3(b)]{Polak1993} has the epi-convergence version for uniformly strict local minima, and \citet[Theorems~1 and~2]{MeiBaiMontanari2018} the smooth version under gradient and Hessian control and a Morse population risk, neither of which we have.

\begin{ProofSketch}{Theorem}{thm:fixed-radius-bridge}
For $d=1$ every unit direction is $\pm1$, so a student network with the fixed directions, masses of total $\sum_kt_k$, and the matching linear skip fits every input exactly, as in the proof of \Cref{thm:headline}. Moving masses and mismatch linear skip linearly from the candidate towards this fit multiplies every empirical residual by $1-\tau$ and stays in $B_J(p,r)$ for small $\tau>0$; minimality on the ball gives $\widehat L_{\mathcal D}(p)\le(1-\tau)^2\widehat L_{\mathcal D}(p)$, hence zero loss (\leanlinkdecl{Empirical.lean}{1516}{PaperLeanFormalization.Empirical.data_fixed_radius_dimension_one}{Lean}). Let now $d\ge2$.
If $R_2(\mathcal D)=0$, every input and teacher-generated label is zero, so the empirical loss is zero at every parameter. We may therefore assume $R_2(\mathcal D)>0$. Multiplication by $d/R_2(\mathcal D)>0$ preserves all ball minima of the actual empirical loss. We compare this rescaled objective with $\Lskip$ in the coordinates $(s,w,u)$ with $u=b^{\mathbf s}-b_{\rm match}(s,w)$: since $b_{\rm match}$ is continuous, this change of coordinates is a homeomorphism with inverse $(s,w,u)\mapsto(u+b_{\rm match}(s,w),s,w)$, and $d_J$ becomes the maximum of the three distances in $s$, $w$ and $u$, so the balls $B_J(p,r)$ are balls of this product metric.

\pfstep{stp:finite-data-compactness}{parameters of $r$-stable local minima lie in compact set.}
Let $p=(b^{\mathbf s},s,w)$ be $r$-stable with $s_i\ge0$, and put $C=\sum_i s_i$, $S=\sum_k|t_k|=\sum_k t_k$, $U=\|u\|$, $A(x)=f_{C;s,w}(x)+\ip ux$, and $T(x)=f_{C;t,v}(x)$. Scaling masses and mismatch linear skip together, $(s,u)\mapsto(as,au)$, stays in $B_J(p,r)$ for $a$ sufficiently close to $1$ and changes the residual to $aA-T$. Differentiating this finite quadratic at its minimum $a=1$ gives
\begin{align}
 \frac1N\sum_j A(x_j)^2
 &=\frac1N\sum_j A(x_j)T(x_j).
 \label{eq:data-scaling}
\end{align}
Substituting this identity into the squared residual gives
\begin{align}
 \widehat L_{\mathcal D}(p)
 =\frac12\left(\frac1N\sum_j T(x_j)^2-\frac1N\sum_j A(x_j)^2\right).
 \label{eq:data-scaling-loss}
\end{align}
Non-negativity of the loss therefore implies
\begin{align}
 \frac1N\sum_j A(x_j)^2
 \le\frac1N\sum_j T(x_j)^2
 \le\frac{R_2(\mathcal D)S^2}{4}.
 \label{eq:data-teacher-bound}
\end{align}
If $m=0$, this already forces zero empirical loss. Otherwise $n\ge m\ge1$, so the following denominators are positive. For the Gaussian moment $P=\E A(G)^2$, the even centered student network is orthogonal to the odd linear skip, and non-negative masses prevent cancellation between centered units. Hence
\begin{align}
 P\ge\frac{C^2}{4n}+U^2.
 \label{eq:data-coercivity}
\end{align}
The function $A^2$ is continuous and homogeneous of degree two. On the unit ball its Lipschitz constant is at most $2(C/2+U)^2\le4nP$. Applying \eqref{eq:modulus} with this test and $\delta\le1/(8dn)$ therefore gives
\begin{align}
 P\le\frac{2d}{R_2(\mathcal D)}\,\frac1N\sum_j A(x_j)^2\le\frac{dS^2}{2}.
 \label{eq:data-moment-absorption}
\end{align}
Consequently,
\begin{align}
 C^2\le2dnS^2,\qquad U^2\le\frac{dS^2}{2}.
 \label{eq:data-parameter-bounds}
\end{align}
Since $0\le s_i\le C$ and all directions are unit vectors, these bounds place the minima in one compact parameter set $K$, independent of the dataset and its radial scale.

\pfstep{stp:finite-data-uniform}{Uniform comparison on the enlarged set}
On the union $K_r$ of closed $d_J$-balls of radius $r$ about $K$, all masses, now possibly negative, and mismatch linear skips are uniformly bounded. Thus every residual $F_{b^{\mathbf s},s,w}-F_{b^{\mathbf t},t,v}$ is bounded by one constant $L$ and is $L$-Lipschitz on the unit ball. Its squared half-residual is continuous, homogeneous of degree two, and $L^2$-Lipschitz there. Applying \eqref{eq:modulus} gives, uniformly on $K_r$,
\begin{align}
 \left|\frac d{R_2(\mathcal D)}\widehat L_{\mathcal D}-\Lskip\right|
 \le dL^2\delta.
 \label{eq:uniform-comparison}
\end{align}

\pfstep{stp:finite-data-stability}{A common zero rules out positive loss}
If the theorem failed for some fixed $r$, there would be datasets $\mathcal D_j$ of accuracy at most $1/j$ and $r$-stable minima $p_j$ of positive empirical loss. Their second moments are positive, and, after discarding finitely many indices, the compactness step places $p_j$ in $K$. Pass to a convergent subsequence. The rescaled objectives $d\widehat L_{\mathcal D_j}/R_2(\mathcal D_j)$ converge uniformly to $\Lskip$ on $K_r$, so \Cref{thm:minimizer-stability} makes the limit $p=(b^{\mathbf s},s,w)$ a population local minimum with non-negative masses. By \Cref{thm:headline}, $\Lskip(p)=0$, and since the residual $F_{b^{\mathbf s},s,w}-F_{b^{\mathbf t},t,v}$ is continuous and the Gaussian density is positive, it vanishes pointwise; hence $\widehat L_{\mathcal D_j}(p)=0$ for every dataset in the sequence. For all sufficiently large $j$, the limit lies in $B_J(p_j,r)$, and stability gives $0<\widehat L_{\mathcal D_j}(p_j)\le\widehat L_{\mathcal D_j}(p)=0$, a contradiction.
\finishes{Theorem}{thm:fixed-radius-bridge}
\end{ProofSketch}

\section{Lean Certificates}
\label{app:lean-certificates}
Every theorem, proposition, lemma, displayed equation, and named proof step of this paper has a counterpart in the Lean~4 development at the root of the accompanying repository, which compiles against Mathlib~\citep{mathlib2020} without admitted proofs; \texttt{verify.py} there checks the five contribution theorems against their recorded statement types, and the build certificate confirms that no paper statement uses an axiom beyond propositional extensionality, choice, and quotient soundness. Throughout the paper, an equation number and a step heading are hyperlinks to the Lean declaration that states, respectively proves, that formula or step; the linked declaration may state a formula in the generality its proof needs, for instance without the unit-norm restriction of the display. The map from paper labels to Lean declarations is shipped as \texttt{paper\_tools/links.json}; this appendix and every equation link are generated from it by \texttt{paper\_tools/gen\_lean\_links.py}. \Cref{app:lean-contributions} prints the five contribution theorems as Lean source; \Cref{app:lean-statements} lists every paper statement with its Lean contract and proof.

\subsection{The five contributions in Lean}
\label{app:lean-contributions}
The five theorems of \texttt{Overview\_theorems.lean} are printed below in the order of \Cref{sec:contributions}, each preceded by the definitions from \texttt{Overview\_definitions.lean} that its statement introduces. Every proof in that file is a one-line delegation to the statement index \texttt{Theorems.lean}; the listings omit these proofs and link to them instead. The code is reproduced verbatim, with Lean's Unicode notation typeset as the editor shows it: \lstinline[style=lean]|⟪w i, x⟫_ℝ| is the inner product $\ip{w_i}{x}$, \lstinline[style=lean]|‖x‖| the Euclidean norm, \lstinline[style=lean]|Fin n → ℝ| a real vector indexed by $\{0,\dots,n-1\}$, and \lstinline[style=lean]|∫ x : Vec d, f x| the Lebesgue integral over $\R^d$.

\subsubsection{The model}
Both features, the Gaussian input law, and the two population losses \eqref{eq:q-losses} are fixed once; every contribution is stated in these terms. Masses are signed real numbers: nonnegativity is an assumption at a candidate, never a constraint on its neighborhood. The losses are literal Bochner integrals, whose value is $0$ for a non-integrable integrand, so a reader should check that their integrands are integrable: the loss--kernel identity behind \Cref{lem:pair-moments} and \Cref{prop:loss_in_residual} is proved from \leanlinkdecl{Preliminaries.lean}{1658}{PaperLeanFormalization.Preliminaries.gaussian_loss_eq_kernel_energy_of_pair_moment}{a bridge} that assumes integrability of every feature pair, discharged by \leanlinkdecl{Preliminaries.lean}{1461}{PaperLeanFormalization.Preliminaries.centered_feature_pair_integrable}{\texttt{centered\_feature\_pair\_integrable}} and \leanlinkdecl{Preliminaries.lean}{1422}{PaperLeanFormalization.Preliminaries.gaussian_plain_pair_integrable}{\texttt{gaussian\_plain\_pair\_integrable}}; the squared centered residual and its product with the linear skip are integrable by \leanlinkdecl{Skip.lean}{314}{PaperLeanFormalization.Skip.centered_residual_square_integrable}{\texttt{centered\_residual\_square\_integrable}} and \leanlinkdecl{Skip.lean}{333}{PaperLeanFormalization.Skip.linear_centered_residual_integrable}{\texttt{linear\_centered\_residual\_integrable}}; and the strict inequality $0<\LR$ in \Cref{thm:plain-trap} is a Lean-checked witness that the losses are not identically zero.
\begin{lstlisting}[style=lean]
/-- The two hidden-unit features of the paper: the centered feature
`C(z) = |z| / 2` and the plain ReLU `R(z) = max 0 z`. -/
inductive Model where
  | centered
  | plainRelu
  deriving DecidableEq

/-- Standard Gaussian density with respect to Euclidean Lebesgue measure. -/
def stdGaussianDensity {d : ℕ} (x : EuclideanSpace ℝ (Fin d)) : ℝ :=
  (sqrt (2 * π))⁻¹ ^ d * exp (-‖x‖ ^ 2 / 2)

/-- Input space `ℝ^d`. -/
abbrev Vec (d : ℕ) := EuclideanSpace ℝ (Fin d)

/-- A width-`n` student: its masses `s` and its directions `w`. -/
abbrev Parameters (d n : ℕ) := (Fin n → ℝ) × (Fin n → Vec d)

/-- Centered population loss: the full formula in one place. -/
def CenteredLoss {d n m : ℕ} (s : Fin n → ℝ) (w : Fin n → Vec d)
    (t : Fin m → ℝ) (v : Fin m → Vec d) : ℝ :=
  (1 / 2 : ℝ) * ∫ x : Vec d,
    ((∑ i, s i * (|⟪w i, x⟫_ℝ| / 2)) -
      ∑ k, t k * (|⟪v k, x⟫_ℝ| / 2)) ^ 2 * stdGaussianDensity x

/-- Plain-ReLU population loss: the full formula in one place. -/
def PlainLoss {d n m : ℕ} (s : Fin n → ℝ) (w : Fin n → Vec d)
    (t : Fin m → ℝ) (v : Fin m → Vec d) : ℝ :=
  (1 / 2 : ℝ) * ∫ x : Vec d,
    ((∑ i, s i * max 0 ⟪w i, x⟫_ℝ) -
      ∑ k, t k * max 0 ⟪v k, x⟫_ℝ) ^ 2 * stdGaussianDensity x
\end{lstlisting}

\subsubsection{Contribution 1: Benign loss landscape with linear skips}
\label{app:lean-contribution-1}
\Cref{thm:headline} is the population statement of the paper's title: with a learned linear skip, a non-negative local minimum of the skip loss against a positive coplanar teacher network has zero loss whenever the student network is at least as wide. The skip loss is the only new definition.
\begin{lstlisting}[style=lean]
/-- Population loss of a plain-ReLU student with a learned linear skip
`bStudent` against a plain-ReLU teacher with linear skip `bTeacher`. -/
def SkipLoss {d n m : ℕ} (bStudent : Vec d) (s : Fin n → ℝ) (w : Fin n → Vec d)
    (bTeacher : Vec d) (t : Fin m → ℝ) (v : Fin m → Vec d) : ℝ :=
  (1 / 2 : ℝ) * ∫ x : Vec d,
    (⟪bStudent, x⟫_ℝ + (∑ i, s i * max 0 ⟪w i, x⟫_ℝ) -
      (⟪bTeacher, x⟫_ℝ + ∑ k, t k * max 0 ⟪v k, x⟫_ℝ)) ^ 2 *
        stdGaussianDensity x
\end{lstlisting}
\begin{lstlisting}[style=lean]
/-- 1. Linear skip: a positive coplanar teacher has zero loss at every
nonnegative local minimum once the student is at least as wide (`thm:headline`). -/
theorem Main {d n m : ℕ} (hd : 1 ≤ d) (hwidth : m ≤ n)
    {bStudent bTeacher : Vec d} {s : Fin n → ℝ} {w : Fin n → Vec d}
    {t : Fin m → ℝ} {v : Fin m → Vec d}
    (hplane : FiniteDimensional.finrank ℝ (Submodule.span ℝ (Set.range v)) ≤ 2)
    (hv : ∀ k, ‖v k‖ = 1) (hw : ∀ i, ‖w i‖ = 1)
    (hs : ∀ i, 0 ≤ s i) (ht : ∀ k, 0 < t k)
    (hmin : IsLocalMinOn
      (fun p : Vec d × Parameters d n => SkipLoss p.1 p.2.1 p.2.2 bTeacher t v)
      {p | ∀ i, ‖p.2.2 i‖ = 1} (bStudent, (s, w))) :
    SkipLoss bStudent s w bTeacher t v = 0
\end{lstlisting}
\noindent The proof of \href{\leanbase/Overview_theorems.lean\#L57}{\texttt{Main}} applies \href{\leanbase/Theorems.lean\#L410}{\texttt{thm\_headline}} (proof: \href{\leanbase/Geometry.lean\#L5218}{Geometry.lean:\allowbreak{}5218}).

\subsubsection{Contribution 2: Spurious minima at arbitrary overparameterization}
\label{app:lean-contribution-2}
\Cref{thm:plain-trap} needs the fixed planar trap teacher network, the planar direction $e(\theta)$, and strict local minimality on the unit-direction parameter space; its three clauses are exactly the theorem's three conclusions.
\begin{lstlisting}[style=lean]
/-- The unit direction `e(θ) = (cos θ, sin θ)` of the plane. -/
def Angle (θ : ℝ) : Vec 2 := ![cos θ, sin θ]

/-- The fixed planar teacher of the trap: three units at angles
`0`, `5π/6`, and `4π/3`, ... -/
def TrapTeacherAngle : Fin 3 → ℝ := ![0 * π, (5 / 6 : ℝ) * π, (4 / 3 : ℝ) * π]

/-- ... each of unit mass. -/
def TrapTeacherMass : Fin 3 → ℝ := fun _ => 1

/-- Strict local minimality within a set: all distinct sufficiently nearby
feasible parameters have strictly larger loss. -/
def IsStrictLocalMinOn {X : Type*} [TopologicalSpace X]
    (f : X → ℝ) (s : Set X) (x : X) : Prop :=
  ∀ᶠ y in (*@$\mathcal{N}$@*)[s] x, y ≠ x → f x < f y
\end{lstlisting}
\begin{lstlisting}[style=lean]
/-- 2. Spurious local minima at arbitrary overparameterization (`thm:plain-trap`).
The fixed teacher has unit masses at angles `0`, `5π/6`, and `4π/3`.

The three clauses are exactly the theorem's conclusions: at every width at
least three there is a positive-mass spurious minimum; at width three one is
strict; at larger widths one is non-strict. The first clause expresses
spuriousness by exhibiting a feasible point of strictly smaller loss.
The strict and non-strict witnesses have positive loss; copying the teacher
and adding zero-mass units gives a zero-loss competitor at these widths. -/
theorem PlainTrap :
    (∀ {n : ℕ} (_ : 3 ≤ n),
      ∃ s : Fin n → ℝ, ∃ w : Fin n → Vec 2,
        (∀ i, ‖w i‖ = 1) ∧ (∀ i, 0 < s i) ∧
        IsLocalMinOn (fun p : Parameters 2 n =>
          PlainLoss p.1 p.2 TrapTeacherMass (fun k => Angle (TrapTeacherAngle k)))
          {p | ∀ i, ‖p.2 i‖ = 1} (s, w) ∧
        ∃ s₀ : Fin n → ℝ, ∃ w₀ : Fin n → Vec 2,
          (∀ i, ‖w₀ i‖ = 1) ∧
          PlainLoss s₀ w₀ TrapTeacherMass (fun k => Angle (TrapTeacherAngle k)) <
            PlainLoss s w TrapTeacherMass (fun k => Angle (TrapTeacherAngle k))) ∧
    (∃ s : Fin 3 → ℝ, ∃ w : Fin 3 → Vec 2,
        (∀ i, ‖w i‖ = 1) ∧ (∀ i, 0 < s i) ∧
        IsStrictLocalMinOn (fun p : Parameters 2 3 =>
          PlainLoss p.1 p.2 TrapTeacherMass (fun k => Angle (TrapTeacherAngle k)))
          {p | ∀ i, ‖p.2 i‖ = 1} (s, w) ∧
        0 < PlainLoss s w TrapTeacherMass (fun k => Angle (TrapTeacherAngle k))) ∧
    (∀ {n : ℕ} (_ : 3 < n),
      ∃ s : Fin n → ℝ, ∃ w : Fin n → Vec 2,
        (∀ i, ‖w i‖ = 1) ∧ (∀ i, 0 < s i) ∧
        IsLocalMinOn (fun p : Parameters 2 n =>
          PlainLoss p.1 p.2 TrapTeacherMass (fun k => Angle (TrapTeacherAngle k)))
          {p | ∀ i, ‖p.2 i‖ = 1} (s, w) ∧
        0 < PlainLoss s w TrapTeacherMass (fun k => Angle (TrapTeacherAngle k)) ∧
        ¬ IsStrictLocalMinOn (fun p : Parameters 2 n =>
          PlainLoss p.1 p.2 TrapTeacherMass (fun k => Angle (TrapTeacherAngle k)))
          {p | ∀ i, ‖p.2 i‖ = 1} (s, w))
\end{lstlisting}
\noindent The proof of \href{\leanbase/Overview_theorems.lean\#L78}{\texttt{PlainTrap}} applies \href{\leanbase/Theorems.lean\#L516}{\texttt{thm\_plain\_trap\_spurious}} (proof: \href{\leanbase/Plain.lean\#L10254}{Plain.lean:\allowbreak{}10254}), \href{\leanbase/Theorems.lean\#L530}{\texttt{thm\_plain\_trap\_strict}} (proof: \href{\leanbase/Plain.lean\#L10271}{Plain.lean:\allowbreak{}10271}), \href{\leanbase/Theorems.lean\#L540}{\texttt{thm\_plain\_trap\_flat}} (proof: \href{\leanbase/Plain.lean\#L10292}{Plain.lean:\allowbreak{}10292}).

\subsubsection{Contribution 3: Learning the teacher subspace}
\label{app:lean-contribution-3}
\Cref{prop:teacher-span-confinement} holds for both features, so the loss is selected by the \lstinline[style=lean]|Model|. The conclusion is containment of every positive-mass student direction in the teacher span.
\begin{lstlisting}[style=lean]
/-- The population loss of either feature, selected by the `Model`. -/
def Loss {d n m : ℕ} : Model → (Fin n → ℝ) → (Fin n → Vec d) →
    (Fin m → ℝ) → (Fin m → Vec d) → ℝ
  | .centered => CenteredLoss
  | .plainRelu => PlainLoss
\end{lstlisting}
\begin{lstlisting}[style=lean]
/-- 3. Teacher-span confinement for both features
(`prop:teacher-span-confinement`).
Every student direction with positive mass lies in the teacher span.
This is containment; it does not assert that the student spans the whole space. -/
theorem Confinement {d n m : ℕ} (hd : 2 ≤ d) (q : Model)
    {s : Fin n → ℝ} {w : Fin n → Vec d} {t : Fin m → ℝ} {v : Fin m → Vec d}
    (hv : ∀ k, ‖v k‖ = 1) (hw : ∀ i, ‖w i‖ = 1) (hs : ∀ i, 0 ≤ s i)
    (hmin : IsLocalMinOn (fun p : Parameters d n => Loss q p.1 p.2 t v)
      {p | ∀ i, ‖p.2 i‖ = 1} (s, w)) :
    ∀ i, 0 < s i → w i ∈ Submodule.span ℝ (Set.range v)
\end{lstlisting}
\noindent The proof of \href{\leanbase/Overview_theorems.lean\#L111}{\texttt{Confinement}} applies \href{\leanbase/Theorems.lean\#L189}{\texttt{prop\_teacher\_span\_confinement}} (proof: \href{\leanbase/Geometry.lean\#L4286}{Geometry.lean:\allowbreak{}4286}).

\subsubsection{Contribution 4: Effective width}
\label{app:lean-contribution-4}
\Cref{prop:plain-collision-ceiling} is stated in the plane with directions \lstinline[style=lean]|Angle θ| and needs no further definition: a critical point is a zero of the Fréchet derivative in the mass and angle coordinates, and oriented directions are counted with period $2\pi$. The hypothesis is written with Mathlib's \lstinline[style=lean]|fderiv|, which is $0$ at a point of non-differentiability; it is not vacuous here, because in mass--angle coordinates the plain loss is an affine rescaling of the finite kernel loss (\leanlinkdecl{Plain.lean}{8645}{PaperLeanFormalization.Plain.plain_loss_eq_excess_div}{\texttt{plain\_loss\_eq\_excess\_div}}), which is globally $C^2$ (\leanlinkdecl{Plain.lean}{1089}{PaperLeanFormalization.Plain.ParameterCalculus.loss_contDiff_two}{\texttt{loss\_contDiff\_two}}); the hypothesis therefore states that the genuine Fréchet derivative vanishes, equivalently \lstinline[style=lean]|HasFDerivAt … 0 (s, θ)|.
\begin{lstlisting}[style=lean]
/-- 4. Effective width (`prop:plain-collision-ceiling`).
In dimension two, every positive-mass plain-ReLU critical point has at most
`2m` distinct oriented student directions. The second clause supplies
attainment: for every odd teacher width at least three there is a critical
point with exactly `2m` directions. Opposite directions count separately.
The `fderiv` hypothesis is a genuine derivative condition: see the conventions
in the module docstring (`Plain.plain_loss_eq_excess_div`,
`Plain.ParameterCalculus.loss_contDiff_two`). -/
theorem EffectiveWidth :
    (∀ {n m : ℕ} {s θ : Fin n → ℝ} {t β : Fin m → ℝ}
        (_ : ∀ i, 0 < s i)
        (_ : fderiv ℝ (fun p : (Fin n → ℝ) × (Fin n → ℝ) =>
          PlainLoss p.1 (fun i => Angle (p.2 i)) t (fun k => Angle (β k))) (s, θ) = 0),
      (Finset.univ.image (fun i => Angle (θ i))).card ≤ 2 * m) ∧
    (∀ (m : ℕ) (_ : 3 ≤ m) (_ : Odd m),
      ∃ (s θ : Fin (m + m) → ℝ) (t β : Fin m → ℝ),
        (∀ i, 0 < s i) ∧ (∀ k, 0 < t k) ∧
        fderiv ℝ (fun p : (Fin (m + m) → ℝ) × (Fin (m + m) → ℝ) =>
          PlainLoss p.1 (fun i => Angle (p.2 i)) t (fun k => Angle (β k))) (s, θ) = 0 ∧
        (Finset.univ.image (fun i => Angle (θ i))).card = 2 * m)
\end{lstlisting}
\noindent The proof of \href{\leanbase/Overview_theorems.lean\#L130}{\texttt{EffectiveWidth}} applies \href{\leanbase/Theorems.lean\#L557}{\texttt{prop\_plain\_collision\_ceiling}} (proof: \href{\leanbase/Plain.lean\#L10531}{Plain.lean:\allowbreak{}10531}), \href{\leanbase/Theorems.lean\#L566}{\texttt{prop\_plain\_collision\_ceiling\_sharp}} (proof: \href{\leanbase/Plain.lean\#L10602}{Plain.lean:\allowbreak{}10602}).

\subsubsection{Contribution 5: Transfer to empirical loss}
\label{app:lean-contribution-5}
\Cref{thm:fixed-radius-bridge} replaces the Gaussian distribution by a finite dataset. The definitions fix the empirical loss, the joint parameter distance through the matching skip, the accuracy condition against the Gaussian functional on degree-two homogeneous tests, and the fixed-radius minimization hypothesis; the theorem then produces one tolerance $\delta$ before the dataset, the labels, the skips, and the student network are chosen.
\begin{lstlisting}[style=lean]
/-- The linear skip that cancels the first Gaussian moment of the residual;
`JointDistance` measures the student's skip relative to it. -/
def MatchingSkip {d n m : ℕ} (bTeacher : Vec d) (t : Fin m → ℝ)
    (v : Fin m → Vec d) (s : Fin n → ℝ) (w : Fin n → Vec d) : Vec d :=
  bTeacher + (1 / 2 : ℝ) • ((∑ k, t k • v k) - ∑ i, s i • w i)

/-- The observed dataset's average squared radius. No directional weights occur. -/
def DataSecondMoment {d N : ℕ} (x : Fin N → Vec d) : ℝ :=
  (∑ j, ‖x j‖ ^ 2) / (N : ℝ)

/-- Ordinary unweighted empirical loss for input vectors and observed labels.
The literal formula also covers zero input vectors; results requiring a
nonempty dataset state `0 < N` explicitly. -/
def DataLoss {d n N : ℕ} (x : Fin N → Vec d) (y : Fin N → ℝ)
    (bStudent : Vec d) (s : Fin n → ℝ) (w : Fin n → Vec d) : ℝ :=
  ((1 / 2 : ℝ) * ∑ j,
    (⟪bStudent, x j⟫_ℝ + (∑ i, s i * max 0 ⟪w i, x j⟫_ℝ) - y j) ^ 2) / (N : ℝ)

/-- Maximum metric in masses, directions, and matching-skip mismatch. -/
def JointDistance {d n m : ℕ} (bTeacher : Vec d) (t : Fin m → ℝ) (v : Fin m → Vec d)
    (p q : Vec d × Parameters d n) : ℝ :=
  max (max (dist p.2.1 q.2.1) (dist p.2.2 q.2.2))
    (dist (p.1 - MatchingSkip bTeacher t v p.2.1 p.2.2)
      (q.1 - MatchingSkip bTeacher t v q.2.1 q.2.2))

/-- The Gaussian expectation of a test function. -/
def gaussianFunctional (d : ℕ) (f : Vec d → ℝ) : ℝ :=
  ∫ x : Vec d, f x * stdGaussianDensity x

/-- Positive homogeneity of degree two, the scaling law shared by every
squared residual of the networks. -/
def IsPosHomogeneousDegTwo {d : ℕ} (F : Vec d → ℝ) : Prop :=
  ∀ τ : ℝ, 0 ≤ τ → ∀ x : Vec d, F (τ • x) = τ ^ 2 * F x

/-- Accuracy of the actual unweighted input dataset against the Gaussian
functional for continuous degree-two homogeneous tests. The observed second
moment supplies the radial normalization. This does not construct or assume
a spherical quadrature rule. -/
def DataAccuracy {d N : ℕ} (x : Fin N → Vec d) (δ : ℝ) : Prop :=
  ∀ H : Vec d → ℝ, Continuous H → IsPosHomogeneousDegTwo H →
    ∀ L : ℝ,
      (∀ z z' : Vec d, ‖z‖ ≤ 1 → ‖z'‖ ≤ 1 → |H z - H z'| ≤ L * ‖z - z'‖) →
      |(d : ℝ) * ((∑ j, H (x j)) / (N : ℝ)) -
          DataSecondMoment x * gaussianFunctional d H| ≤
        (d : ℝ) * DataSecondMoment x * L * δ

/-- The observed labels come exactly from the teacher's plain-ReLU network
with its linear skip. Here `x j` is input number `j`, `y j` its label,
`t k` the mass of teacher unit `k`, and `v k` its direction.
There is no observation noise in this hypothesis. -/
def TeacherGeneratedLabels {d m N : ℕ} (x : Fin N → Vec d) (y : Fin N → ℝ)
    (bTeacher : Vec d) (t : Fin m → ℝ) (v : Fin m → Vec d) : Prop :=
  ∀ j, y j = ⟪bTeacher, x j⟫_ℝ + ∑ k, t k * max 0 ⟪v k, x j⟫_ℝ

/-- The candidate minimizes empirical loss over the entire closed ball of
radius `r`, intersected with the unit-direction parameter space.

A parameter is `(linear skip, (masses, directions))`. The distance is
`JointDistance`, defined above: the maximum of the mass distance, direction
distance, and distance between matching-skip mismatches. Thus this is the
paper's joint parameter ball, with its stated choice of coordinates.

`IsMinOn` means the candidate's loss is no larger than the loss at *every*
feasible point in this ball. This is stronger than ordinary local minimality,
which only promises some possibly smaller neighborhood. The ball permits
signed mass perturbations; nonnegative masses are required only at the
candidate in `ZeroLossForFixedRadiusMinima` below. -/
def MinimizesOnJointBall {d n m N : ℕ} (x : Fin N → Vec d) (y : Fin N → ℝ)
    (bTeacher : Vec d) (t : Fin m → ℝ) (v : Fin m → Vec d) (r : ℝ)
    (candidate : Vec d × Parameters d n) : Prop :=
  IsMinOn (fun p : Vec d × Parameters d n => DataLoss x y p.1 p.2.1 p.2.2)
    {p | (∀ i, ‖p.2.2 i‖ = 1) ∧ JointDistance bTeacher t v p candidate ≤ r}
    candidate

/-- The desired deterministic conclusion for one dataset, one teacher's
hidden units, one student width, and one radius.

For every teacher skip and candidate student: if the labels come from that
teacher, the candidate has nonnegative masses and unit directions, and it
minimizes on the prescribed ball, then its empirical loss is zero.
For nonempty data, zero loss means every observed label is fitted exactly. -/
def ZeroLossForFixedRadiusMinima {d m N : ℕ} (n : ℕ)
    (x : Fin N → Vec d) (y : Fin N → ℝ)
    (t : Fin m → ℝ) (v : Fin m → Vec d) (r : ℝ) : Prop :=
  ∀ (bTeacher bStudent : Vec d) (s : Fin n → ℝ) (w : Fin n → Vec d),
    TeacherGeneratedLabels x y bTeacher t v →
    (∀ i, 0 ≤ s i) → (∀ i, ‖w i‖ = 1) →
    MinimizesOnJointBall x y bTeacher t v r (bStudent, (s, w)) →
    DataLoss x y bStudent s w = 0
\end{lstlisting}
\begin{lstlisting}[style=lean]
/-- 5. Transfer to empirical loss (`thm:fixed-radius-bridge`).

Fix the dimensions, teacher hidden units, and a positive radius `r`.
The teacher has positive masses and unit directions spanning at most two
dimensions; the student is at least as wide as the teacher. Every positive
input dimension is covered; in dimension one the conclusion holds for every
dataset, so the tolerance is not needed there.

There is an accuracy tolerance `δ > 0` that works for *every* nonempty dataset
and every candidate student satisfying the displayed conditions. In particular,
`δ` is chosen before the sample size, inputs, labels, skip vectors, and student
parameters. `DataAccuracy` is the explicit Gaussian approximation condition
in `Overview_definitions.lean`, using the observed average squared input radius. -/
theorem EmpiricalTransfer {d n m : ℕ} (hd : 1 ≤ d) (hwidth : m ≤ n)
    {t : Fin m → ℝ} {v : Fin m → Vec d}
    (hplane : FiniteDimensional.finrank ℝ (Submodule.span ℝ (Set.range v)) ≤ 2)
    (hteacherDirections : ∀ k, ‖v k‖ = 1) (hteacherMasses : ∀ k, 0 < t k)
    {r : ℝ} (hr : 0 < r) :
    ∃ δ : ℝ, 0 < δ ∧
      ∀ (N : ℕ) (x : Fin N → Vec d) (y : Fin N → ℝ),
        0 < N → DataAccuracy x δ → ZeroLossForFixedRadiusMinima n x y t v r
\end{lstlisting}
\noindent The proof of \href{\leanbase/Overview_theorems.lean\#L157}{\texttt{EmpiricalTransfer}} applies \href{\leanbase/Theorems.lean\#L593}{\texttt{thm\_fixed\_radius\_bridge}} (proof: \href{\leanbase/Empirical.lean\#L1599}{Empirical.lean:\allowbreak{}1599}).

\subsection{Statements and proofs}
\label{app:lean-statements}
The statement column is the paper-facing contract in \texttt{Theorems.lean}; the proof column is the chapter declaration it delegates to. Separate rows distinguish clauses of one statement.

\begingroup
\footnotesize
\setlength{\tabcolsep}{3pt}
\renewcommand{\arraystretch}{1.12}
\begin{longtable}{@{}>{\raggedright\arraybackslash}p{0.25\linewidth}>{\raggedright\arraybackslash}p{0.19\linewidth}>{\raggedright\arraybackslash}p{0.19\linewidth}>{\raggedright\arraybackslash}p{0.31\linewidth}@{}}
\toprule
Paper statement / clause & Statement in Lean & Proof & Scope \\
\midrule
\endfirsthead
\toprule
Paper statement / clause & Statement in Lean & Proof & Scope \\
\midrule
\endhead
\bottomrule
\endfoot
Contribution 1: \Cref{thm:headline} & \href{\leanbase/Overview_theorems.lean\#L57}{Overview\_\allowbreak{}theorems.lean:\allowbreak{}57} & \Cref{app:lean-contribution-1} & Benign loss landscape with linear skips; exact contract of \Cref{thm:headline}. \\
Contribution 2: \Cref{thm:plain-trap} & \href{\leanbase/Overview_theorems.lean\#L78}{Overview\_\allowbreak{}theorems.lean:\allowbreak{}78} & \Cref{app:lean-contribution-2} & Spurious minima at arbitrary overparameterization; exact contract of \Cref{thm:plain-trap}. \\
Contribution 3: \Cref{prop:teacher-span-confinement} & \href{\leanbase/Overview_theorems.lean\#L111}{Overview\_\allowbreak{}theorems.lean:\allowbreak{}111} & \Cref{app:lean-contribution-3} & Learning the teacher subspace; exact contract of \Cref{prop:teacher-span-confinement}. \\
Contribution 4: \Cref{prop:plain-collision-ceiling} & \href{\leanbase/Overview_theorems.lean\#L130}{Overview\_\allowbreak{}theorems.lean:\allowbreak{}130} & \Cref{app:lean-contribution-4} & Effective width; exact contract of \Cref{prop:plain-collision-ceiling}. \\
Contribution 5: \Cref{thm:fixed-radius-bridge} & \href{\leanbase/Overview_theorems.lean\#L157}{Overview\_\allowbreak{}theorems.lean:\allowbreak{}157} & \Cref{app:lean-contribution-5} & Transfer to empirical loss; exact contract of \Cref{thm:fixed-radius-bridge}. \\
Definition~\ref{def:networks} & --- & --- & Definition components and parameter conventions; no separate theorem binding. See \eqref{eq:models-raw}. \\
Definition~\ref{def:losses} & --- & --- & Definition components and parameter conventions; no separate theorem binding. See \eqref{eq:losses}, \eqref{eq:q-losses}. \\
Definition~\ref{def:minima} & --- & --- & Definition components and parameter conventions; no separate theorem binding. See \eqref{eq:parameter-classes}. \\
Lemma~\ref{lem:raw-transfer}\newline q & \href{\leanbase/Theorems.lean\#L40}{Theorems.lean:\allowbreak{}40} & \href{\leanbase/Preliminaries.lean\#L2703}{Preliminaries.lean:\allowbreak{}2703} & Exact contract and local proof. \\
Lemma~\ref{lem:raw-transfer}\newline linear skip & \href{\leanbase/Theorems.lean\#L59}{Theorems.lean:\allowbreak{}59} & \href{\leanbase/Preliminaries.lean\#L2727}{Preliminaries.lean:\allowbreak{}2727} & Exact contract and local proof. \\
Lemma~\ref{lem:equivalence_loss_skip_centered}\newline critical & \href{\leanbase/Theorems.lean\#L116}{Theorems.lean:\allowbreak{}116} & \href{\leanbase/Skip.lean\#L604}{Skip.lean:\allowbreak{}604} & Exact contract and local proof. \\
Lemma~\ref{lem:equivalence_loss_skip_centered}\newline identities & \href{\leanbase/Theorems.lean\#L83}{Theorems.lean:\allowbreak{}83} & \href{\leanbase/Skip.lean\#L513}{Skip.lean:\allowbreak{}513} & Exact contract and local proof. \\
Lemma~\ref{lem:equivalence_loss_skip_centered}\newline local minimum & \href{\leanbase/Theorems.lean\#L93}{Theorems.lean:\allowbreak{}93} & \href{\leanbase/Skip.lean\#L564}{Skip.lean:\allowbreak{}564} & Exact contract and local proof. \\
Lemma~\ref{lem:equivalence_loss_skip_centered}\newline strict local minimum & \href{\leanbase/Theorems.lean\#L104}{Theorems.lean:\allowbreak{}104} & \href{\leanbase/Skip.lean\#L583}{Skip.lean:\allowbreak{}583} & Exact contract and local proof. \\
Lemma~\ref{lem:pair-moments} & \href{\leanbase/Theorems.lean\#L133}{Theorems.lean:\allowbreak{}133} & \href{\leanbase/Preliminaries.lean\#L1649}{Preliminaries.lean:\allowbreak{}1649} & Exact contract and local proof. \\
Proposition~\ref{prop:loss_in_residual} & \href{\leanbase/Theorems.lean\#L146}{Theorems.lean:\allowbreak{}146} & \href{\leanbase/Preliminaries.lean\#L2297}{Preliminaries.lean:\allowbreak{}2297} & Energy pairing, actual mass derivative, and the actual spherical directional derivative along a great circle; no replacement by an arbitrary named gradient expression. \\
Proposition~\ref{prop:teacher-span-confinement} & \href{\leanbase/Theorems.lean\#L189}{Theorems.lean:\allowbreak{}189} & \href{\leanbase/Geometry.lean\#L4286}{Geometry.lean:\allowbreak{}4286} & Input dimension at least two; both activation models; non-negative student masses, arbitrary signed teacher masses, unit directions and local minimality. Active student directions lie in the teacher span. \\
Proposition~\ref{prop:zero_mass_exact_fit} & \href{\leanbase/Theorems.lean\#L199}{Theorems.lean:\allowbreak{}199} & \href{\leanbase/Geometry.lean\#L4774}{Geometry.lean:\allowbreak{}4774} & Exact contract and local proof. \\
Proposition~\ref{prop:nc-dead-student} & \href{\leanbase/Theorems.lean\#L211}{Theorems.lean:\allowbreak{}211} & \href{\leanbase/Geometry.lean\#L4815}{Geometry.lean:\allowbreak{}4815} & Exact contract and local proof. \\
Lemma~\ref{lem:nc-dead-counterexamples}\newline line, (a) & \href{\leanbase/Theorems.lean\#L228}{Theorems.lean:\allowbreak{}228} & \href{\leanbase/Plain.lean\#L11111}{Plain.lean:\allowbreak{}11111} & The opposed zero-mass student unit on the line is a local minimum on the unit-direction parameter space, with the explicit loss $(s^2+1)/4$ of the opposed student unit and loss $0$ for the teacher copy. \\
Lemma~\ref{lem:nc-dead-counterexamples}\newline plane, (b) & \href{\leanbase/Theorems.lean\#L228}{Theorems.lean:\allowbreak{}228} & \href{\leanbase/Plain.lean\#L10907}{Plain.lean:\allowbreak{}10907} & For the fixed five-unit signed teacher network and either activation, every zero-mass one-student parameter with angle in $(-\pi/6,\pi/6)$ is a local minimum on the unit-direction space with positive loss and a lower-loss one-student competitor. \\
Remark~\ref{rem:planar_residual_equivalence} & \href{\leanbase/Theorems.lean\#L170}{Theorems.lean:\allowbreak{}170} & \href{\leanbase/Planar.lean\#L5025}{Planar.lean:\allowbreak{}5025} & Exact contract and local proof. \\
Lemma~\ref{lem:critical-points-are-double-zeros} & \href{\leanbase/Theorems.lean\#L258}{Theorems.lean:\allowbreak{}258} & \href{\leanbase/Planar.lean\#L5100}{Planar.lean:\allowbreak{}5100} & Exact contract and local proof. \\
Lemma~\ref{lem:beam-kernel} & \href{\leanbase/Theorems.lean\#L268}{Theorems.lean:\allowbreak{}268} & \href{\leanbase/Beam.lean\#L169}{Beam.lean:\allowbreak{}169} & Exact contract and local proof. \\
Lemma~\ref{lem:gap-sign} & \href{\leanbase/Theorems.lean\#L275}{Theorems.lean:\allowbreak{}275} & \href{\leanbase/Planar.lean\#L4755}{Planar.lean:\allowbreak{}4755} & Canonical representatives in [0,pi), translated left cut at zero. The Green representation, empty-gap equivalence, sign and both endpoint curvatures are all included. \\
Lemma~\ref{lem:line-count} & \href{\leanbase/Theorems.lean\#L299}{Theorems.lean:\allowbreak{}299} & \href{\leanbase/Planar.lean\#L4860}{Planar.lean:\allowbreak{}4860} & Canonical real line representatives, positive/negative net supports, double zeros and cyclic weak interlacing; the support cardinality consequence is included. \\
Lemma~\ref{lem:opposed-rotation} & \href{\leanbase/Theorems.lean\#L314}{Theorems.lean:\allowbreak{}314} & \href{\leanbase/Planar.lean\#L5192}{Planar.lean:\allowbreak{}5192} & Exact contract and local proof. \\
Lemma~\ref{lem:S-is-small} & \href{\leanbase/Theorems.lean\#L347}{Theorems.lean:\allowbreak{}347} & \href{\leanbase/Geometry.lean\#L5021}{Geometry.lean:\allowbreak{}5021} & Exact contract and local proof. \\
Proposition~\ref{prop:S-plus-eq-student-lines} & \href{\leanbase/Theorems.lean\#L325}{Theorems.lean:\allowbreak{}325} & \href{\leanbase/Planar.lean\#L5213}{Planar.lean:\allowbreak{}5213} & Exact contract and local proof. \\
Lemma~\ref{lem:no-collisions} & \href{\leanbase/Theorems.lean\#L336}{Theorems.lean:\allowbreak{}336} & \href{\leanbase/Planar.lean\#L5241}{Planar.lean:\allowbreak{}5241} & Exact contract and local proof. \\
Proposition~\ref{prop:locamin_are_line_separated} & \href{\leanbase/Theorems.lean\#L358}{Theorems.lean:\allowbreak{}358} & \href{\leanbase/Planar.lean\#L5256}{Planar.lean:\allowbreak{}5256} & Exact contract and local proof. \\
Proposition~\ref{prop:beam-descent} & \href{\leanbase/Theorems.lean\#L370}{Theorems.lean:\allowbreak{}370} & \href{\leanbase/Planar.lean\#L5575}{Planar.lean:\allowbreak{}5575} & Actual Gaussian affine second derivative is strictly negative. The fresh proof derives interlacing and constructs its direction from the stated positive equal-width separated critical configuration. \\
Theorem~\ref{thm:centered} & \href{\leanbase/Theorems.lean\#L397}{Theorems.lean:\allowbreak{}397} & \href{\leanbase/Geometry.lean\#L5198}{Geometry.lean:\allowbreak{}5198} & All $d\ge1$; the scalar total-mass proof handles $d=1$, and the planar reduction handles $d\ge2$. Empty widths are included. \\
Theorem~\ref{thm:headline} & \href{\leanbase/Theorems.lean\#L410}{Theorems.lean:\allowbreak{}410} & \href{\leanbase/Geometry.lean\#L5218}{Geometry.lean:\allowbreak{}5218} & All $d\ge1$; a fixed-direction residual-scaling path handles $d=1$, with the original loss decomposition for $d\ge2$. \\
Theorem~\ref{thm:plain-trap}\newline every width & \href{\leanbase/Theorems.lean\#L504}{Theorems.lean:\allowbreak{}504} & \href{\leanbase/Plain.lean\#L10233}{Plain.lean:\allowbreak{}10233} & Exact contract and local proof. \\
Theorem~\ref{thm:plain-trap}\newline flat above three & \href{\leanbase/Theorems.lean\#L540}{Theorems.lean:\allowbreak{}540} & \href{\leanbase/Plain.lean\#L10292}{Plain.lean:\allowbreak{}10292} & Exact contract and local proof. \\
Theorem~\ref{thm:plain-trap}\newline non-global & \href{\leanbase/Theorems.lean\#L516}{Theorems.lean:\allowbreak{}516} & \href{\leanbase/Plain.lean\#L10254}{Plain.lean:\allowbreak{}10254} & Explicit feasible same-width competitor with strictly smaller loss, establishing genuine non-global spuriousness. \\
Theorem~\ref{thm:plain-trap}\newline strict three & \href{\leanbase/Theorems.lean\#L530}{Theorems.lean:\allowbreak{}530} & \href{\leanbase/Plain.lean\#L10271}{Plain.lean:\allowbreak{}10271} & Exact contract and local proof. \\
Lemma~\ref{lem:box-certificate} & \href{\leanbase/Theorems.lean\#L428}{Theorems.lean:\allowbreak{}428} & \href{\leanbase/Plain.lean\#L582}{Plain.lean:\allowbreak{}582} & A C2 objective on arbitrary finite coordinate blocks, with a Euclidean Hessian floor and dual Euclidean center-gradient bound, has a unique strict minimum in the interior of the max-norm box. \\
Lemma~\ref{lem:box-certificate}\newline spurious & \href{\leanbase/Theorems.lean\#L448}{Theorems.lean:\allowbreak{}448} & \href{\leanbase/Plain.lean\#L645}{Plain.lean:\allowbreak{}645} & Positive affine normalization, positive actual loss on the box and an exact-fit competitor imply strict spuriousness. Positive mass coordinates are an optional separate consequence, not a premise for spuriousness. \\
Lemma~\ref{lem:positive-curvature-splitting} & \href{\leanbase/Theorems.lean\#L465}{Theorems.lean:\allowbreak{}465} & \href{\leanbase/Plain.lean\#L9968}{Plain.lean:\allowbreak{}9968} & Any positive finite partition of the chosen positive mass preserves the represented network, actual loss and local minimality when its residual curvature is positive. Two or more children give an injective constant-loss curve and non-strictness. The local provider indexes k+1 children; the central entry point explicitly adapts to every $k\ge1$. \\
Proposition~\ref{prop:plain-collision-ceiling}\newline bound & \href{\leanbase/Theorems.lean\#L557}{Theorems.lean:\allowbreak{}557} & \href{\leanbase/Plain.lean\#L10531}{Plain.lean:\allowbreak{}10531} & Exact contract and local proof. \\
Proposition~\ref{prop:plain-collision-ceiling}\newline sharp & \href{\leanbase/Theorems.lean\#L566}{Theorems.lean:\allowbreak{}566} & \href{\leanbase/Plain.lean\#L10602}{Plain.lean:\allowbreak{}10602} & Sharpness is witnessed for odd teacher widths $m\ge3$, as in the printed explicit family; not a claim of attainment for every m. \\
Remark~\ref{rem:finite-data-accuracy}\newline accuracy & \href{\leanbase/Theorems.lean\#L613}{Theorems.lean:\allowbreak{}613} & \href{\leanbase/Empirical.lean\#L2901}{Empirical.lean:\allowbreak{}2901} & Model-independent DataAccuracy for identically distributed inputs with an orthogonally invariant common distribution and finite positive second moment; one measurable high-probability event works for every sufficiently large sample size and every test. \\
Theorem~\ref{thm:minimizer-stability} & \href{\leanbase/Theorems.lean\#L581}{Theorems.lean:\allowbreak{}581} & \href{\leanbase/Empirical.lean\#L351}{Empirical.lean:\allowbreak{}351} & Exact contract and local proof. \\
Theorem~\ref{thm:fixed-radius-bridge} & \href{\leanbase/Theorems.lean\#L593}{Theorems.lean:\allowbreak{}593} & \href{\leanbase/Empirical.lean\#L1599}{Empirical.lean:\allowbreak{}1599} & All $d\ge1$; dimension one needs no accuracy hypothesis. Ordinary finite DataLoss with teacher-generated labels. The accuracy threshold is uniform over non-empty datasets, teacher linear skips and non-negative fixed-radius minima. Zero-input datasets are included; signed competitors remain in the ball. \\
\end{longtable}
\endgroup

\section{AI use statement}
\label{app:ai-use}

Generative AI played a substantial role in the research underlying
this paper, extending beyond language editing or transcription into
formal syntax.
We used language models for literature exploration, conjecture
generation, proof and counterexample search, informal mathematical
reasoning, numerical experimentation, development of supporting
software, auto-formalization, Lean~4 proof development, and proof
simplification.
We also used AI assistance for manuscript formulations and for
locating and interpreting relevant declarations in the Lean
development.
During July--September 2026, we used Codex with Sol and Claude Code with
Opus and Fable; the final re-formalization described below used Codex
with Astra Ultra.
Lean~4~\citep{deMouraUllrich2021} provided the formal proof-checking
environment.

The authors selected and refined the research questions, guided the
exploration, reviewed the mathematical meaning of proposed statements,
and reorganized the resulting arguments.
In particular, the proof ideas in the main paper were reconstructed
and written largely manually by the authors, with AI assistance for
formulations and navigation of the formal development.
This does not imply that their underlying ideas were exclusively
human-generated: AI was heavily utilized for idea generation and contributed substantially to the discovery and
development of the mathematics.
The authors take responsibility for the final statements, arguments,
citations, code, and accompanying artifacts.

As a rough measure of computational usage, the project consumed approximately 40 full weekly Codex Pro20 usage allowances and 20 full weekly Claude Max20 usage allowances in the time span from the beginning of July to the end of September 2026. We report usage in these terms because complete token counts are not available and, moreover, token counts alone do not capture the heterogeneous computational resources involved in long-horizon agentic workflows.

\subsection{Scope of the contribution and reproducibility}
\label{app:ai-reproducibility}

\noindent\textbf{This paper contributes mathematical results, not an empirical evaluation of an AI research system.}
We do not introduce the agent orchestration as a separately evaluated method, nor do we claim fully autonomous discovery, comparative model performance, proof-search success rates, or improvements in research efficiency.
The workflow account below documents the provenance of the results, the extent of AI assistance, and the limitations encountered during the project.

\subsection{Progression of the research workflow}
\label{app:ai-workflow}

The workflow evolved through the following phases.
The phases describe the progression of the project rather than a
single fixed pipeline that would have been planned or applied perfectly in serial.

\begin{itemize}
    \item \textbf{Phase 1: Text/TeX-only exploration.}
    We developed an agentic proof-search setup with specialized roles
    for literature exploration, conjecturing, proving, and
    verification/utility judging, and experimented with multiple
    language models and orchestration pipelines.
    Mathematical knowledge was organized as a directed hypergraph:
    statements formed the vertices, while proofs formed hyperedges
    connecting their required ingredients to their conclusions.
    Statements and proofs were stored in individual Markdown files.
    Python code orchestrated the workflow by invoking OpenCode with
    different models as subprocesses.
    Verification remained informal, relying on judge agents to
    examine arguments step by step or attempt to reproduce results.
    The physical analogy behind \Cref{sec:benign}, that the angular residual is the
    radial displacement of a thin elastic ring under point forces and hence obeys
    the classical ring equation (\Cref{rem:ring}), was found in this phase, when an
    agent searching the literature for the differential equation of the kernel
    matched it with the elasticity of curved rods.
    Hallucinations and failures of this verification process
    motivated the transition to formal proof checking.

    \item \textbf{Phase 2: Auto-formalization and consolidation.}
    We used language-model-assisted auto-formalization to translate
    selected exploratory results and supporting mathematics from
    the literature into a shared Lean~4 library.
    Compilation was supplemented by semantic review: we examined
    whether the formal statements expressed the intended
    mathematical claims, including their definitions, implicit
    assumptions, and quantifiers.
    We also checked the existence of objects satisfying relevant
    assumptions to guard against vacuous formulations.
    Prompts and reusable agent instructions were revised in
    response to formalization errors and discrepancies between
    formal statements and their intended meaning.

    \item \textbf{Phase 3: Agentic proof search with a shared
    Lean library.}
    Prompt- and skill-based agents explored focused subproblems,
    developed candidate proofs, reused verified lemmas across
    related settings, and were incentivized to combine intermediate results
    into stronger statements.
    Human guidance included selecting and decomposing research
    questions, reviewing proposed claims, and transferring useful
    ideas between subproblems. Any time the search would stall, the authors would ask the agents to read relevant literature or find new related work, brainstorm ideas and directions, or evaluate ideas given by the authors.
    When the proof search was at a branching point, the authors would discuss significance, difficulties and generalizability of proof techniques with the agents in order to decide where to focus next.
    Numerical experiments helped test conjectures and identify
    counterexamples, while failed approaches were preserved to
    inform subsequent searches.
    Candidate Lean proofs were compiled, and incorporation into
    the verified library required both a completed formal proof
    and review of the statement's mathematical meaning. Integration into the Lean library would also include a rebuild of its changed parts.
    When a major new result would have been integrated into the Lean library, the authors would check the respective Lean proposition and let the agents generalize from unnecessary theorem requirements.

    \item \textbf{Phase 4: Proof simplification and fresh searches.}
    We conducted additional searches to simplify arguments and
    reduce the complexity and build times of the Lean development.
    For selected statements, fresh searches received the statement
    and necessary definitions either in Lean or in natural language, but not the
    existing proof.
    These searches sought alternative or simpler arguments;
    proposed replacements were subject to the same formal and
    semantic checks.

    \item \textbf{Phase 5: Attempted proof translation and
    author-led exposition.}
    We used Codex and Claude Code to translate the earlier Lean
    proofs into an appendix of conventional mathematical proofs.
    Despite substantial prompting effort and repeated instructions
    and revisions concerning mathematical presentation, the
    generated drafts remained difficult to read and did not provide
    satisfactory exposition.
    We therefore extracted the proof ideas, formulas, and logical
    dependencies and substantially reorganized the arguments.
    The authors wrote the resulting proof ideas largely
    manually, while continuing to use AI for formulations and
    assistance in navigating and interpreting the Lean theorems.
    These reconstructed proof ideas, rather than the earlier
    automatically generated proof appendix, are the exposition
    presented in the main paper.
    The assessment of the generated drafts describes our experience
    in this project; it is not a systematic evaluation of the
    models' mathematical writing capabilities.

    \item \textbf{Phase 6: Re-formalization of the simplified
    proof structure.}
    The author-led reconstruction substantially simplified the
    organization of the arguments.
    We then instructed Codex with Astra Ultra to produce Lean proofs
    following the structure of the revised proof ideas.
    The resulting development forms the supplementary Lean
    certificate accompanying the current manuscript.
\end{itemize}

\subsection{Verification and author responsibility}
\label{app:ai-verification}

Admission into the certified development required complete Lean
proofs without \texttt{sorry}/\texttt{admit}, additional
project-specific axioms, or unchecked computational shortcuts.
Numerical certificates used in the proofs were justified by exact
Lean-checked arithmetic.
Compilation and transitive axiom audits were supplemented by review
of the relationship between the formal statements and the
mathematics stated in the paper.

For the five contribution theorem entry points, the accompanying
\texttt{verify.py} script checks the recorded statement types and
audits their axiom dependencies against propositional extensionality,
choice, and quotient soundness.
The coverage and correspondence of the formal development are
documented in \Cref{app:lean-certificates}.
The authors remain responsible for the mathematical exposition and
for the accuracy of all claims about the formal certificates.

\subsection{Released skills and limits of the workflow record}
\label{app:ai-skills}

\noindent\textbf{The supplementary material includes the final versions
of the agent skills used during the project as Markdown
(\texttt{.md}) files.}
These reusable instructions document how agents were directed during
the research and make the workflow inspectable.

The skills evolved during the project in response to failed
approaches, verification problems, and changing research needs.
The released files therefore represent their final versions; they
should not be interpreted as instructions that were used unchanged
in every earlier run.
They are also not a complete chronological record of all prompts,
model interactions, intermediate states, or rejected candidates.

We provide these materials to disclose substantive AI involvement
and document the procedures used, not as evidence that the workflow
is superior to alternatives or sufficient for autonomous discovery.
Accordingly, the absence of a benchmark or an ablation study of the
agent workflow limits conclusions about that workflow, but is
separate from the verification of the mathematical results claimed
in this paper.
Those results are supported by the mathematical arguments and the
accompanying certificates, with the authors responsible for their
correctness and correspondence.
\end{document}